\documentclass{article} 
\usepackage{iclr2026_conference,times}

\usepackage{amsmath,amsfonts,bm}

\def\eqref#1{equation~\ref{#1}}

\def\1{\bm{1}}

\DeclareMathAlphabet{\mathsfit}{\encodingdefault}{\sfdefault}{m}{sl}
\SetMathAlphabet{\mathsfit}{bold}{\encodingdefault}{\sfdefault}{bx}{n}

\def\gD{{\mathcal{D}}}

\def\gH{{\mathcal{H}}}

\def\gN{{\mathcal{N}}}

\def\gR{{\mathcal{R}}}

\usepackage{hyperref}
\usepackage{url}
\usepackage{xspace}
\usepackage{adjustbox}
\usepackage{booktabs}
\usepackage{amssymb}
\usepackage{wrapfig} 
\usepackage{tcolorbox}
\usepackage{subcaption}
\usepackage{enumitem} 
\usepackage{cite} 
\usepackage[table]{xcolor}

\definecolor{HeatRed}{HTML}{FF0000}
\definecolor{HeatOrange}{HTML}{FFA500}
\definecolor{HeatYellow}{HTML}{FFFF00}
\definecolor{HeatLightGreen}{HTML}{90EE90}
\definecolor{HeatGreen}{HTML}{008000}
\definecolor{HeatDarkGreen}{HTML}{006400}
\definecolor{HeatWhite}{HTML}{FFFFFF} 
\definecolor{HeatVeryLow}{HTML}{D73027} 
\definecolor{HeatLow}{HTML}{F46D43}
\definecolor{HeatMidLow}{HTML}{FDAE61}
\definecolor{HeatMid}{HTML}{FEE08B}
\definecolor{HeatMidHigh}{HTML}{D9EF8B}
\definecolor{HeatHigh}{HTML}{A6D96A}
\definecolor{HeatVeryHigh}{HTML}{66BD63}
\definecolor{HeatExcellent}{HTML}{1A9850}

\newcommand{\heatmapcell}[2]{%
    \ifdim #1pt < 10pt \cellcolor{HeatVeryLow}\else
    \ifdim #1pt < 25pt \cellcolor{HeatLow}\else
    \ifdim #1pt < 40pt \cellcolor{HeatMidLow}\else
    \ifdim #1pt < 55pt \cellcolor{HeatMid}\else
    \ifdim #1pt < 70pt \cellcolor{HeatMidHigh}\else
    \ifdim #1pt < 85pt \cellcolor{HeatHigh}\else
    \ifdim #1pt < 95pt \cellcolor{HeatVeryHigh}\else
    \cellcolor{HeatExcellent}\fi\fi\fi\fi\fi\fi\fi
    \textcolor{black}{#2}
}

\newcommand{\proj}{HaystackCraft\xspace}

\title{Haystack Engineering: Context Engineering for Heterogeneous and Agentic Long-Context Evaluation}

\iclrfinalcopy 

\author{Mufei Li\textsuperscript{†}\thanks{Work done during Mufei's internship at Meta}, Dongqi Fu\textsuperscript{‡}, Limei Wang\textsuperscript{‡}, Si Zhang\textsuperscript{‡}, Hanqing Zeng\textsuperscript{‡}, Kaan Sancak\textsuperscript{‡},\\ \textbf{Ruizhong Qiu}\textsuperscript{§}, \textbf{Haoyu Wang}\textsuperscript{†}, \textbf{Xiaoxin He}\textsuperscript{¶}, \textbf{Xavier Bresson}\textsuperscript{¶}, \textbf{Yinglong Xia}\textsuperscript{‡},\\
\textbf{Chonglin Sun}\textsuperscript{‡}, \textbf{Pan Li}\textsuperscript{†}
\bigskip \\
\textsuperscript{†}Georgia Institute of Technology, \texttt{\{mufei.li, haoyu.wang, panli\}@gatech.edu} \\
\textsuperscript{‡}Meta AI, \texttt{\{dongqifu, limeiwang, sizhang, zengh, kaansancak, yxia,}\\ \texttt{ clsun\}@meta.com} \\
\textsuperscript{§}University of Illinois Urbana–Champaign, \texttt{\{rq5\}@illinois.edu} \\
\textsuperscript{¶}National University of Singapore, \texttt{\{xiaoxin, xaviercs\}@comp.nus.edu.sg}
}

\begin{document}

\maketitle

\begin{abstract}
Modern long-context large language models (LLMs) perform well on synthetic ``needle-in-a-haystack'' (NIAH) benchmarks, but such tests overlook how noisy contexts arise from biased retrieval and agentic workflows. We argue that \textbf{haystack engineering} is necessary to construct noisy long contexts that faithfully capture key real-world factors---distraction from heterogeneous biased retrievers and cascading errors in agentic workflows---to test models' long-context robustness. We instantiate it through \textbf{\proj}, a new NIAH benchmark built on the full English Wikipedia hyperlink network with multi-hop questions. \proj evaluates how heterogeneous retrieval strategies (e.g., sparse, dense, hybrid, and graph-based) affect distractor composition, haystack ordering, and downstream LLM performance. \proj further extends NIAH to dynamic, LLM-dependent settings that simulate agentic operations, where models refine queries, reflect on their past reasonings, and decide when to stop. Experiments with 15 long-context models show that (1) while stronger dense retrievers can introduce more challenging distractors, graph-based reranking simultaneously improves retrieval effectiveness and mitigates more harmful distractors; (2) in agentic tests, even advanced models like Gemini 2.5 Pro and GPT-5 suffer cascading failures from self-generated distractors or struggle to perform early stops. These results highlight persistent challenges in agentic long-context reasoning and establish \proj as a valuable testbed for future progress. Our implementation is available at \href{https://github.com/Graph-COM/HaystackCraft}{https://github.com/Graph-COM/HaystackCraft}.
\end{abstract}

\section{Introduction}

Effective context engineering~\citep{mei2025survey}---optimizing information for LLMs' contexts---and robust long-context reasoning are essential for large language models (LLMs) to enable sophisticated agents and handle complex, information-intensive tasks. Recent algorithmic and engineering innovations have significantly expanded LLMs' context windows and enhanced their long-context reasoning capabilities~\citep{SU2024127063, peng2024yarn, dao2022flashattention, dao2023flashattention2, liu2024ringattention, xiao2024efficient, yuan-etal-2025-native, kwon2023efficient}. Consequently, modern LLMs can process extended contexts and often achieve near-perfect recall on synthetic ``\textbf{needle-in-a-haystack (NIAH)}'' benchmarks~\citep{yen2025helmet}. These benchmarks test whether a model can retrieve and reason over relevant information \textit{needle} buried in a large \textit{haystack} context that also contains many distractors. Yet such synthetic setups neglect a fundamental question: how are noisy long contexts constructed in real-world applications?

To engineer long contexts in practice, retrieval-augmented generation (RAG)~\citep{NEURIPS2020_6b493230} is one of the most widely adopted strategies, where external retrievers rank candidate context documents with respect to queries. However, retrieval systems are imperfect and inherently biased, introducing retriever-dependent ranked distractors. To be specific, sparse retrievers such as BM25~\citep{conf/trec/RobertsonWJHG94, 10.1561/1500000019} often populate haystacks with lexically similar but semantically irrelevant documents, while dense retrievers~\citep{karpukhin-etal-2020-dense} surface semantically close but potentially factually incorrect ``near misses''. Because no single retriever is universally optimal~\citep{thakur2021beir}, it is crucial to study how heterogeneous retrieval strategies shape the context and consequently affect NIAH performance.

\begin{figure}
    \centering
    \includegraphics[width=\linewidth]{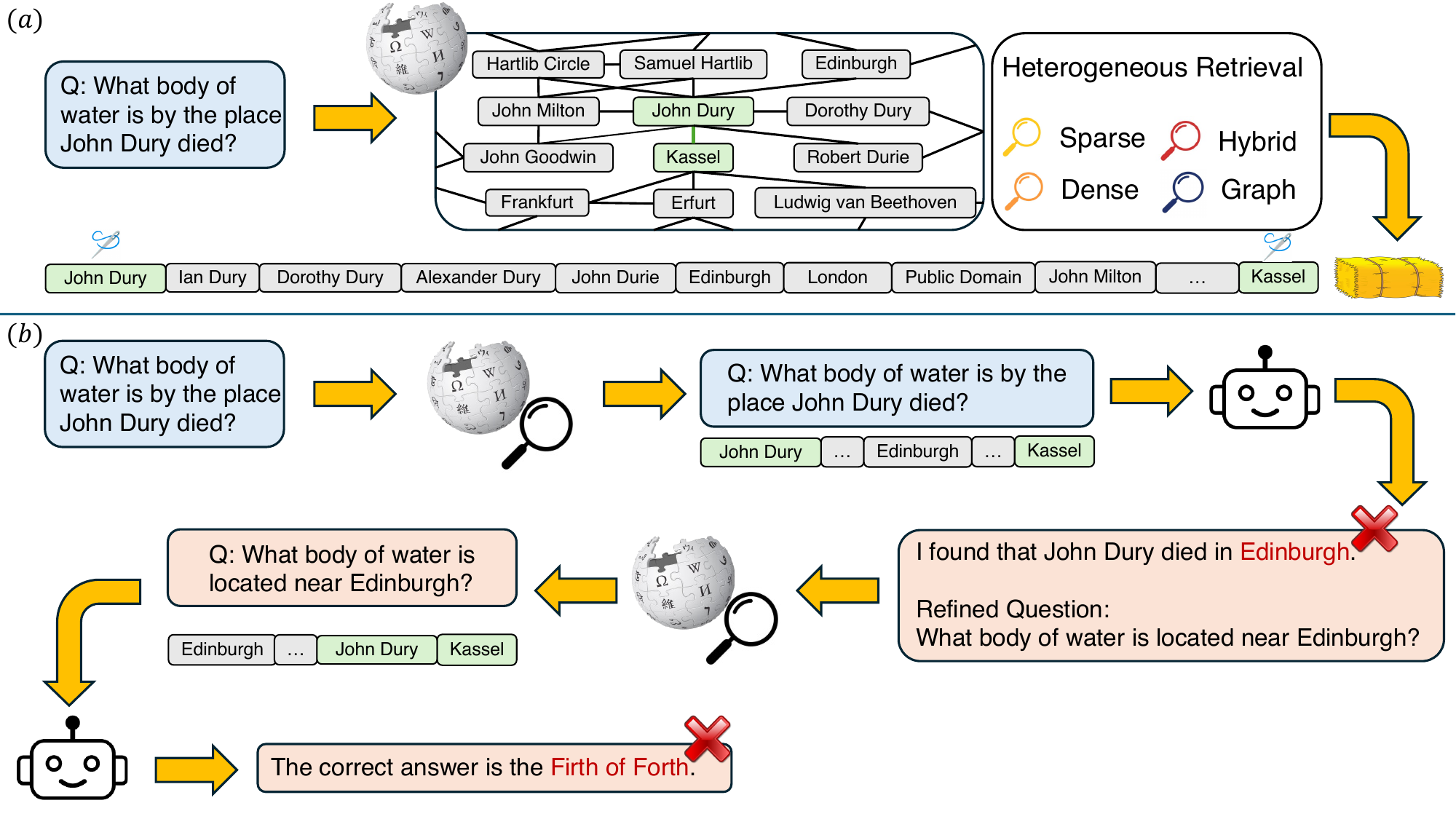}
    \caption{Overview of the core challenges that \proj addresses. 
    (a) \textbf{Retrieval-Dependent Haystacks.} The composition and ordering of the noisy long context (``haystack'') are shaped by the retrieval strategy (e.g., sparse, dense, hybrid, and graph-based). (b) \textbf{Agentic Error Propagation.} In dynamic agentic workflows, early errors—such as misidentifying John Dury's death place—can propagate through query refinements. This leads to cascading failures where the agent deviates from the original query's intent and inflates distractor rankings.
    \label{fig:theme}}
    \vspace{-8mm}
\end{figure}

Beyond heterogeneous retrieval biases, interdependencies between needles and distractors introduce another layer of complexity. Many queries are multi-hop, which requires connecting multiple scattered, logically related needle pieces~\citep{yang-etal-2018-hotpotqa}. In networked corpora such as social or hyperlink networks, these pieces are explicitly connected to each other and to potential distractors, yielding the challenge of identifying a smaller ``needle subgraph'' within a larger ``haystack graph''. Graph-based retrieval methods are central to modern information retrieval and search engines~\citep{Page1999ThePC}. Collectively, these factors create a realistic and nuanced interplay that has been largely overlooked in prior NIAH studies (Fig.~\ref{fig:theme} (a)).

Furthermore, advances in reasoning LLMs~\citep{openai2024learning, DeepSeek-R1} enable increasingly LLM-driven agentic context engineering~\citep{yao2023react} for challenging tasks like DeepResearch~\citep{google2025deepresearch}. Rather than passively digest provided contexts and directly jump to the conclusion, LLMs can perform multi-round research~\citep{trivedi-etal-2023-interleaving, jiang-etal-2023-active}, actively refining the query to improve the retrieval quality~\citep{wang-etal-2023-query2doc, ma-etal-2023-query} and reflecting on their own past reasoning~\citep{shinn2023reflexion, asai2024selfrag}. In such dynamic systems with adaptive retrieval and iterative reasoning, LLMs become a distractor source themselves. Early-stage errors---such as noisy retrievals or hallucinated facts---can propagate and compound, leading to cascading failures over iterations or gradual deviation from the original query intent (Fig.~\ref{fig:theme} (b)). As a result, static, single-round, and LLM-independent NIAH evaluations are insufficient, motivating dynamic, multi-round, LLM-dependent test environments.

In order to mitigate the simulation-to-reality gap for LLMs' long-context utility, we argue that \textbf{haystack engineering} is necessary. Just as context engineering seeks to provide optimal LLM contexts, haystack engineering addresses the challenge of constructing realistic noisy long contexts. While context engineering emphasizes best-case conditions, haystack engineering emphasizes faithful haystack constructions shaped by heterogeneous retrieval strategies and cascading agentic errors. 

We instantiate this concept through \textbf{\proj}, a new NIAH benchmark built on the full English Wikipedia hyperlink network and multi-hop questions. \proj systematically examines how retriever choice shapes distractor composition, haystack ordering, and the resulting LLM performance. It evaluates widely adopted retrieval strategies, including sparse, dense, hybrid, and graph-based methods. To contextualize the evaluation in agentic context engineering, we introduce novel LLM-dependent, dynamic NIAH tests. Featuring crucial agentic operations like query refinement and summarization, \proj challenges models in two dynamic long-context settings: (1) an enforced multi-round scenario to measure robustness against cascading errors and (2) a variable-round scenario to examine if models can proactively escape cascading errors by early stop. 

We perform extensive studies covering 15 long-context LLMs, featuring general-purpose, reasoning, open-source, and commercial models. First, we find that retrieval strategy strongly impacts haystack difficulty. While dense retrievers introduce harder distractors than sparse ones, graph-based reranking with Personalized PageRank (PPR) simultaneously improves retrieval effectiveness and mitigates more harmful distractors, improving NIAH performance by up to $44\%$. Second, our dynamic NIAH tests reveal that current models are surprisingly brittle in agentic workflows. Even advanced models like Gemini 2.5 Pro and GPT-5 suffer from cascading self-distraction when multiple reasoning rounds are enforced. Crucially, models tend to be more robust to single-round noisy long contexts (``width'') than to noisy reasoning iterations (``depth''). Even when models are allowed for an early stop, most models fail to terminate the process appropriately. Overall, our evaluations suggest that long-context challenges in realistic, agentic context engineering are far from solved---and that \proj provides a valuable testbed for measuring progress on these issues.
\section{Related Work}

\textbf{Long-Context Benchmarks.} The original NIAH test inserts a single needle sentence into increasingly large haystacks~\citep{kamradt2023needle}. LV-Eval~\citep{yuan2024lveval}, RULER~\citep{hsieh2024ruler}, and BABILong~\citep{kuratov2024babilong} extend the test in terms of question types and corpus sources. However, all these attempts construct query-independent distractors, rather than retriever-dependent contexts, as in practical applications like RAG. HELMET~\citep{yen2025helmet} takes a step toward realism by using a dense retriever for distractor construction, but it does not capture retriever heterogeneity, network-structured corpora, or the influence of retriever-ranked haystack ordering. Beyond NIAH, other benchmarks assess long-context reasoning in downstream tasks and domain-specific applications~\citep{shaham-etal-2022-scrolls,dong2023bamboo, shaham-etal-2023-zeroscrolls, an-etal-2024-l, bai-etal-2024-longbench, bai2024longbench2, wang-etal-2024-leave, zhang-etal-2024-bench, wang2025novelqa}. However, these benchmarks lack the flexibility that NIAH provides in context size and distractor composition. Furthermore, unlike \proj, all existing long-context benchmarks employ static, LLM-independent contexts. This approach is insufficient for evaluating LLMs in dynamic, multi-round agentic systems.

\textbf{Multi-Round Benchmarks.} MT-bench pioneered the evaluation of multi-turn conversations~\citep{zheng2023judging}, with follow-ups refining conversation patterns and question taxonomies~\citep{kwan-etal-2024-mt, duan-etal-2024-botchat, bai-etal-2024-mt, laban2025llms}, extending conversation lengths~\citep{deshpande-etal-2025-multichallenge}, and considering tool usage~\citep{wang2024mint}. These benchmarks purely focus on multi-round conversations rather than agentic context engineering. LLM agent benchmarks like AgentBench~\citep{liu2023agentbench} introduce challenging tasks that necessitate multi-round agentic context engineering, but they do not study how the long-context ``width'' at each step contributes to cascading errors across the reasoning ``depth''. Consequently, none of these works explicitly study the joint ``wide and deep'' long-context challenges that \proj tackle.
\section{\proj}

Context engineering aims to select, structure, and optimize an LLM's input context to maximize its reasoning effectiveness~\citep{NEURIPS2020_6b493230, yao2023react, mei2025survey}. Its practice shapes LLMs' long-context challenges. We introduce the complementary concept of \textbf{haystack engineering}: the principled construction of realistic noisy long contexts that faithfully model the complexities and failure modes of real-world context engineering pipelines. While context engineering seeks to improve performance, haystack engineering aims to create challenging test conditions to measure model robustness, shaped by factors like heterogeneous retrieval strategies and cascading errors in agentic workflows. We present \textbf{\proj}, a benchmark that instantiates this principle.

In this section, we first formalize the NIAH problem arising from RAG, which highlights the central role of the retrieval strategy and motivates studying representative, heterogeneous retrievers. We then introduce, to our knowledge, the first dynamic, LLM-dependent NIAH challenge, designed to characterize long-context challenges in agentic context engineering. Finally, we describe how \proj is grounded in the full English Wikipedia hyperlink network with multi-hop questions, ensuring a realistic and challenging evaluation setting.

\subsection{NIAH Testing for RAG: Retrieval, Needle, and Haystack}

\label{sec:NIAH_RAG}

RAG is a popular context engineering strategy due to its simplicity and broad applicability. In a standard RAG pipeline, a retrieval strategy first fetches the top-$N$ documents deemed most relevant to a query. These documents, along with the query, form the input context for an LLM. To achieve high recall, the hyperparameter $N$ is often set to a value much larger than the number of ground-truth supporting documents for queries. This practice inevitably sacrifices precision and introduces challenging ``near-miss'' distractors that have high retrieval scores~\citep{xu2024retrieval}. The problem is exacerbated when a query requires logically combining information from multiple supporting documents, as in multi-hop question answering (QA). Our empirical studies in Sec.~\ref{sec:retrieval_effectiveness} demonstrate that a larger $N$ is required to achieve comparable retrieval recall for multi-hop questions.

Correspondingly, the requirement for a large $N$ inherently is prone to create the NIAH problem, assuming that perfect retrieval could be achieved with a sufficiently large $N$, or equivalently, a sufficiently large context size. Building on this observation, we formalize the NIAH problem from a RAG perspective:
Let $\gD$ be the document corpus. For any given query $q$, a set of ground-truth documents $\gN_q\subset \gD$ is required to answer it; we refer to this set as the \textbf{needle}. A retrieval strategy $\gR$ scores and ranks all documents in $\gD$ based on their predicted relevance to $q$. 

To construct the \textbf{haystack} $\gH_q^{\gR}(S)$ for a target context size of $S$ tokens, we first include all needle documents from $\gN_q$. We then fill the remaining token budget by adding the top-ranked distractors from $\gD\setminus \gN_q$. If including the final distractor would exceed the budget, we truncate that to fit. Finally, $\gH_q^{\gR}(S)$ is linearized into a sequence of documents $(d_1,\cdots, d_{|\gH_q^{\gR}(S)|})$ according to an ordering policy $\pi(q, \gR, \gH_q^{\gR}(S))$ (e.g., by retrieval ranking), before being passed to the LLM. See Appendix~\ref{sec:NIAH_prompt} for the detailed prompts we use.

\subsection{Assessing Heterogeneous Retrieval Strategies}

\textbf{Retriever Strategy ($\gR$) and Haystack Composition.} The above formulation highlights the central role of retrieval strategy ($\gR$) in haystack engineering. Different retrieval strategies introduce distinct biases into the distractor composition, which consequently shape the reasoning challenge for LLMs according to the strategy's specific failure modes. Since no single method is universally optimal in terms of both effectiveness and efficiency, it is crucial to consider heterogeneous retrievers. To this end, \proj incorporates a broad spectrum of retrievers, including:
\begin{enumerate}[leftmargin=*]
    \item \textbf{Sparse (BM25)}~\citep{conf/trec/RobertsonWJHG94, 10.1561/1500000019}: A classical sparse retriever that measures lexical similarity.
    \item \textbf{Dense (Qwen3-Embedding-0.6B)}~\citep{qwen3embedding}: A dense retriever that captures semantic similarity. We choose this model for its competitive retrieval performance on MMTEB~\citep{enevoldsen2025mmteb}, small size, and applicability to long documents. 
    \item \textbf{Hybrid (BM25 + Qwen3-Embedding-0.6B)}: A combination of the two using reciprocal rank fusion~\citep{10.1145/1571941.1572114, Microsoft2025Hybrid}, which is robust to differences in score magnitudes across retrievers. As sparse and dense retrievers are complementary, a hybrid of them often yields better performance in practice~\citep{lee-etal-2023-complementarity}.
\end{enumerate}

\textbf{Graph-Based Reranking for Multi-Hop QA.} Complex queries, such as those in multi-hop QA~\citep{trivedi-etal-2022-musique}, require synthesizing information from multiple interconnected documents (the needle set $\gN_q$). Standard retrievers score documents independently and thus overlook the relational structure among documents (e.g., hyperlinks or citations), limiting their ability to surface supporting chains. This frames the task as finding a ``needle subgraph'' within a larger ``haystack graph''. Graph structure provides powerful retrieval signals. For instance, PageRank~\citep{Page1999ThePC}, a foundational algorithm for modern search engines, leverages this by considering a document structurally important if it is heavily referenced by other important documents. Building on this idea, we employ Personalized PageRank (PPR)~\citep{10.1145/511446.511513} reranking to study the impact of graph-based retrieval on distractor composition and downstream LLM performance. Specifically, after retrieving an initial candidate set with a base retriever, we perform PPR reranking seeded on the top-ranked documents to integrate structural information.

\textbf{Haystack Ordering ($\pi$)}. LLMs exhibit strong positional biases due to autoregressive generation and positional encodings, and the order of documents can significantly impact their long-context performance~\citep{liu-etal-2024-lost, xiao2024efficient, yang2025ape}. While prior NIAH benchmarks often randomize document order to account for this issue, practical RAG systems present documents in a ranked order determined by the retriever. To bridge this simulation-to-reality gap, we evaluate both retriever-ranked ordering and random permutations. This dual approach allows us to assess LLM performance in a realistic RAG setting while also diagnosing the effects of positional bias.

\subsection{Dynamic NIAH Testing for Agentic Context Engineering}
\label{sec:dynamic_NIAH}

Standard RAG can be ineffective when dealing with imperfect queries or complex tasks. User queries might be ambiguous or contain grammatical errors, which harm effective retrieval. Furthermore, standard RAG struggles with multi-hop queries, which are composed of logically interdependent subqueries. In this case, retrieving enough evidence requires answering earlier subqueries first. For instance, to answer ``\textit{What \colorbox{red!20}{continent} is the \colorbox{orange!20}{country} encompassing Luahoko located in?}'', a system must first find that \colorbox{orange!20}{Luahoko is in Tonga} before a second retrieval for the \colorbox{red!20}{continent of Tonga}.

Agentic context engineering~\citep{yao2023react} can mitigate these limitations by transforming LLMs from passive retrieval consumers into proactive researchers. In such systems, LLMs can dynamically initiate further retrievals as needed~\citep{trivedi-etal-2023-interleaving, jiang-etal-2023-active}, refine queries to optimize retrieval quality (e.g., replacing the query above with ``\textit{\colorbox{blue!20}{What continent is Tonga located in?}}'')~\citep{wang-etal-2023-query2doc, ma-etal-2023-query}, and reflect on their past analyses~\citep{shinn2023reflexion, asai2024selfrag} until they can confidently draw a conclusion.

However, agentic context engineering introduces a new challenge: \textbf{LLMs themselves become a potential source of distraction}.
Recent studies show that even advanced reasoning models struggle to recognize their own reasoning errors, often \emph{reinforcing initial mistakes} rather than \emph{correcting them}~\citep{huang2024large, he-etal-2025-large}.
Early-stage errors, such as noisy retrievals or flawed reasoning, can propagate and compound through LLMs' generation, leading to cascading failures or a gradual deviation from the original query's intent.
While related issues have been previously observed~\citep{laban2025llms}, the interplay between wider context windows and deeper agentic iterations introduces a critical failure mode not captured by existing static NIAH benchmarks and multi-round benchmarks.
This gap highlights a need for benchmarks that can test these integrated ``wide and deep'' long-context challenges.
Existing static, single-round NIAH tests are insufficient, motivating our development of a dynamic, multi-round, and LLM-dependent test environment.

We perform an extension of our previous NIAH formulation in Section~\ref{sec:NIAH_RAG} for comparable results, simplicity, and controllability, while capturing key characteristics: multi-round retrieval, query refinement, and self-reflection. The process is iterative. We start with the original query $q^{(0)}=q$ and an empty LLM reasoning history $\mathcal{C}^{(0)}=()$. At each round $t$, we use the latest query $q^{(t)}$ to construct the haystack $\mathcal{H}_{q^{(t)}}^\mathcal{R}(S)$. The LLM receives $q^{(t)}$, the history $\mathcal{C}^{(t)}$, and the ordered haystack. In intermediate rounds, the LLM outputs a refined query $q^{(t+1)}$ and its latest analysis $\mathcal{A}^{(t+1)}$, and we update the history $\mathcal{C}^{(t+1)}=(\mathcal{A}^{(1)}, \cdots, \mathcal{A}^{(t+1)})$. In the final round, the LLM outputs its answer. 

We evaluate models in two dynamic settings, with detailed prompts provided in Appendix~\ref{sec:dynamic_NIAH_prompts}: 
\begin{itemize}[leftmargin=*]
    \item \textbf{Enforced Multi-Round}. We enforce models to perform a \textbf{fixed} number of reasoning rounds to measure their robustness against cascading errors.
    \item \textbf{Variable-Round}. We allow models to \textbf{decide when to stop}, testing their ability to balance iterative refinement against the risk of cascading errors.
\end{itemize}

\subsection{Corpus and QA Samples}
\label{sec:corpus_QA_samples}

To instantiate the static and dynamic NIAH tests introduced above, we need a networked corpus and QA dataset that support heterogeneous retrieval strategies and multi-hop reasoning. In this subsection, we detail our choice of corpus and QA samples that satisfy the need.

\textbf{Networked Corpus.} We ground our benchmark in the entire English Wikipedia hyperlink network. This choice is deliberate: Wikipedia is a dominant information source for retriever development and QA dataset curation and serves as a widely recognized proxy for general knowledge~\citep{chen-etal-2017-reading}. It provides a centralized testbed for studying realistic haystack engineering with diverse retrievers. Furthermore, its large scale and natural network structure, formed by in-text references (hyperlinks), make it an ideal corpus for studying graph-based retrieval. We process the 2025-04-04 Wikipedia dump using WikiExtractor~\citep{Wikiextractor2015}, resulting in a network that comprises $6, 954, 909$ articles interconnected by $97, 442, 472$ unique hyperlinks. 

\textbf{Long Retrieval Unit.} We choose to use full articles as the retrieval unit, rather than smaller, broken chunks. This approach mirrors modern search engines, which return entire documents, and avoids fragmenting a document's logical flow as with common chunking practices in RAG. By preserving article integrity, we present a more realistic and demanding long-context reasoning challenge.

\textbf{QA Datasets.} We use two established datasets: Natural Questions (NQ)~\citep{47761} for single-hop questions and MuSiQue~\citep{trivedi-etal-2022-musique} for multi-hop questions. The multi-hop questions require reasoning over up to four supporting documents, presenting a challenge that motivates agentic context engineering. Both datasets are built on Wikipedia, providing a unified source for needles and distractors. We choose MuSiQue over alternatives~\citep{yang-etal-2018-hotpotqa, ho-etal-2020-constructing} as it is specifically designed to be less susceptible to reasoning shortcuts. Since both datasets were curated on earlier Wikipedia versions, we manually filter all samples to ensure validity under our updated corpus, addressing issues like outdated knowledge and ambiguity. This yields a final set of 500 high-quality samples. Further details are available in Appendix~\ref{appendix:dataset_details}.

\textbf{Data Contamination Mitigation.} A critical concern in LLM evaluation is data contamination, where exposure to benchmark data during pretraining inflates performance~\citep{sainz-etal-2023-nlp}. While the models we evaluate have almost certainly been trained on versions of Wikipedia and even the QA datasets, our benchmark's design inherently mitigates this risk. The core task demands locating the ``needle'' within a long context of plausible, retriever-selected distractors—rather than simple fact recall. This challenge is amplified for our multi-hop questions, which require synthesizing information across multiple documents, a process robust to memorization. Furthermore, our use of a recent Wikipedia dump post-dates the training cutoffs of most current LLMs, minimizing data overlap. Our empirical results in Section~\ref{sec: emp_NIAH_perofrmance} confirm this mitigation: all models show substantial performance degradation as context size increases, demonstrating that they are actively reasoning over the provided text, not merely recalling memorized answers.

\section{Experiments}

\subsection{NIAH with Heterogeneous Retrieval Strategies}

To investigate the impact of retrieval strategies, we evaluate $12$ open-source and commercial long-context LLMs across input context sizes of $S\in\{8K, 16K, 32K, 64K, 128K\}$. Our selection spans both reasoning models---three Qwen3 variants~\citep{qwen3}, Gemini 2.5 Flash-Lite, and o4-mini---and leading general-purpose models, including GPT-4.1 mini and the open-source Llama-3.1~\citep{dubey2024llama3herdmodels}, Qwen2.5-1M~\citep{qwen2.5-1M}, and Gemma 3~\citep{gemmateam2025gemma3} families. Following the practice of MuSiQue, we use the F1 score as the QA metric. Our analysis in Appendix~\ref{appendix:contamination} confirms that multi-hop questions are less susceptible to data contamination, making them preferable for our study. To ensure a fair comparison, we standardize token counts using the Qwen2.5-1M tokenizer. For more LLM and retriever setup details, see Appendix~\ref{appendix:setup_details}.

\begin{figure}[t]
    \centering
    \includegraphics[width=\linewidth]{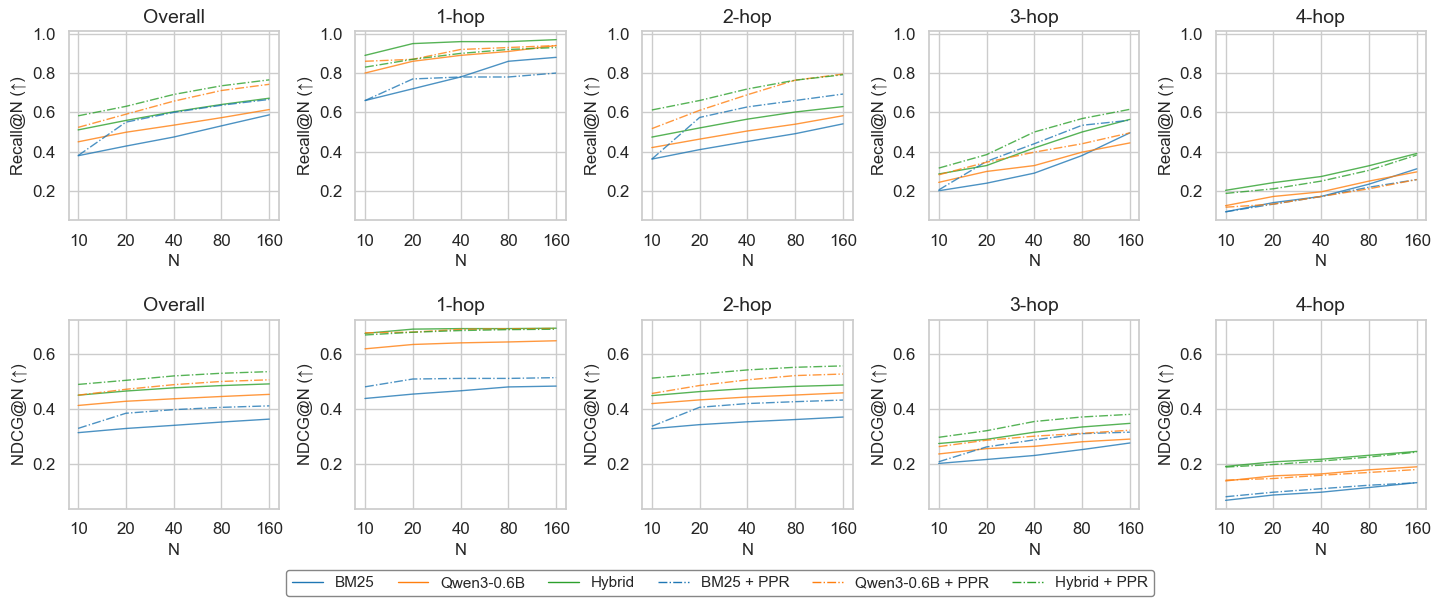}
    \caption{(1) Retrieval performance improves as \# retrieved documents $(N)$ increases. (2) Multi-hop questions pose larger retrieval challenges. (3) Reranking with PPR consistently boosts performance, especially for multi-hop questions. See Appendix~\ref{sec:detailed_retrieval_eval} for the raw numbers.
    \label{fig:ndcg_recall}}
    \vspace{-6mm}
\end{figure}

\label{sec:retrieval_effectiveness}

\textbf{Retrieval Effectiveness.} As a preliminary step, we first evaluate the effectiveness of the retrieval strategies to ensure they construct meaningful distractors. We measure both Recall $@N$, which quantifies the coverage of ground-truth supporting documents, and NDCG $@N$~\citep{10.1145/345508.345545, 10.1145/582415.582418}, which additionally accounts for the ranking. We study the scaling behavior of retrieval by increasing $N$, the number of retrieved documents, which directly corresponds to increasing the context size in our NIAH setting.

Fig.~\ref{fig:ndcg_recall} shows that retrieval performance for all methods improves as $N$ increases, justifying constructing longer contexts with more distractors. Among the base retrievers, the dense retriever (Qwen3-0.6B) consistently outperforms the sparse retriever (BM25) in both metrics, and combining them with a hybrid retriever further improves the performance. The retrieval effectiveness decreases as the question hop count (\# supporting documents) increases, validating our claim in Sec.~\ref{sec:NIAH_RAG} that multi-hop questions necessitate a larger $N$, leading to more distractors. Graph-based reranking with PPR boosts all base retrievers in both coverage and ranking, especially for multi-hop questions.

\label{sec: emp_NIAH_perofrmance}

\begin{figure}[t]
    \centering
    \includegraphics[width=\linewidth]{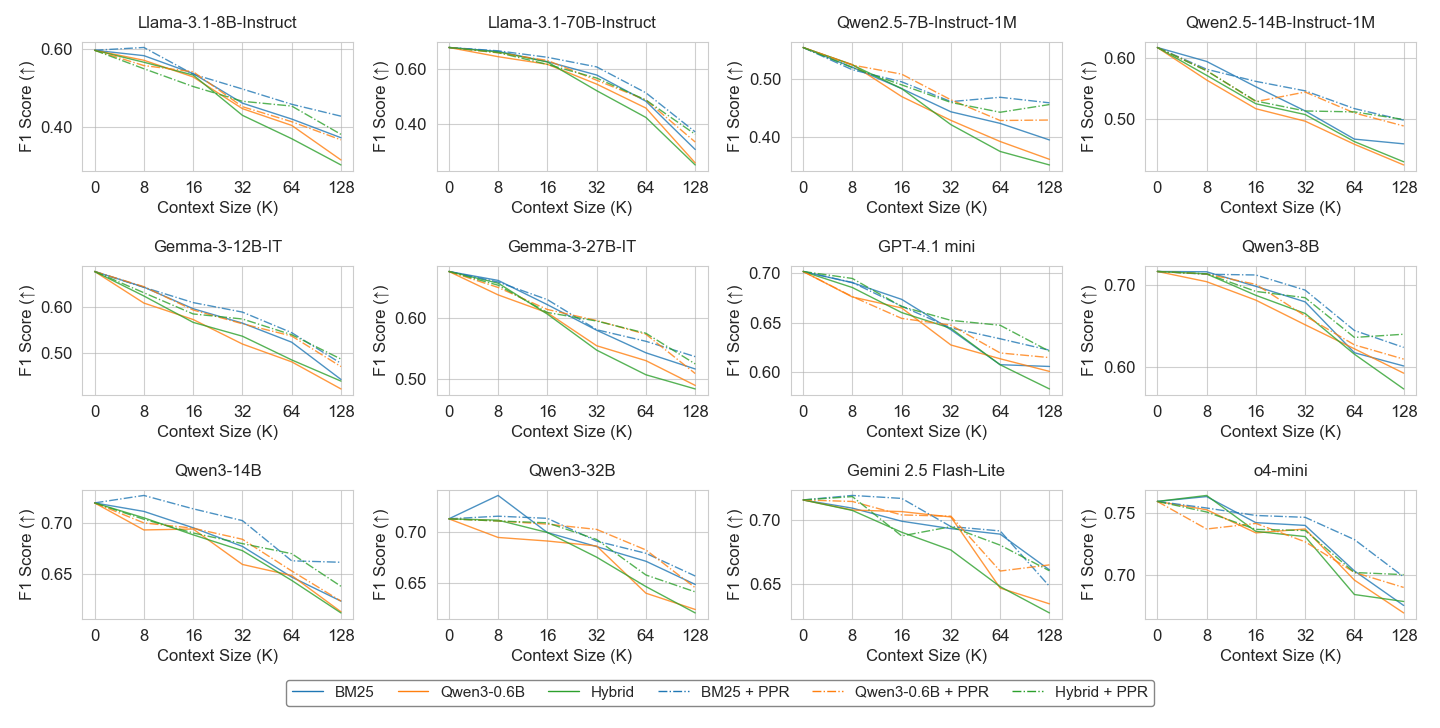}
    \caption{Impact of retrieval strategy on NIAH performance as context size increases. $0$ stands for the case without distractors. All models experience a performance drop as context size increases. Graph-based reranking (dashed lines) consistently improves performance for larger context sizes. See Appendix~\ref{appendix:static_NIAH_rank_order} for the raw numbers.
    \label{fig:retriever_choice}}
\end{figure}

\textbf{Impact of Retriever Strategy on NIAH Performance.} To study the overall impact of the retrieval strategy ($\mathcal{R}$) on haystack composition ($\gH_q^{\gR}(S)$) and ordering, we first employ retrieval ranking for haystack ordering ($\pi$). Fig.~\ref{fig:retriever_choice} presents the evaluation results. All LLMs, including advanced commercial and reasoning models, suffer significant performance degradation as context size increases to $128K$ tokens, regardless of the retrieval strategy. For 11 out of 12 cases, the dense retriever (Qwen3-0.6B) introduces more challenging distractors than the sparse retriever (BM25) at larger context sizes. However, combining them with a hybrid retriever does not necessarily introduce more challenging distractors. 

\textbf{Impact of Graph-Based Retrieval.} Using PPR for graph-based reranking leads to significant NIAH performance improvement. By comparing the solid lines with the dashed lines in Fig.~\ref{fig:retriever_choice}, we observe that for nearly every model and base retriever, the performance curve paired with PPR is noticeably higher, especially at context sizes of $64K$ and $128K$. This demonstrates that exploiting the relational structure among documents is a powerful method for mitigating distraction. For instance, an improvement of $44\%$ was observed for Llama-3.1-70B-Instruct with the hybrid retriever, highlighting how prioritizing structurally central documents can mitigate more harmful structurally isolated lexical and semantic distractors.

\textbf{Retrieval Effectiveness vs NIAH Performance.} Previous study by~\citet{jin2025longcontext} suggests that better retrievers introduce harder distractors for shorter-context reasoning and single-hop QA. Our study discloses a deeper insight, demonstrating that the interplay between the retriever mechanism and task nature plays a crucial role. While hybrid retriever substantially improves retrieval recall and ranking, it fails to introduce more challenging distractors. In contrast, graph-based reranking simultaneously improves retrieval effectiveness and mitigates harmful distractors. Our study highlights the critical role of retrieval strategy design in long-context engineering.

\begin{figure}[t]
    \centering
    \includegraphics[width=\linewidth]{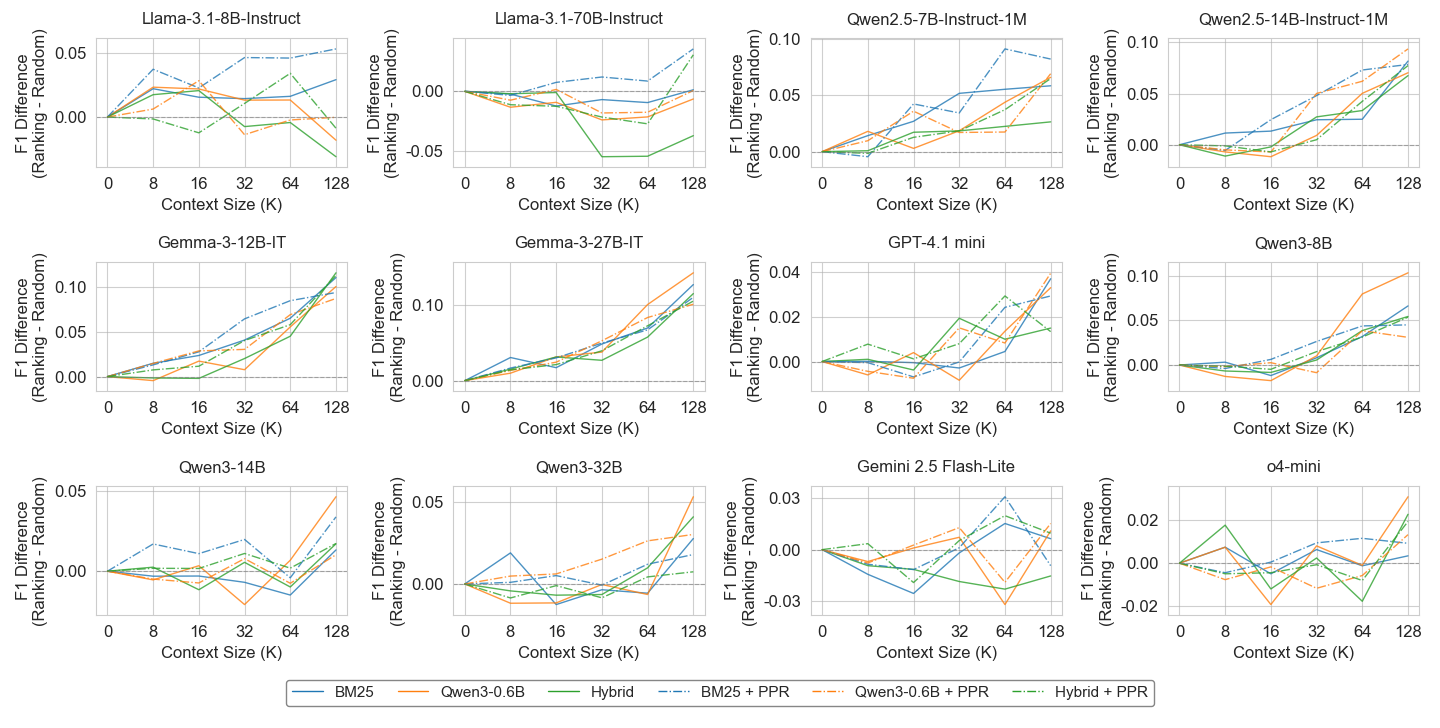}
    \caption{F1 score difference between retriever-ranked and random haystack orderings. The ordering impact is highly model-dependent. The Gemma-3 and Qwen2.5-1M families derive a significant and growing benefit from retriever-ranked ordering as context size expands. See Appendix~\ref{appendix:static_NIAH_random} for the raw NIAH performance numbers with random haystack orderings.
    \label{fig:order_diff}
    }
    \vspace{-3mm}
\end{figure}

\textbf{Impact of Haystack Ordering.} To isolate the effect of haystack ordering $(\pi)$, we compare the performance of retriever-ranked ordering against the average of three random permutations. The results in Fig.~\ref{fig:order_diff} reveal complex and highly model-dependent patterns. While some models, such as the Gemma-3 and Qwen2.5-1M families, derive a significant and growing benefit from retriever-ranked ordering as context size expands, others exhibit a more volatile, retriever-dependent, or even negative response. This finding carries a crucial implication: to faithfully assess a model's practical long-context utility in RAG systems, evaluations must mirror the canonical, retriever-ranked input. Furthermore, contrasting this setup with random permutations allows us to better understand the positional biases of individual models. 

\subsection{Dynamic NIAH}

To assess the ``wide and deep'' challenges in agentic context engineering (Sec.~\ref{sec:dynamic_NIAH}), we perform dynamic NIAH evaluation with multi-round reasoning. We randomly choose 100 QA samples and evaluate eight LLMs, including state-of-the-art models Gemini 2.5 Pro and GPT-5. We exclusively use retriever-ranked haystack ordering, as this realistic setup ensures that LLM's query refinement is directly reflected in the context. Flawed refinements degrade the document ranking, posing a dynamic challenge that faithfully evaluates a model’s robustness to cascading errors. We consider two representative retrieval strategies: BM25 + PPR, a graph-based strategy effective at mitigating more harmful distractors, and Qwen3-0.6B, a dense retriever that introduces more challenging distractors.

 \begin{figure}[t]
     \centering

     \begin{subfigure}{\linewidth} 
         \centering
         \includegraphics[width=\linewidth]{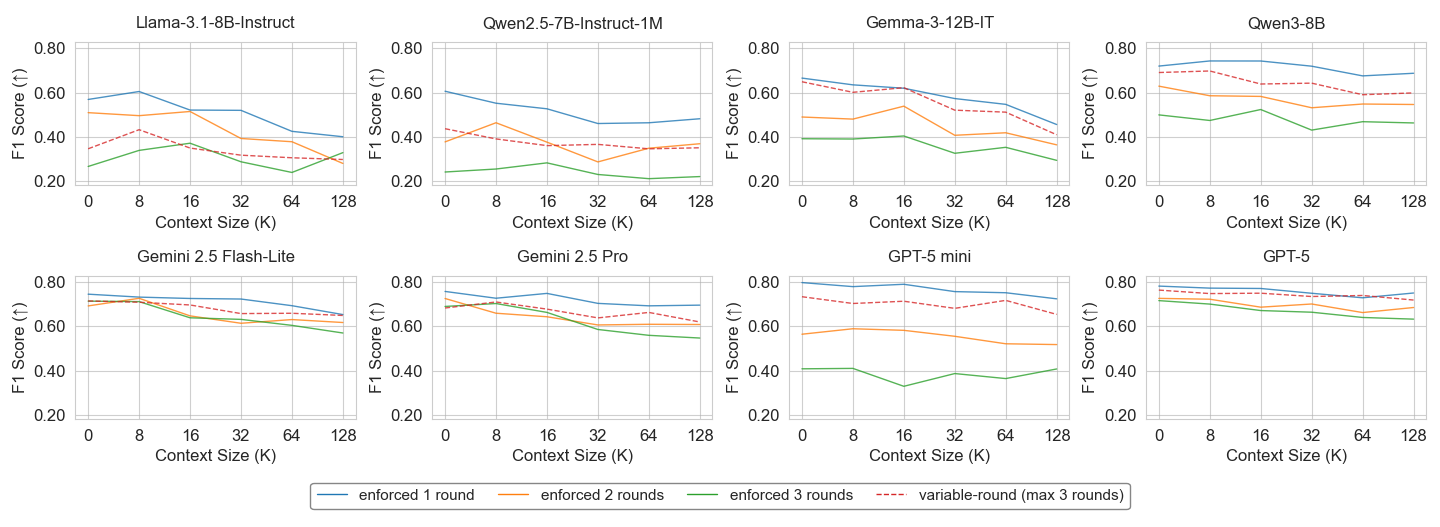}
         \caption{Haystack construction with BM25 + PPR.}
         \label{fig:dynamic_NIAH_bm25_ppr}
     \end{subfigure}

     \begin{subfigure}{\linewidth} 
         \centering
         \includegraphics[width=\linewidth]{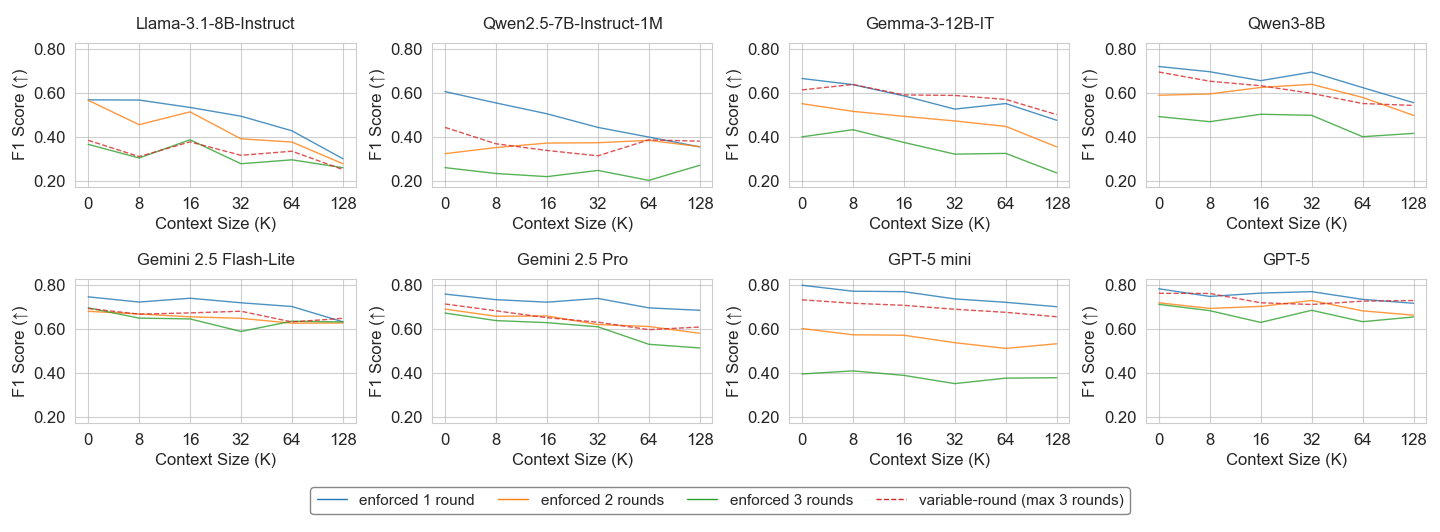}
         \caption{Haystack construction with Qwen3-0.6B.}
         \label{fig:dynamic_NIAH_qwen}
     \end{subfigure}

     \caption{Dynamic NIAH performance. $0$ stands for the case without distractors. (1) Enforced multi-round reasoning leads to performance drop. (2) Models are generally more robust to wider contexts than deeper reasoning. (3) Models fail to perform early stop properly (variable-round). For raw experiment numbers, see Appendix~\ref{appendix:dynamic_NIAH_raw}.}
     \vspace{-6mm}
     \label{fig:dynamic_NIAH}
\end{figure}

\textbf{Enforced Multi-Round: More Rounds Amplify Errors.} We first evaluate model robustness by enforcing a constant number (2 or 3) of reasoning rounds. Fig.~\ref{fig:dynamic_NIAH} shows that all models, including the most advanced Gemini 2.5 Pro and GPT-5, are vulnerable to cascading errors. Across retrieval strategies and context sizes, performance generally worsens with more rounds. Rather than mitigating distraction, additional iterations often amplify early mistakes or inject new noise. Interestingly, the degradation is not always monotonic with context size in multi-round settings, and more rounds can be more damaging than longer noisy contexts in a single pass. Crucially, static NIAH performance is not a reliable predictor of multi-round robustness: for instance, GPT-5 mini performs comparably to GPT-5 in the static setting but collapses under enforced multi-round reasoning, revealing weaker agentic robustness.

\textbf{Variable-Round: Self-Correction Is Difficult.} We further investigate if models can balance iterative refinement against cascading errors when allowed to stop early before exhausting three rounds. None of the models reliably improve upon their single-round performance. GPT-5 achieves the best relative performance but still fails to convert multi-round reasoning into sustained improvements.

\textbf{Representative Failure Patterns.} Appendix~\ref{appendix:frequent_cascading_failure_patterns} presents representative failure cases. 1) Early-stage errors can propagate and compound through query refinement and summarization, leading to cascading failures that are hard to correct. 2) LLMs can deviate from the original query intent, changing its nature or form. 3) Long-context challenges can still prevent LLMs from reasoning and retrieving relevant information.

\textbf{Implications.} These findings reveal a consistent pattern: current LLMs are more robust to noisy long contexts than noisy reasoning iterations. In practice, practitioners should prioritize the use of a larger context window size (``width'') over more reasoning iterations (``depth''). These results underscore the unsolved long-context challenges in agentic context engineering and establish \proj as a valuable testbed for measuring and advancing agentic robustness.

\section{Conclusion}

Our benchmark, \proj, demonstrates that retrieval strategies critically shape distractor composition and ordering. Furthermore, our novel dynamic tests reveal that even state-of-the-art models like Gemini 2.5 Pro and GPT-5 remain vulnerable to cascading self-distractions and fail to self-correct. These findings highlight that robust agentic long-context reasoning is far from solved and establish HaystackCraft as a valuable testbed for measuring future progress.


\subsubsection*{Acknowledgments}
M. Li, H. Wang, and P. Li are partially supported by the NSF under awards IIS-2239565, IIS-2428777, and CCF-2402816; the JP-Morgan Chase Faculty Award; the OpenAI Researcher Access Program Credit; the Google Gemini Academic Program; and the IDEaS Cyberinfrastructure Awards. XB is supported by NUS Grant ID R-252-000-B97-133 and MOE AcRF T1 Grant ID 251RES2423. The authors would also like to thank Xi Wang for valuable feedback on early drafts. 

\bibliography{iclr2026_conference}

\begin{thebibliography}{72}
\providecommand{\natexlab}[1]{#1}
\providecommand{\url}[1]{\texttt{#1}}
\expandafter\ifx\csname urlstyle\endcsname\relax
  \providecommand{\doi}[1]{doi: #1}\else
  \providecommand{\doi}{doi: \begingroup \urlstyle{rm}\Url}\fi

\bibitem[An et~al.(2024)An, Gong, Zhong, Zhao, Li, Zhang, Kong, and Qiu]{an-etal-2024-l}
Chenxin An, Shansan Gong, Ming Zhong, Xingjian Zhao, Mukai Li, Jun Zhang, Lingpeng Kong, and Xipeng Qiu.
\newblock {{L}-Eval: Instituting Standardized Evaluation for Long Context Language Models}.
\newblock In \emph{Proceedings of the 62nd Annual Meeting of the Association for Computational Linguistics (Volume 1: Long Papers)}, pp.\  14388--14411, 2024.

\bibitem[Asai et~al.(2024)Asai, Wu, Wang, Sil, and Hajishirzi]{asai2024selfrag}
Akari Asai, Zeqiu Wu, Yizhong Wang, Avirup Sil, and Hannaneh Hajishirzi.
\newblock {Self-{RAG}: Learning to Retrieve, Generate, and Critique through Self-Reflection}.
\newblock In \emph{International Conference on Learning Representations}, 2024.

\bibitem[Attardi(2015)]{Wikiextractor2015}
Giusepppe Attardi.
\newblock {WikiExtractor}.
\newblock \url{https://github.com/attardi/wikiextractor}, 2015.

\bibitem[Bai et~al.(2024{\natexlab{a}})Bai, Liu, Bu, He, Liu, Zhou, Lin, Su, Ge, Zheng, and Ouyang]{bai-etal-2024-mt}
Ge~Bai, Jie Liu, Xingyuan Bu, Yancheng He, Jiaheng Liu, Zhanhui Zhou, Zhuoran Lin, Wenbo Su, Tiezheng Ge, Bo~Zheng, and Wanli Ouyang.
\newblock {{MT}-Bench-101: A Fine-Grained Benchmark for Evaluating Large Language Models in Multi-Turn Dialogues}.
\newblock In \emph{Proceedings of the 62nd Annual Meeting of the Association for Computational Linguistics (Volume 1: Long Papers)}, pp.\  7421--7454, 2024{\natexlab{a}}.

\bibitem[Bai et~al.(2024{\natexlab{b}})Bai, Lv, Zhang, Lyu, Tang, Huang, Du, Liu, Zeng, Hou, Dong, Tang, and Li]{bai-etal-2024-longbench}
Yushi Bai, Xin Lv, Jiajie Zhang, Hongchang Lyu, Jiankai Tang, Zhidian Huang, Zhengxiao Du, Xiao Liu, Aohan Zeng, Lei Hou, Yuxiao Dong, Jie Tang, and Juanzi Li.
\newblock {{L}ong{B}ench: A Bilingual, Multitask Benchmark for Long Context Understanding}.
\newblock In \emph{Proceedings of the 62nd Annual Meeting of the Association for Computational Linguistics (Volume 1: Long Papers)}, pp.\  3119--3137, 2024{\natexlab{b}}.

\bibitem[Bai et~al.(2024{\natexlab{c}})Bai, Tu, Zhang, Peng, Wang, Lv, Cao, Xu, Hou, Dong, Tang, and Li]{bai2024longbench2}
Yushi Bai, Shangqing Tu, Jiajie Zhang, Hao Peng, Xiaozhi Wang, Xin Lv, Shulin Cao, Jiazheng Xu, Lei Hou, Yuxiao Dong, Jie Tang, and Juanzi Li.
\newblock {LongBench v2: Towards Deeper Understanding and Reasoning on Realistic Long-context Multitasks}.
\newblock \emph{arXiv preprint arXiv:2412.15204}, 2024{\natexlab{c}}.

\bibitem[Chen et~al.(2017)Chen, Fisch, Weston, and Bordes]{chen-etal-2017-reading}
Danqi Chen, Adam Fisch, Jason Weston, and Antoine Bordes.
\newblock {Reading {W}ikipedia to Answer Open-Domain Questions}.
\newblock In \emph{Proceedings of the 55th Annual Meeting of the Association for Computational Linguistics (Volume 1: Long Papers)}, pp.\  1870--1879, 2017.

\bibitem[Cormack et~al.(2009)Cormack, Clarke, and Buettcher]{10.1145/1571941.1572114}
Gordon~V. Cormack, Charles L~A Clarke, and Stefan Buettcher.
\newblock {Reciprocal rank fusion outperforms condorcet and individual rank learning methods}.
\newblock In \emph{Proceedings of the 32nd International ACM SIGIR Conference on Research and Development in Information Retrieval}, pp.\  758–759, 2009.

\bibitem[Dao(2024)]{dao2023flashattention2}
Tri Dao.
\newblock {Flash{A}ttention-2: Faster Attention with Better Parallelism and Work Partitioning}.
\newblock In \emph{International Conference on Learning Representations}, 2024.

\bibitem[Dao et~al.(2022)Dao, Fu, Ermon, Rudra, and R{\'e}]{dao2022flashattention}
Tri Dao, Daniel~Y. Fu, Stefano Ermon, Atri Rudra, and Christopher R{\'e}.
\newblock {Flash{A}ttention: Fast and Memory-Efficient Exact Attention with {IO}-Awareness}.
\newblock In \emph{Advances in Neural Information Processing Systems}, 2022.

\bibitem[DeepSeek-AI et~al.(2025)]{DeepSeek-R1}
DeepSeek-AI et~al.
\newblock {DeepSeek-R1: Incentivizing Reasoning Capability in LLMs via Reinforcement Learning}.
\newblock \emph{arXiv preprint arXiv:2501.12948}, 2025.

\bibitem[Deshpande et~al.(2025)Deshpande, Sirdeshmukh, Mols, Jin, Hernandez-Cardona, Lee, Kritz, Primack, Yue, and Xing]{deshpande-etal-2025-multichallenge}
Kaustubh Deshpande, Ved Sirdeshmukh, Johannes~Baptist Mols, Lifeng Jin, Ed-Yeremai Hernandez-Cardona, Dean Lee, Jeremy Kritz, Willow~E. Primack, Summer Yue, and Chen Xing.
\newblock {{M}ulti{C}hallenge: A Realistic Multi-Turn Conversation Evaluation Benchmark Challenging to Frontier {LLM}s}.
\newblock In \emph{Findings of the Association for Computational Linguistics: ACL 2025}, pp.\  18632--18702, 2025.

\bibitem[Dong et~al.(2023)Dong, Tang, Li, Zhao, and Wen]{dong2023bamboo}
Zican Dong, Tianyi Tang, Junyi Li, Wayne~Xin Zhao, and Ji-Rong Wen.
\newblock {BAMBOO: A Comprehensive Benchmark for Evaluating Long Text Modeling Capacities of Large Language Models}.
\newblock \emph{arXiv preprint arXiv:2309.13345}, 2023.

\bibitem[Duan et~al.(2024)Duan, Wei, Wang, Liu, Fang, Zhang, Lin, and Chen]{duan-etal-2024-botchat}
Haodong Duan, Jueqi Wei, Chonghua Wang, Hongwei Liu, Yixiao Fang, Songyang Zhang, Dahua Lin, and Kai Chen.
\newblock {{B}ot{C}hat: Evaluating {LLM}s' Capabilities of Having Multi-Turn Dialogues}.
\newblock In \emph{Findings of the Association for Computational Linguistics: NAACL 2024}, pp.\  3184--3200, 2024.

\bibitem[Dubey et~al.(2024)]{dubey2024llama3herdmodels}
Abhimanyu Dubey et~al.
\newblock The llama 3 herd of models.
\newblock \emph{arXiv preprint arXiv:2407.21783}, 2024.

\bibitem[Enevoldsen et~al.(2025)]{enevoldsen2025mmteb}
Kenneth Enevoldsen et~al.
\newblock {{MMTEB}: Massive Multilingual Text Embedding Benchmark}.
\newblock In \emph{International Conference on Learning Representations}, 2025.

\bibitem[{Google}(2025)]{google2025deepresearch}
{Google}.
\newblock {Gemini Deep Research}.
\newblock \url{https://gemini.google/overview/deep-research/}, 2025.
\newblock Accessed: 2025-09-14.

\bibitem[Haveliwala(2002)]{10.1145/511446.511513}
Taher~H. Haveliwala.
\newblock {Topic-sensitive PageRank}.
\newblock In \emph{Proceedings of the 11th International Conference on World Wide Web}, pp.\  517–526, 2002.

\bibitem[He et~al.(2025)He, Li, Liu, Wang, Bu, Zhang, Peng, Zhang, Zheng, Su, and Zheng]{he-etal-2025-large}
Yancheng He, Shilong Li, Jiaheng Liu, Weixun Wang, Xingyuan Bu, Ge~Zhang, Z.y. Peng, Zhaoxiang Zhang, Zhicheng Zheng, Wenbo Su, and Bo~Zheng.
\newblock {Can Large Language Models Detect Errors in Long Chain-of-Thought Reasoning?}
\newblock In \emph{Proceedings of the 63rd Annual Meeting of the Association for Computational Linguistics (Volume 1: Long Papers)}, pp.\  18468--18489, 2025.

\bibitem[Ho et~al.(2020)Ho, Duong~Nguyen, Sugawara, and Aizawa]{ho-etal-2020-constructing}
Xanh Ho, Anh-Khoa Duong~Nguyen, Saku Sugawara, and Akiko Aizawa.
\newblock {Constructing A Multi-hop {QA} Dataset for Comprehensive Evaluation of Reasoning Steps}.
\newblock In \emph{Proceedings of the 28th International Conference on Computational Linguistics}, pp.\  6609--6625, 2020.

\bibitem[Hsieh et~al.(2024)Hsieh, Sun, Kriman, Acharya, Rekesh, Jia, and Ginsburg]{hsieh2024ruler}
Cheng-Ping Hsieh, Simeng Sun, Samuel Kriman, Shantanu Acharya, Dima Rekesh, Fei Jia, and Boris Ginsburg.
\newblock {{RULER}: What{\textquoteright}s the Real Context Size of Your Long-Context Language Models?}
\newblock In \emph{Conference on Language Modeling}, 2024.

\bibitem[Huang et~al.(2024)Huang, Chen, Mishra, Zheng, Yu, Song, and Zhou]{huang2024large}
Jie Huang, Xinyun Chen, Swaroop Mishra, Huaixiu~Steven Zheng, Adams~Wei Yu, Xinying Song, and Denny Zhou.
\newblock {Large Language Models Cannot Self-Correct Reasoning Yet}.
\newblock In \emph{International Conference on Learning Representations}, 2024.

\bibitem[J\"{a}rvelin \& Kek\"{a}l\"{a}inen(2000)J\"{a}rvelin and Kek\"{a}l\"{a}inen]{10.1145/345508.345545}
Kalervo J\"{a}rvelin and Jaana Kek\"{a}l\"{a}inen.
\newblock {IR evaluation methods for retrieving highly relevant documents}.
\newblock In \emph{Proceedings of the 23rd Annual International ACM SIGIR Conference on Research and Development in Information Retrieval}, pp.\  41–48, 2000.

\bibitem[J\"{a}rvelin \& Kek\"{a}l\"{a}inen(2002)J\"{a}rvelin and Kek\"{a}l\"{a}inen]{10.1145/582415.582418}
Kalervo J\"{a}rvelin and Jaana Kek\"{a}l\"{a}inen.
\newblock {Cumulated gain-based evaluation of IR techniques}.
\newblock \emph{ACM Trans. Inf. Syst.}, 20\penalty0 (4):\penalty0 422–446, 2002.

\bibitem[Jiang et~al.(2023)Jiang, Xu, Gao, Sun, Liu, Dwivedi-Yu, Yang, Callan, and Neubig]{jiang-etal-2023-active}
Zhengbao Jiang, Frank Xu, Luyu Gao, Zhiqing Sun, Qian Liu, Jane Dwivedi-Yu, Yiming Yang, Jamie Callan, and Graham Neubig.
\newblock {Active Retrieval Augmented Generation}.
\newblock In \emph{Proceedings of the 2023 Conference on Empirical Methods in Natural Language Processing}, pp.\  7969--7992, 2023.

\bibitem[Jin et~al.(2025)Jin, Yoon, Han, and Arik]{jin2025longcontext}
Bowen Jin, Jinsung Yoon, Jiawei Han, and Sercan~O Arik.
\newblock {Long-Context {LLM}s Meet {RAG}: Overcoming Challenges for Long Inputs in {RAG}}.
\newblock In \emph{International Conference on Learning Representations}, 2025.

\bibitem[Kamath et~al.(2025)]{gemmateam2025gemma3}
Aishwarya Kamath et~al.
\newblock {Gemma 3 Technical Report}.
\newblock \emph{arXiv preprint arXiv:2503.19786}, 2025.

\bibitem[Kamradt(2023)]{kamradt2023needle}
Gregory Kamradt.
\newblock {Needle In A Haystack - Pressure Testing LLMs}.
\newblock \url{https://github.com/gkamradt/LLMTest_NeedleInAHaystack}, 2023.
\newblock Accessed: Apr. 15, 2025.

\bibitem[Karpukhin et~al.(2020)Karpukhin, Oguz, Min, Lewis, Wu, Edunov, Chen, and Yih]{karpukhin-etal-2020-dense}
Vladimir Karpukhin, Barlas Oguz, Sewon Min, Patrick Lewis, Ledell Wu, Sergey Edunov, Danqi Chen, and Wen-tau Yih.
\newblock {Dense Passage Retrieval for Open-Domain Question Answering}.
\newblock In \emph{Proceedings of the 2020 Conference on Empirical Methods in Natural Language Processing (EMNLP)}, pp.\  6769--6781, 2020.

\bibitem[Kuratov et~al.(2024)Kuratov, Bulatov, Anokhin, Rodkin, Sorokin, Sorokin, and Burtsev]{kuratov2024babilong}
Yuri Kuratov, Aydar Bulatov, Petr Anokhin, Ivan Rodkin, Dmitry~Igorevich Sorokin, Artyom Sorokin, and Mikhail Burtsev.
\newblock {{BABIL}ong: Testing the Limits of {LLM}s with Long Context Reasoning-in-a-Haystack}.
\newblock In \emph{Advances in Neural Information Processing Systems Datasets and Benchmarks Track}, 2024.

\bibitem[Kwan et~al.(2024)Kwan, Zeng, Jiang, Wang, Li, Shang, Jiang, Liu, and Wong]{kwan-etal-2024-mt}
Wai-Chung Kwan, Xingshan Zeng, Yuxin Jiang, Yufei Wang, Liangyou Li, Lifeng Shang, Xin Jiang, Qun Liu, and Kam-Fai Wong.
\newblock {{MT}-Eval: A Multi-Turn Capabilities Evaluation Benchmark for Large Language Models}.
\newblock In \emph{Proceedings of the 2024 Conference on Empirical Methods in Natural Language Processing}, pp.\  20153--20177, 2024.

\bibitem[Kwiatkowski et~al.(2019)Kwiatkowski, Palomaki, Redfield, Collins, Parikh, Alberti, Epstein, Polosukhin, Kelcey, Devlin, Lee, Toutanova, Jones, Chang, Dai, Uszkoreit, Le, and Petrov]{47761}
Tom Kwiatkowski, Jennimaria Palomaki, Olivia Redfield, Michael Collins, Ankur Parikh, Chris Alberti, Danielle Epstein, Illia Polosukhin, Matthew Kelcey, Jacob Devlin, Kenton Lee, Kristina~N. Toutanova, Llion Jones, Ming-Wei Chang, Andrew Dai, Jakob Uszkoreit, Quoc Le, and Slav Petrov.
\newblock {Natural Questions: a Benchmark for Question Answering Research}.
\newblock \emph{Transactions of the Association of Computational Linguistics}, 2019.

\bibitem[Kwon et~al.(2023)Kwon, Li, Zhuang, Sheng, Zheng, Yu, Gonzalez, Zhang, and Stoica]{kwon2023efficient}
Woosuk Kwon, Zhuohan Li, Siyuan Zhuang, Ying Sheng, Lianmin Zheng, Cody~Hao Yu, Joseph~E. Gonzalez, Hao Zhang, and Ion Stoica.
\newblock {Efficient Memory Management for Large Language Model Serving with PagedAttention}.
\newblock In \emph{Proceedings of the ACM SIGOPS 29th Symposium on Operating Systems Principles}, 2023.

\bibitem[Laban et~al.(2025)Laban, Hayashi, Zhou, and Neville]{laban2025llms}
Philippe Laban, Hiroaki Hayashi, Yingbo Zhou, and Jennifer Neville.
\newblock {LLMs Get Lost In Multi-Turn Conversation}.
\newblock \emph{arXiv preprint arXiv:2505.06120}, 2025.

\bibitem[Lee et~al.(2023)Lee, Hwang, Lee, Choi, and Park]{lee-etal-2023-complementarity}
Dohyeon Lee, Seung-won Hwang, Kyungjae Lee, Seungtaek Choi, and Sunghyun Park.
\newblock {On Complementarity Objectives for Hybrid Retrieval}.
\newblock In \emph{Proceedings of the 61st Annual Meeting of the Association for Computational Linguistics (Volume 1: Long Papers)}, pp.\  13357--13368, 2023.

\bibitem[Lewis et~al.(2020)Lewis, Perez, Piktus, Petroni, Karpukhin, Goyal, K\"{u}ttler, Lewis, Yih, Rockt\"{a}schel, Riedel, and Kiela]{NEURIPS2020_6b493230}
Patrick Lewis, Ethan Perez, Aleksandra Piktus, Fabio Petroni, Vladimir Karpukhin, Naman Goyal, Heinrich K\"{u}ttler, Mike Lewis, Wen-tau Yih, Tim Rockt\"{a}schel, Sebastian Riedel, and Douwe Kiela.
\newblock {Retrieval-Augmented Generation for Knowledge-Intensive NLP Tasks}.
\newblock In \emph{Advances in Neural Information Processing Systems}, pp.\  9459--9474, 2020.

\bibitem[Liu et~al.(2024{\natexlab{a}})Liu, Zaharia, and Abbeel]{liu2024ringattention}
Hao Liu, Matei Zaharia, and Pieter Abbeel.
\newblock {RingAttention with Blockwise Transformers for Near-Infinite Context}.
\newblock In \emph{International Conference on Learning Representations}, 2024{\natexlab{a}}.

\bibitem[Liu et~al.(2024{\natexlab{b}})Liu, Lin, Hewitt, Paranjape, Bevilacqua, Petroni, and Liang]{liu-etal-2024-lost}
Nelson~F. Liu, Kevin Lin, John Hewitt, Ashwin Paranjape, Michele Bevilacqua, Fabio Petroni, and Percy Liang.
\newblock {Lost in the Middle: How Language Models Use Long Contexts}.
\newblock \emph{Transactions of the Association for Computational Linguistics}, 12:\penalty0 157--173, 2024{\natexlab{b}}.

\bibitem[Liu et~al.(2023)Liu, Yu, Zhang, Xu, Lei, Lai, Gu, Ding, Men, Yang, Zhang, Deng, Zeng, Du, Zhang, Shen, Zhang, Su, Sun, Huang, Dong, and Tang]{liu2023agentbench}
Xiao Liu, Hao Yu, Hanchen Zhang, Yifan Xu, Xuanyu Lei, Hanyu Lai, Yu~Gu, Hangliang Ding, Kaiwen Men, Kejuan Yang, Shudan Zhang, Xiang Deng, Aohan Zeng, Zhengxiao Du, Chenhui Zhang, Sheng Shen, Tianjun Zhang, Yu~Su, Huan Sun, Minlie Huang, Yuxiao Dong, and Jie Tang.
\newblock Agentbench: Evaluating llms as agents.
\newblock \emph{arXiv preprint arXiv: 2308.03688}, 2023.

\bibitem[Ma et~al.(2023)Ma, Gong, He, Zhao, and Duan]{ma-etal-2023-query}
Xinbei Ma, Yeyun Gong, Pengcheng He, Hai Zhao, and Nan Duan.
\newblock {Query Rewriting in Retrieval-Augmented Large Language Models}.
\newblock In \emph{Proceedings of the 2023 Conference on Empirical Methods in Natural Language Processing}, pp.\  5303--5315, 2023.

\bibitem[Mei et~al.(2025)Mei, Yao, Ge, Wang, Bi, Cai, Liu, Li, Li, Zhang, Zhou, Mao, Xia, Guo, and Liu]{mei2025survey}
Lingrui Mei, Jiayu Yao, Yuyao Ge, Yiwei Wang, Baolong Bi, Yujun Cai, Jiazhi Liu, Mingyu Li, Zhong-Zhi Li, Duzhen Zhang, Chenlin Zhou, Jiayi Mao, Tianze Xia, Jiafeng Guo, and Shenghua Liu.
\newblock {A Survey of Context Engineering for Large Language Models}.
\newblock \emph{arXiv preprint 2507.13334}, 2025.

\bibitem[Microsoft(2025)]{Microsoft2025Hybrid}
Microsoft.
\newblock {Relevance Scoring in Hybrid Search Using Reciprocal Rank Fusion (RRF)}.
\newblock \url{https://learn.microsoft.com/en-us/azure/search/hybrid-search-ranking}, 2025.
\newblock Accessed: 2025-06-27.

\bibitem[OpenAI(2024)]{openai2024learning}
OpenAI.
\newblock {Learning to reason with LLMs}.
\newblock \url{https://openai.com/index/learning-to-reason-with-llms/}, 2024.
\newblock Accessed: Sep. 11, 2025.

\bibitem[Page et~al.(1999)Page, Brin, Motwani, and Winograd]{Page1999ThePC}
Lawrence Page, Sergey Brin, Rajeev Motwani, and Terry Winograd.
\newblock {The PageRank Citation Ranking : Bringing Order to the Web}.
\newblock In \emph{The Web Conference}, 1999.

\bibitem[Peng et~al.(2024)Peng, Quesnelle, Fan, and Shippole]{peng2024yarn}
Bowen Peng, Jeffrey Quesnelle, Honglu Fan, and Enrico Shippole.
\newblock {Ya{RN}: Efficient Context Window Extension of Large Language Models}.
\newblock In \emph{International Conference on Learning Representations}, 2024.

\bibitem[Robertson \& Zaragoza(2009)Robertson and Zaragoza]{10.1561/1500000019}
Stephen Robertson and Hugo Zaragoza.
\newblock {The Probabilistic Relevance Framework: BM25 and Beyond}.
\newblock \emph{Found. Trends Inf. Retr.}, 3\penalty0 (4):\penalty0 333–389, 2009.

\bibitem[Robertson et~al.(1994)Robertson, Walker, Jones, Hancock-Beaulieu, and Gatford]{conf/trec/RobertsonWJHG94}
Stephen~E. Robertson, Steve Walker, Susan Jones, Micheline Hancock-Beaulieu, and Mike Gatford.
\newblock {Okapi at TREC-3.}
\newblock In \emph{TREC}, volume 500-225 of \emph{NIST Special Publication}, pp.\  109--126, 1994.

\bibitem[Sainz et~al.(2023)Sainz, Campos, Garc{\'i}a-Ferrero, Etxaniz, de~Lacalle, and Agirre]{sainz-etal-2023-nlp}
Oscar Sainz, Jon Campos, Iker Garc{\'i}a-Ferrero, Julen Etxaniz, Oier~Lopez de~Lacalle, and Eneko Agirre.
\newblock {{NLP} Evaluation in trouble: On the Need to Measure {LLM} Data Contamination for each Benchmark}.
\newblock In \emph{Findings of the Association for Computational Linguistics: EMNLP 2023}, pp.\  10776--10787, 2023.

\bibitem[Shaham et~al.(2022)Shaham, Segal, Ivgi, Efrat, Yoran, Haviv, Gupta, Xiong, Geva, Berant, and Levy]{shaham-etal-2022-scrolls}
Uri Shaham, Elad Segal, Maor Ivgi, Avia Efrat, Ori Yoran, Adi Haviv, Ankit Gupta, Wenhan Xiong, Mor Geva, Jonathan Berant, and Omer Levy.
\newblock {{SCROLLS}: Standardized {C}ompa{R}ison Over Long Language Sequences}.
\newblock In \emph{Proceedings of the 2022 Conference on Empirical Methods in Natural Language Processing}, pp.\  12007--12021, 2022.

\bibitem[Shaham et~al.(2023)Shaham, Ivgi, Efrat, Berant, and Levy]{shaham-etal-2023-zeroscrolls}
Uri Shaham, Maor Ivgi, Avia Efrat, Jonathan Berant, and Omer Levy.
\newblock {{Z}ero{SCROLLS}: A Zero-Shot Benchmark for Long Text Understanding}.
\newblock In \emph{Findings of the Association for Computational Linguistics: EMNLP 2023}, pp.\  7977--7989, 2023.

\bibitem[Shinn et~al.(2023)Shinn, Cassano, Gopinath, Narasimhan, and Yao]{shinn2023reflexion}
Noah Shinn, Federico Cassano, Ashwin Gopinath, Karthik~R Narasimhan, and Shunyu Yao.
\newblock {Reflexion: language agents with verbal reinforcement learning}.
\newblock In \emph{Advances in Neural Information Processing Systems}, 2023.

\bibitem[Su et~al.(2024)Su, Ahmed, Lu, Pan, Bo, and Liu]{SU2024127063}
Jianlin Su, Murtadha Ahmed, Yu~Lu, Shengfeng Pan, Wen Bo, and Yunfeng Liu.
\newblock {RoFormer: Enhanced transformer with Rotary Position Embedding}.
\newblock \emph{Neurocomputing}, 568:\penalty0 127063, 2024.

\bibitem[Thakur et~al.(2021)Thakur, Reimers, R{\"u}ckl{\'e}, Srivastava, and Gurevych]{thakur2021beir}
Nandan Thakur, Nils Reimers, Andreas R{\"u}ckl{\'e}, Abhishek Srivastava, and Iryna Gurevych.
\newblock {{BEIR}: A Heterogeneous Benchmark for Zero-shot Evaluation of Information Retrieval Models}.
\newblock In \emph{Advances in Neural Information Processing Systems Datasets and Benchmarks Track}, 2021.

\bibitem[Trivedi et~al.(2022)Trivedi, Balasubramanian, Khot, and Sabharwal]{trivedi-etal-2022-musique}
Harsh Trivedi, Niranjan Balasubramanian, Tushar Khot, and Ashish Sabharwal.
\newblock {{M}u{S}i{Q}ue: Multihop Questions via Single-hop Question Composition"}.
\newblock \emph{Transactions of the Association for Computational Linguistics}, 10:\penalty0 539--554, 2022.

\bibitem[Trivedi et~al.(2023)Trivedi, Balasubramanian, Khot, and Sabharwal]{trivedi-etal-2023-interleaving}
Harsh Trivedi, Niranjan Balasubramanian, Tushar Khot, and Ashish Sabharwal.
\newblock {Interleaving Retrieval with Chain-of-Thought Reasoning for Knowledge-Intensive Multi-Step Questions}.
\newblock In \emph{Proceedings of the 61st Annual Meeting of the Association for Computational Linguistics (Volume 1: Long Papers)}, pp.\  10014--10037, 2023.

\bibitem[Wang et~al.(2025)Wang, Ning, Pan, Wu, Guo, Deng, Bao, Hu, Zhang, Wang, and Zhang]{wang2025novelqa}
Cunxiang Wang, Ruoxi Ning, Boqi Pan, Tonghui Wu, Qipeng Guo, Cheng Deng, Guangsheng Bao, Xiangkun Hu, Zheng Zhang, Qian Wang, and Yue Zhang.
\newblock {Novel{QA}: Benchmarking Question Answering on Documents Exceeding 200K Tokens}.
\newblock In \emph{International Conference on Learning Representations}, 2025.

\bibitem[Wang et~al.(2023)Wang, Yang, and Wei]{wang-etal-2023-query2doc}
Liang Wang, Nan Yang, and Furu Wei.
\newblock {Query2doc: Query Expansion with Large Language Models}.
\newblock In \emph{Proceedings of the 2023 Conference on Empirical Methods in Natural Language Processing}, pp.\  9414--9423, 2023.

\bibitem[Wang et~al.(2024{\natexlab{a}})Wang, Chen, Cheng, Liao, Zhang, Wu, Yu, Xu, Zhang, Luo, Li, Yang, Huang, and Li]{wang-etal-2024-leave}
Minzheng Wang, Longze Chen, Fu~Cheng, Shengyi Liao, Xinghua Zhang, Bingli Wu, Haiyang Yu, Nan Xu, Lei Zhang, Run Luo, Yunshui Li, Min Yang, Fei Huang, and Yongbin Li.
\newblock {Leave No Document Behind: Benchmarking Long-Context {LLM}s with Extended Multi-Doc {QA}}.
\newblock In \emph{Proceedings of the 2024 Conference on Empirical Methods in Natural Language Processing}, pp.\  5627--5646, 2024{\natexlab{a}}.

\bibitem[Wang et~al.(2024{\natexlab{b}})Wang, Wang, Liu, Chen, Yuan, Peng, and Ji]{wang2024mint}
Xingyao Wang, Zihan Wang, Jiateng Liu, Yangyi Chen, Lifan Yuan, Hao Peng, and Heng Ji.
\newblock {{MINT}: Evaluating {LLM}s in Multi-turn Interaction with Tools and Language Feedback}.
\newblock In \emph{International Conference on Learning Representations}, 2024{\natexlab{b}}.

\bibitem[Xiao et~al.(2024)Xiao, Tian, Chen, Han, and Lewis]{xiao2024efficient}
Guangxuan Xiao, Yuandong Tian, Beidi Chen, Song Han, and Mike Lewis.
\newblock {Efficient Streaming Language Models with Attention Sinks}.
\newblock In \emph{International Conference on Learning Representations}, 2024.

\bibitem[Xu et~al.(2024)Xu, Ping, Wu, McAfee, Zhu, Liu, Subramanian, Bakhturina, Shoeybi, and Catanzaro]{xu2024retrieval}
Peng Xu, Wei Ping, Xianchao Wu, Lawrence McAfee, Chen Zhu, Zihan Liu, Sandeep Subramanian, Evelina Bakhturina, Mohammad Shoeybi, and Bryan Catanzaro.
\newblock {Retrieval meets Long Context Large Language Models}.
\newblock In \emph{International Conference on Learning Representations}, 2024.

\bibitem[Yang et~al.(2025{\natexlab{a}})]{qwen2.5-1M}
An~Yang et~al.
\newblock Qwen2.5-1m technical report.
\newblock \emph{arXiv preprint arXiv:2501.15383}, 2025{\natexlab{a}}.

\bibitem[Yang et~al.(2025{\natexlab{b}})]{qwen3}
An~Yang et~al.
\newblock {Qwen3 Technical Report}.
\newblock \emph{arXiv preprint arXiv:2505.09388}, 2025{\natexlab{b}}.

\bibitem[Yang et~al.(2025{\natexlab{c}})Yang, Chen, and Chen]{yang2025ape}
Xinyu Yang, Tianqi Chen, and Beidi Chen.
\newblock {{APE}: Faster and Longer Context-Augmented Generation via Adaptive Parallel Encoding}.
\newblock In \emph{International Conference on Learning Representations}, 2025{\natexlab{c}}.

\bibitem[Yang et~al.(2018)Yang, Qi, Zhang, Bengio, Cohen, Salakhutdinov, and Manning]{yang-etal-2018-hotpotqa}
Zhilin Yang, Peng Qi, Saizheng Zhang, Yoshua Bengio, William Cohen, Ruslan Salakhutdinov, and Christopher~D. Manning.
\newblock {HotpotQA: A Dataset for Diverse, Explainable Multi-hop Question Answering}.
\newblock In \emph{Proceedings of the 2018 Conference on Empirical Methods in Natural Language Processing}, pp.\  2369--2380, 2018.

\bibitem[Yao et~al.(2023)Yao, Zhao, Yu, Du, Shafran, Narasimhan, and Cao]{yao2023react}
Shunyu Yao, Jeffrey Zhao, Dian Yu, Nan Du, Izhak Shafran, Karthik~R Narasimhan, and Yuan Cao.
\newblock {ReAct: Synergizing Reasoning and Acting in Language Models}.
\newblock In \emph{International Conference on Learning Representations}, 2023.

\bibitem[Yen et~al.(2025)Yen, Gao, Hou, Ding, Fleischer, Izsak, Wasserblat, and Chen]{yen2025helmet}
Howard Yen, Tianyu Gao, Minmin Hou, Ke~Ding, Daniel Fleischer, Peter Izsak, Moshe Wasserblat, and Danqi Chen.
\newblock {HELMET: How to Evaluate Long-context Models Effectively and Thoroughly}.
\newblock In \emph{International Conference on Learning Representations}, 2025.

\bibitem[Yuan et~al.(2025)Yuan, Gao, Dai, Luo, Zhao, Zhang, Xie, Wei, Wang, Xiao, Wang, Ruan, Zhang, Liang, and Zeng]{yuan-etal-2025-native}
Jingyang Yuan, Huazuo Gao, Damai Dai, Junyu Luo, Liang Zhao, Zhengyan Zhang, Zhenda Xie, Yuxing Wei, Lean Wang, Zhiping Xiao, Yuqing Wang, Chong Ruan, Ming Zhang, Wenfeng Liang, and Wangding Zeng.
\newblock {Native Sparse Attention: Hardware-Aligned and Natively Trainable Sparse Attention}.
\newblock In \emph{Proceedings of the 63rd Annual Meeting of the Association for Computational Linguistics (Volume 1: Long Papers)}, pp.\  23078--23097, 2025.

\bibitem[Yuan et~al.(2024)Yuan, Ning, Zhou, Yang, Li, Zhuang, Tan, Yao, Lin, Li, Dai, Yan, and Wang]{yuan2024lveval}
Tao Yuan, Xuefei Ning, Dong Zhou, Zhijie Yang, Shiyao Li, Minghui Zhuang, Zheyue Tan, Zhuyu Yao, Dahua Lin, Boxun Li, Guohao Dai, Shengen Yan, and Yu~Wang.
\newblock {LV-Eval: A Balanced Long-Context Benchmark with 5 Length Levels Up to 256K}.
\newblock \emph{arXiv preprint arXiv:2402.05136}, 2024.

\bibitem[Zhang et~al.(2024)Zhang, Chen, Hu, Xu, Chen, Hao, Han, Thai, Wang, Liu, and Sun]{zhang-etal-2024-bench}
Xinrong Zhang, Yingfa Chen, Shengding Hu, Zihang Xu, Junhao Chen, Moo Hao, Xu~Han, Zhen Thai, Shuo Wang, Zhiyuan Liu, and Maosong Sun.
\newblock {$\infty${B}ench: Extending Long Context Evaluation Beyond 100{K} Tokens}.
\newblock In \emph{Proceedings of the 62nd Annual Meeting of the Association for Computational Linguistics (Volume 1: Long Papers)}, pp.\  15262--15277, 2024.

\bibitem[Zhang et~al.(2025)Zhang, Li, Long, Zhang, Lin, Yang, Xie, Yang, Liu, Lin, Huang, and Zhou]{qwen3embedding}
Yanzhao Zhang, Mingxin Li, Dingkun Long, Xin Zhang, Huan Lin, Baosong Yang, Pengjun Xie, An~Yang, Dayiheng Liu, Junyang Lin, Fei Huang, and Jingren Zhou.
\newblock Qwen3 embedding: Advancing text embedding and reranking through foundation models.
\newblock \emph{arXiv preprint arXiv:2506.05176}, 2025.

\bibitem[Zheng et~al.(2023)Zheng, Chiang, Sheng, Zhuang, Wu, Zhuang, Lin, Li, Li, Xing, Zhang, Gonzalez, and Stoica]{zheng2023judging}
Lianmin Zheng, Wei-Lin Chiang, Ying Sheng, Siyuan Zhuang, Zhanghao Wu, Yonghao Zhuang, Zi~Lin, Zhuohan Li, Dacheng Li, Eric Xing, Hao Zhang, Joseph~E. Gonzalez, and Ion Stoica.
\newblock {Judging {LLM}-as-a-Judge with {MT}-Bench and Chatbot Arena}.
\newblock In \emph{Advances in Neural Information Processing Systems Datasets and Benchmarks Track}, 2023.

\end{thebibliography}
\bibliographystyle{iclr2026_conference}

\appendix
\section{LLM Usage Disclosure}

We use Gemini 2.5 Pro, GPT-5, and Grok for writing enhancements, primarily to improve grammar and overall text flow. We also use their DeepResearch capabilities to retrieve related works for more contextualized discussions. All LLM outputs were reviewed and verified by the authors to ensure accuracy and avoid factual errors or hallucinations.

\section{NIAH Prompt}
\label{sec:NIAH_prompt}

\begin{tcolorbox}[colback=gray!5!white, colframe=gray!75!black, title=Input Prompt for NIAH Evaluation]
\scriptsize

\raggedright Read the following articles and answer the question below. \\
\vspace{3mm}
\{ordered haystack\}
\vspace{3mm} \\
What is the correct answer to this question: \{question\} \\
\vspace{3mm}
Format your response as follows: ``The correct answer is (insert answer here)''.
\end{tcolorbox}
\label{fig:prompt_NIAH}

\section{Dynamic NIAH Prompts}
\label{sec:dynamic_NIAH_prompts}

\subsection{Enforced Multi-Round}

\begin{tcolorbox}[colback=gray!5!white, colframe=gray!75!black, title=Input Prompt for Intermediate Rounds in Enforced Multi-Round NIAH Evaluation]
\scriptsize

\raggedright Read your previous analyses and the following articles. Analyze the question below. \\
\vspace{3mm}
Previous Analyses: \{analyses\} \\
\vspace{3mm}
Articles: \{ordered haystack\} \\
\vspace{3mm}
Question: \{question\} \\
\vspace{3mm}
Based on your previous analyses and the potentially new articles provided, summarize your findings related to the question and refine the question. \\
\vspace{3mm}
Format your response as follows: \\
\vspace{3mm}
Summary: (Summarize what you found in the articles that relates to the question, including any partial answers, relevant context, or gaps in information.) \\
\vspace{3mm}
Refined Question: (Copy the original question or replace it with a more specific question based on your findings.) 
\end{tcolorbox}
\label{fig:prompt_NIAH_enforced_multi_intermediate}

\begin{tcolorbox}[colback=gray!5!white, colframe=gray!75!black, title=Input Prompt for Final Round in Enforced Multi-Round NIAH Evaluation]
\scriptsize

\raggedright Read your previous analyses and the following articles, and answer the question below. \\
\vspace{3mm}
Previous Analyses: \{analyses\} \\
\vspace{3mm}
Articles: \{ordered haystack\} \\
\vspace{3mm} 
What is the correct answer to this question: \{question\} \\
\vspace{3mm}
Format your response as follows: ``The correct answer is (insert answer here)''.
\end{tcolorbox}
\label{fig:prompt_NIAH_enforced_multi_final}

\subsection{Variable-Round}

\begin{tcolorbox}[colback=gray!5!white, colframe=gray!75!black, title=Input Prompt for Variable-Round NIAH Evaluation]
\scriptsize

\raggedright Read your previous analyses and the following articles. Analyze the question below. \\
\vspace{3mm}
Previous Analyses: \{analyses\} \\
\vspace{3mm}
Articles: \{ordered haystack\} \\
\vspace{3mm} 
Question: \{question\} \\
\vspace{3mm}
Based on your previous analyses and the potentially new articles provided, decide if you are confident in answering the question or if you need additional information. \\
\vspace{3mm}
If you have complete information to fully answer the question, format your response as follows: "The correct answer is (insert answer here)". \\
\vspace{3mm}
If you need more information, format your response as follows:\\
Summary: (Summarize what you found in the articles that relates to the question, including any partial answers, relevant context, or gaps in information.) \\
\vspace{3mm}
Refined Question: (Copy the original question or replace it with a more specific question based on your findings.) 
\end{tcolorbox}
\label{fig:prompt_NIAH_variable}

\section{More Dataset Details}
\label{appendix:dataset_details}

In preparing the Wikipedia hyperlink network, we filter out empty and redirect pages. 

Table~\ref{tab:hop_count} provides a dataset breakdown over hop count.

\begin{table}[h]
\caption{Question breakdown over hop count.}
\vspace{3mm}
\centering
\begin{adjustbox}{width=0.18\textwidth}
\begin{tabular}{lc}
\toprule
\# hops & \% \\
\midrule
1 & $20$ \\
2 & $58$ \\
3 & $15.6$ \\
4 & $6.4$\\
\bottomrule
\end{tabular}
\end{adjustbox}
\label{tab:hop_count}
\end{table}

\section{Additional Setup Details}

\label{appendix:setup_details}

\subsection{LLM Setup}

For each LLM, we utilize the recommended inference hyperparameters as specified on its Hugging Face model card. These settings include sampling parameters like temperature, Top-P, Top-K, and Min-P, along with the ``thinking budget'' for thinking LLMs. All models considered in this work possess native long-context support for at least $128K$ tokens, with the exception of the Qwen3 models. To ensure the Qwen3 models could process a $128K$-token input and generate a $32K$-token output, we extend their context window to $164K$ tokens using YaRN~\citep{peng2024yarn}.

\subsection{PPR Setup}

We perform a hyperparameter search for PPR per retriever using $10 \%$ of the QA samples. For retrieval criteria, we adopt Normalized Discounted Cumulative Gain (NDCG) @ $10K$~\citep{10.1145/345508.345545, 10.1145/582415.582418} for ranking ground truth supporting documents among the corpus. Table~\ref{tab:ppr_hyper} presents the best hyperparameters for each retriever based on three random seeds.

\begin{table}[h]
\caption{Retriever-specific PPR hyperparameters.}
\vspace{3mm}
\centering
\begin{adjustbox}{width=0.5\textwidth}
\begin{tabular}{lcc}
\toprule
Retriever & \# Seed Documents &  Damping Factor \\
\midrule
BM25 & 10 & 0.5 \\
Qwen3-0.6B & 5 & 0.5\\
Hybrid & 5 & 0.85\\
\bottomrule
\end{tabular}
\end{adjustbox}
\label{tab:ppr_hyper}
\end{table}

\section{Evaluation for Data Contamination}

\label{appendix:contamination}

To quantify data contamination, we evaluate LLM performance under two conditions: 1) without context, to test reliance on parametric knowledge, and 2) with ground-truth supporting documents. We measure F1 scores across an increasing number of the question hop count to assess how performance varies with reasoning complexity.

\begin{figure}[h]
    \centering
    \includegraphics[width=\linewidth]{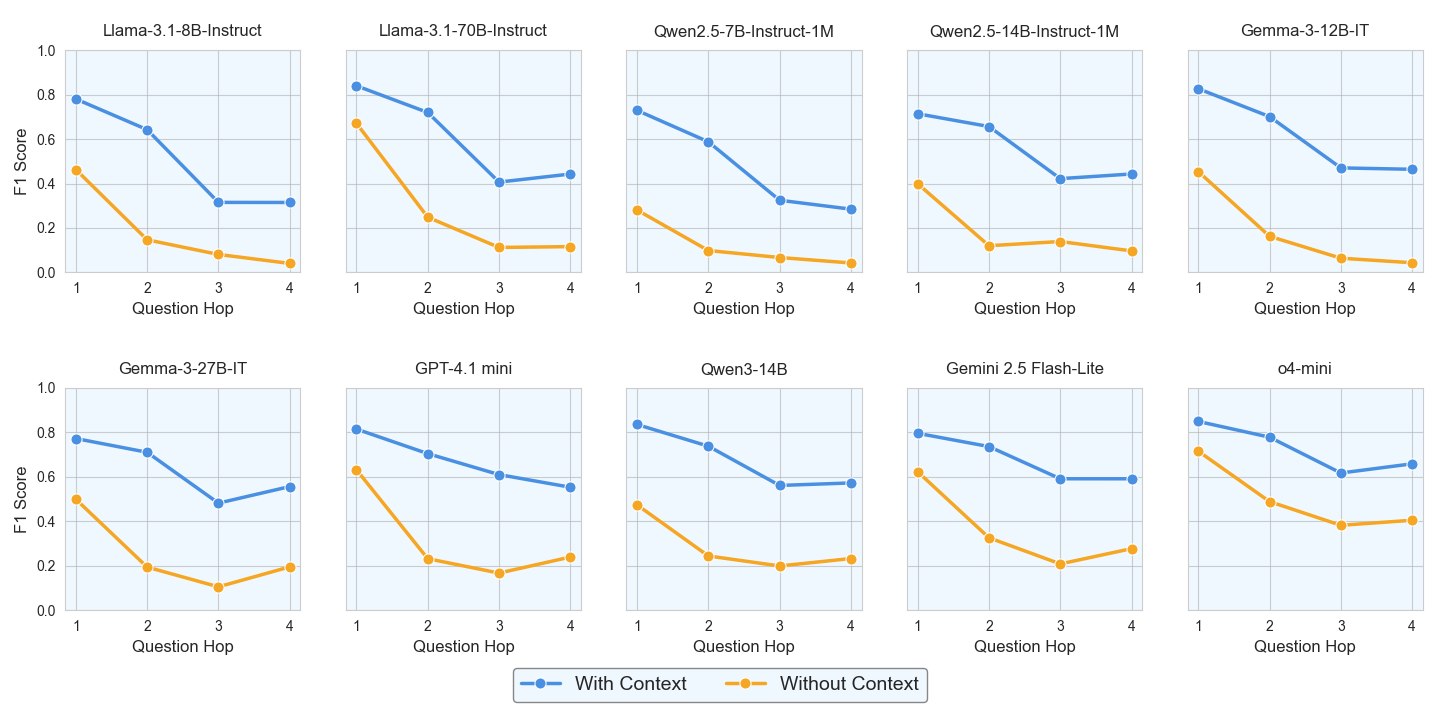}
    \caption{
    LLM performance with vs without context across question hop.
    \label{fig:context_vs_no_context}}
\end{figure}

Fig.~\ref{fig:context_vs_no_context} presents the evaluation results. 
\begin{itemize}
    \item \textbf{Contamination is evident.} All models achieve non-zero F1 scores without context. This indicates a degree of data contamination.
    \item \textbf{Context is crucial.} Despite contamination, providing ground-truth documents substantially improves the performance of all models.
    \item \textbf{Complexity remains a challenge.} F1 scores generally decrease as the question hop count increases, even when context is provided. This also suggests that evaluation with multi-hop questions suffers less from data contamination.
\end{itemize}

\section{Case Studies for Failure Patterns in Dynamic NIAH}

\label{appendix:frequent_cascading_failure_patterns}

\begin{tcolorbox}[colback=gray!5!white, colframe=gray!75!black, title=Gemma-3-12B-IT propagates its reasoning errors via query refinement and summarization.]
\scriptsize

\raggedright [Original query]: What body of water is by the place John Dury died? \\
\vspace{3mm}
[First-round response]: \\
\vspace{3mm}
Summary: The provided text discusses John Dury and his life, focusing on his religious and intellectual work. It states that he died in Edinburgh. Edinburgh is located on the Firth of Forth. Therefore, John Dury died near the Firth of Forth.  \; {\color{red}{// Wrong. He was born in Edinburgh.}} \\
\vspace{3mm}
Refined Question: What body of water is located near Edinburgh, where John Dury died? \; {\color{red}{// The mistake got consolidated and will inflate the ranking of distractor documents.}} \\
\vspace{3mm}
[Second-round response]: \\
\vspace{3mm}
Summary: The provided articles offer several pieces of information related to John Dury and the geography of Edinburgh. Firstly, the article on Edinburgh states it’s located ``near the Firth of Forth''. Secondly, the article on John Dury mentions that he died in Edinburgh, and Edinburgh is located on the Firth of Forth. Therefore, John Dury died near the Firth of Forth. The articles do not offer any other specific bodies of water near Edinburgh. \\
\vspace{3mm}
Refined Question: What body of water is Edinburgh situated on, in which John Dury died?  \; {\color{red}{// The LLM cannot correct consolidated early mistakes.}} \\
\vspace{3mm}
[Final answer prediction]: The correct answer is the Firth of Forth. \; {\color{red}{// The final answer is then naturally wrong.}} \\
\end{tcolorbox}

\begin{tcolorbox}[colback=gray!5!white, colframe=gray!75!black, title=GPT-5 deviated from the original query intent.]
\scriptsize

\raggedright [Original query]: What year saw the formation of the Brazilian republic? \\
\vspace{3mm}
[Refined query after the first round]: In what year (and on what date) was the Brazilian Republic proclaimed? \; {\color{red}{// GPT-5 added ``on what date''}} \\
\vspace{3mm}
[Refined query after the second round]: In what year, and on what date, was the Brazilian Republic proclaimed? (Answer: 15 November 1889) \; {\color{red}{// GPT-5 reinforced the earlier query deviation.}} \\
\vspace{3mm}
[Final answer prediction]: The correct answer is 15 November 1889. \; {\color{red}{// The correct answer should be just 1889.}} 
\end{tcolorbox}

\begin{tcolorbox}[colback=gray!5!white, colframe=gray!75!black, title=GPT-5 mini changed open questions to yes/no questions.]
\scriptsize

\raggedright [Original query]: What country is the member of the band that performed Visions of Eden a citizen of? \\
\vspace{3mm}
[Refined query after the first round]: Which country is David DeFeis (the Virgin Steele member who performed on Visions of Eden) a citizen of? \\
\vspace{3mm}
[Refined query after the second round]: Is David DeFeis a citizen of the United States? If so, can you cite a source that explicitly confirms his country of citizenship (e.g., place of birth, official biography or other primary documentation)? \; {\color{red}{// The question got changed to an yes/no question.}} \\
\vspace{3mm}
[Final answer prediction]: Yes. \; {\color{red}{// GPT-5 mini even failed to answer its own second question, despite the deviation.}} \\
\end{tcolorbox}

\begin{tcolorbox}[colback=gray!5!white, colframe=gray!75!black, title=GPT-5 mini changed open questions to multiple-choice questions.]
\scriptsize

\raggedright [Original query]: What year did unification of the country Mintu is located happen? \\
\vspace{3mm}
[Refined query after the first round]: Do you mean the year Sarawak became part of the federation of Malaysia (i.e., when Malaya, North Borneo/Sabah, Sarawak and Singapore united) — which was 1963 — or do you mean the earlier independence of the Federation of Malaya in 1957? \\
\vspace{3mm}
[Refined query after the second round]: Do you mean (A) the year Sarawak (and thus Mintu) joined the Federation of Malaysia — Malaysia Day, 16 September 1963 — or (B) the earlier independence of the Federation of Malaya — Hari Merdeka, 31 August 1957? \; {\color{red}{// The question got changed to a multiple-choice question.}} \\
\vspace{3mm}
[Final answer prediction]: A. \\
\end{tcolorbox}

\begin{tcolorbox}[colback=gray!5!white, colframe=gray!75!black, title=Gemini 2.5 Pro fails due to long-context challenges.]
\scriptsize

\raggedright [Original query]: What is the nickname of the state where John Kerry's VP pick was a senator? \\
\vspace{3mm}
[First-round response]: \\
\vspace{3mm}
Summary: The articles identify John Kerry's vice presidential running mate in the 2004 presidential election as John Edwards. At the time of his selection, Edwards was a U.S. Senator from North Carolina. The article titled ``North Carolina'' provides two slogans used on the state's license plates: ``First in Flight'' and ``First in Freedom''. While the article also refers to ``Tar heel voters'', it does not explicitly state that ``The Tar Heel State'' is the official nickname. \\
\vspace{3mm}
Refined Question: What is the nickname of the state where John Kerry's VP pick was a senator?  \; {\color{red}{// Gemini 2.5 Pro simply fails in identifying relevant text about ``Old North State''. It also does not update the query properly by incorporating partial evidence.}}\\
\vspace{3mm}
\raggedright [Second-round response]: Summary: The articles confirm that John Kerry's vice presidential running mate in the 2004 presidential election was John Edwards. Multiple articles state that Edwards was a U.S. Senator from North Carolina at the time of his selection. The article on North Carolina mentions two slogans used on the state's license plates, ``First in Flight'' and ``First in Freedom'', but none of the provided texts explicitly identify the state's official nickname.\\
\vspace{3mm}
Refined Question: What is the nickname of the state where John Kerry's VP pick was a senator? \\
\vspace{3mm}
[Final answer prediction]: Based on the provided articles, the correct answer is that the texts do not explicitly state an official nickname for North Carolina.
\end{tcolorbox}

\section{Raw Experiment Results for Retrieval Evaluation}

\label{sec:detailed_retrieval_eval}

See Table~\ref{tab:retrieval_eval_recall} and Table~\ref{tab:retrieval_eval_ndcg}.

\begin{table}[t]
\caption{Recall$@N$ of retrieval strategies for coverage evaluation $(\times 10^{-2}, \uparrow)$, with a breakdown over question hop. We present the results in a way that allows comparing the impact of using PPR or not, and we highlight the better results.}
\centering
\begin{adjustbox}{width=0.85\textwidth}
\begin{tabular}{lccccccc}
\toprule
Base Retriever & Hop & PPR & $@10$ & $@20$ & $@40$ & $@80$ & $@160$\\
\midrule
BM25 & Overall  &            & 37.93 & 42.83 & 47.43 & 53.13 & 58.73 \\
     &          & \checkmark & \textbf{38}    & \textbf{55.02} & \textbf{59.97} & \textbf{63.63} & \textbf{66.58} \\
\midrule 
BM25 & 1-hop    &            & \textbf{66}    & 72    & \textbf{78}    & \textbf{86} & \textbf{88} \\
     &          & \checkmark & \textbf{66}    & \textbf{77}     & \textbf{78}   & 78 & 80 \\
\midrule 
BM25 & 2-hop    &            & \textbf{36.21}   & 41.03 & 45.17 & 49.14 & 54.14\\
     &          & \checkmark & \textbf{36.21} & \textbf{57.41} & \textbf{62.76} & \textbf{66.03} & \textbf{69.31} \\
\midrule 
BM25 & 3-hop    &            & 20.09   & 23.93 & 29.06 & 38.03 & 49.57 \\
     &          & \checkmark & \textbf{20.51} & \textbf{35.04} & \textbf{44.02} & \textbf{53.42} & \textbf{55.98} \\
\midrule 
BM25 & 4-hop    &            & \textbf{9.38}    & \textbf{14.06} & \textbf{17.19} & \textbf{23.44} & \textbf{31.25} \\
     &          & \checkmark & \textbf{9.38}    & 13.28 & \textbf{17.19} & 21.88 & 25.78 \\
\midrule
Qwen3-0.6B & Overall &            & 45    & 49.87 & 53.48 & 57.3  & 61.43 \\
           &         & \checkmark & \textbf{52.35} & \textbf{59.05} & \textbf{65.7} & \textbf{71.12} & \textbf{74.28} \\
\midrule 
Qwen3-0.6B & 1-hop   &            & 80    & 86    & 89    & 91    & \textbf{94}    \\
           &         & \checkmark & \textbf{86}    & \textbf{87}    & \textbf{92} & \textbf{93} & \textbf{94} \\
\midrule 
Qwen3-0.6B & 2-hop   &            & 42.07 & 46.38 & 50.52 & 53.97 & 58.28 \\
           &         & \checkmark & \textbf{51.72} & \textbf{61.03} & \textbf{68.97} & \textbf{76.38} & \textbf{79.48}\\
\midrule
Qwen3-0.6B & 3-hop   &            & 24.36 & 29.91 & 32.91 & 39.74 & 44.44 \\
           &         & \checkmark & \textbf{28.21} & \textbf{34.62} & \textbf{39.74} & \textbf{44.02} & \textbf{49.57} \\
\midrule
Qwen3-0.6B & 4-hop   &            & \textbf{12.5}  & \textbf{17.19} & \textbf{19.53} & \textbf{25}    & \textbf{29.69} \\
           &         & \checkmark & 11.72 & 13.28 & 17.19 & 21.09 & 25.78 \\
\midrule 
Hybrid & Overall &            & 51.07 & 55.88 & 60.28 & 64 & 67.2 \\
       &         & \checkmark & \textbf{58.23} & \textbf{63.05} & \textbf{69.1} & \textbf{73.52} & \textbf{76.55} \\
\midrule 
Hybrid & 1-hop   &            & \textbf{89}    & \textbf{95}    & \textbf{96}        & \textbf{96} & \textbf{97} \\
       &         & \checkmark & 83    & 87    & 90 & 92 & 93 \\
\midrule 
Hybrid & 2-hop   &            & 47.41 & 52.07 & 56.55 & 60.17 & 62.93\\
       &         & \checkmark & \textbf{61.21} & \textbf{66.03} & \textbf{71.9} & \textbf{76.38} & \textbf{79.14} \\
\midrule
Hybrid & 3-hop   &            & 28.63 & 32.91 & 41.88 & 50 & 56.41\\
       &         & \checkmark & \textbf{31.62} & \textbf{38.46} & \textbf{50} & \textbf{56.84} & \textbf{61.54} \\
\midrule
Hybrid & 4-hop   &            & \textbf{20.31} & \textbf{24.22} & \textbf{27.34} & \textbf{32.81} & \textbf{39.06} \\
       &         & \checkmark & 18.75 & 21.09 & 25 & 30.47 & 38.28 \\
\bottomrule
\end{tabular}
\end{adjustbox}
\label{tab:retrieval_eval_recall}
\end{table}

\begin{table}[t]
\caption{NDCG$@N$ of retrieval strategies for ranking evaluation $(\times 10^{-2}, \uparrow)$, with a breakdown over question hop. We present the results in a way that allows comparing the impact of using PPR or not, and we highlight the better results.}
\centering
\begin{adjustbox}{width=0.85\textwidth}
\begin{tabular}{lccccccc}
\toprule
Base Retriever & Hop & PPR & $@10$ & $@20$ & $@40$ & $@80$ & $@160$\\
\midrule
BM25 & Overall  &            & 31.31 & 32.83 & 33.96 & 35.16 & 36.22\\
     &          & \checkmark & \textbf{32.86} & \textbf{38.35} & \textbf{39.66} & \textbf{40.49} & \textbf{41.03}\\
\midrule 
BM25 & 1-hop    &            & 43.71 & 45.3 & 46.5        & 47.9  & 48.2\\
     &          & \checkmark & \textbf{47.94} & \textbf{50.77} & \textbf{50.99} & \textbf{50.99} & \textbf{51.27}\\
\midrule 
BM25 & 2-hop    &            & 32.74 & 34.22 & 35.25 & 36.07 & 36.97\\
     &          & \checkmark & \textbf{33.64} & \textbf{40.52} & \textbf{41.87} & \textbf{42.55} & \textbf{43.14}\\
\midrule 
BM25 & 3-hop    &            & 20.14 & 21.58 & 23.03 & 25.17 & 27.56\\
     &          & \checkmark & \textbf{20.76} & \textbf{26.13} & \textbf{28.7} & \textbf{30.95} & \textbf{31.49}\\
\midrule 
BM25 & 4-hop    &            & 6.76  & 8.69  & 9.71       & 11.4  & \textbf{13.18} \\
     &          & \checkmark & \textbf{8.06} & \textbf{9.71} & \textbf{10.98} & \textbf{12.24} & 13.12\\
\midrule
Qwen3-0.6B & Overall &            & 41.18 & 42.7 & 43.6 & 44.43 & 45.2\\
           &         & \checkmark & \textbf{44.91} & \textbf{47.04} & \textbf{48.72} & \textbf{49.89} & \textbf{50.49}\\
\midrule 
Qwen3-0.6B & 1-hop   &            & 61.72 & 63.3 & 63.89 & 64.22 & 64.65\\
           &         & \checkmark & \textbf{67.53} & \textbf{67.79} & \textbf{68.79} & \textbf{68.96} & \textbf{69.11} \\
\midrule 
Qwen3-0.6B & 2-hop   &            & 41.86 & 43.2 & 44.24 & 44.96 & 45.75\\
           &         & \checkmark & \textbf{45.53} & \textbf{48.42} & \textbf{50.46} & \textbf{52.02} & \textbf{52.58}\\
\midrule
Qwen3-0.6B & 3-hop   &            & 23.58 & 25.51 & 26.35 & 28.01 & 28.97\\
           &         & \checkmark & \textbf{26.25} & \textbf{28.57} & \textbf{30.03} & \textbf{31.07} & \textbf{32.2}\\
\midrule
Qwen3-0.6B & 4-hop   &            & 13.8  & \textbf{15.63} & \textbf{16.35} & \textbf{17.83} & \textbf{18.92} \\
           &         & \checkmark & \textbf{14.09} & 14.67 & 15.86 & 16.87 & 17.93\\
\midrule 
Hybrid & Overall &            & 44.91 & 46.41 & 47.57 & 48.39 & 49\\
       &         & \checkmark & \textbf{48.81} & \textbf{50.3} & \textbf{51.89} & \textbf{52.86} & \textbf{53.45}\\
\midrule 
Hybrid & 1-hop   &            & \textbf{67.33} & \textbf{68.87} & \textbf{69.06} & \textbf{69.06} & \textbf{69.21}\\
       &         & \checkmark & 66.75 & 67.74 & 68.36 & 68.69 & 68.84\\
\midrule 
Hybrid & 2-hop   &            & 44.75 & 46.21 & 47.34 & 48.1  & 48.59\\
       &         & \checkmark & \textbf{51.09} & \textbf{52.58} & \textbf{54.07} & \textbf{55.03} & \textbf{55.53}\\
\midrule
Hybrid & 3-hop   &            & 27.37 & 28.89 & 31.46 & 33.37 & 34.68\\
       &         & \checkmark & \textbf{29.62} & \textbf{32.01} & \textbf{35.36} & \textbf{37} & \textbf{37.95}\\
\midrule
Hybrid & 4-hop   &            & \textbf{19.06} & \textbf{20.71} & \textbf{21.66} & \textbf{23.1}  & \textbf{24.51} \\
       &         & \checkmark & 18.82 & 19.76 & 21 & 22.47 & 24.24\\
\bottomrule
\end{tabular}
\end{adjustbox}
\label{tab:retrieval_eval_ndcg}
\end{table}

\section{Raw Experiment Results for Static NIAH with Retrieval-Ranked Haystack Ordering}

\label{appendix:static_NIAH_rank_order}

\begin{itemize}[leftmargin=*]
    \item BM25: Table~\ref{tab:static_NIAH_rank_bm25}
    \item Qwen3-0.6B: Table~\ref{tab:static_NIAH_rank_qwen3_0.6B}
    \item Hybrid: Table~\ref{tab:static_NIAH_rank_hybrid}
    \item BM25 + PPR: Table~\ref{tab:static_NIAH_rank_bm25_ppr}
    \item Qwen3-0.6B + PPR: Table~\ref{tab:static_NIAH_rank_qwen3_0.6B_ppr}
    \item Hybrid + PPR: Table~\ref{tab:static_NIAH_rank_hybrid_ppr}
\end{itemize}

\begin{table}[t]
\caption{Static NIAH performance in F1 score $(\times 10^{-2}, \uparrow)$ using BM25 for haystack construction, where retriever-ranked haystack ordering is used. 0 stands for the case without distractors.}
\centering
\begin{adjustbox}{width=0.8\textwidth}
\begin{tabular}{lcccccc}
\toprule
Context Size (\# Tokens) & 0 & 8K & 16K & 32K & 64K & 128K \\
\midrule
    Llama-3.1-8B-Instruct & \heatmapcell{59.8}{59.8} &
    \heatmapcell{58.41}{58.41} &
    \heatmapcell{53.81}{53.81} &
    \heatmapcell{46.16}{46.16} & 
    \heatmapcell{42.06}{42.06} 
    & \heatmapcell{37.24}{37.24} 
    \\
    Llama-3.1-70B-Instruct & 
    \heatmapcell{67.7}{67.7} &
    \heatmapcell{66.28}{66.28} &
    \heatmapcell{62.56}{62.56} &
    \heatmapcell{57.72}{57.72} &
    \heatmapcell{48.2}{48.2} &
    \heatmapcell{30.71}{30.71} \\
    Qwen2.5-7B-Instruct-1M & \heatmapcell{55.56}{55.56} &
    \heatmapcell{52.12}{52.12} &
    \heatmapcell{48.42}{48.42} &
    \heatmapcell{44.39}{44.39} &
    \heatmapcell{42.38}{42.38} 
    & \heatmapcell{39.5}{39.5} 
    \\
    Qwen2.5-14B-Instruct-1M & 
    \heatmapcell{61.76}{61.76} &
    \heatmapcell{59.46}{59.46} &
    \heatmapcell{55.28}{55.28} &
    \heatmapcell{51.3}{51.3} &
    \heatmapcell{46.65}{46.65} &
    \heatmapcell{45.87}{45.87} \\
    Gemma-3-12B-IT & 
    \heatmapcell{67.49}{67.49} &
    \heatmapcell{64.15}{64.15} &
    \heatmapcell{59.54}{59.54} &
    \heatmapcell{56.47}{56.47} &
    \heatmapcell{52.4}{52.4} &
    \heatmapcell{44.45}{44.45} \\
    Gemma-3-27B-IT & 
    \heatmapcell{67.71}{67.71} &
    \heatmapcell{66.24}{66.24} & 
    \heatmapcell{62.25}{62.25} &
    \heatmapcell{58.05}{58.05} &
    \heatmapcell{54.34}{54.34} &
    \heatmapcell{51.68}{51.68} \\
    GPT-4.1 mini & 
    \heatmapcell{70.19}{70.19} &
    \heatmapcell{69.05}{69.05} &
    \heatmapcell{67.36}{67.36} &
    \heatmapcell{64.28}{64.28} &
    \heatmapcell{60.73}{60.73} &
    \heatmapcell{60.55}{60.55} \\
    Qwen3-8B & 
    \heatmapcell{71.66}{71.66} & 
    \heatmapcell{71.62}{71.62} &
    \heatmapcell{69.84}{69.84} & 
    \heatmapcell{67.94}{67.94} &
    \heatmapcell{61.72}{61.72} &
    \heatmapcell{60.1}{60.1} \\
    Qwen3-14B &
    \heatmapcell{71.9}{71.9} & 
    \heatmapcell{71.08}{71.08} &
    \heatmapcell{69.5}{69.5} &
    \heatmapcell{67.7}{67.7} &
    \heatmapcell{64.74}{64.74} 
    & 
    \heatmapcell{62.42}{62.42}\\
    Qwen3-32B & 
    \heatmapcell{71.32}{71.32} & 
    \heatmapcell{73.62}{73.62} &
    \heatmapcell{69.97}{69.97} &
    \heatmapcell{68.59}{68.59} &
    \heatmapcell{67.15}{67.15} & 
    \heatmapcell{64.87}{64.87} \\
    Gemini 2.5 Flash-Lite & 
    \heatmapcell{71.6}{71.6} &
    \heatmapcell{70.96}{70.96} &
    \heatmapcell{69.94}{69.94} & \heatmapcell{69.36}{69.36} &
    \heatmapcell{68.93}{68.93} &
    \heatmapcell{66.14}{66.14} \\
    o4-mini & 
    \heatmapcell{75.95}{75.95} &
    \heatmapcell{76.34}{76.34} &
    \heatmapcell{74.24}{74.24} &
    \heatmapcell{74.02}{74.02} &
    \heatmapcell{70.36}{70.36} &
    \heatmapcell{67.58}{67.58} \\
\bottomrule
\end{tabular}
\end{adjustbox}
\label{tab:static_NIAH_rank_bm25}
\end{table}

\begin{table}[t]
\caption{Static NIAH performance in F1 score $(\times 10^{-2}, \uparrow)$ using Qwen3-0.6B for haystack construction, where retriever-ranked haystack ordering is used. 0 stands for the case without distractors.}
\centering
\begin{adjustbox}{width=0.8\textwidth}
\begin{tabular}{lcccccc}
\toprule
Context Size (\# Tokens) & 0 & 8K & 16K & 32K & 64K & 128K \\
\midrule
    Llama-3.1-8B-Instruct & \heatmapcell{59.8}{59.8} & \heatmapcell{57.22}{57.22} & \heatmapcell{52.92}{52.92} & \heatmapcell{44.74}{44.74} & \heatmapcell{40.42}{40.42} & \heatmapcell{31.51}{31.51} \\
    Llama-3.1-70B-Instruct & 
    \heatmapcell{67.7}{67.7} &
    \heatmapcell{64.36}{64.36} &
    \heatmapcell{61.6}{61.6} &
    \heatmapcell{54.42}{54.42} &
    \heatmapcell{45.73}{45.73} &
    \heatmapcell{25.85}{25.85} \\
    Qwen2.5-7B-Instruct-1M & \heatmapcell{55.56}{55.56} & 
    \heatmapcell{52.59}{52.59} & 
    \heatmapcell{47.04}{47.04} & 
    \heatmapcell{42.9}{42.9} & 
    \heatmapcell{39.23}{39.23} & 
    \heatmapcell{36.14}{36.14}  
     \\
    Qwen2.5-14B-Instruct-1M & \heatmapcell{61.76}{61.76} &
    \heatmapcell{56.4}{56.4} &
    \heatmapcell{51.66}{51.66} &
    \heatmapcell{49.61}{49.61} &
    \heatmapcell{45.8}{45.8} &
    \heatmapcell{42.39}{42.39} \\
    Gemma-3-12B-IT & 
    \heatmapcell{67.49}{67.49} &
    \heatmapcell{60.75}{60.75} &
    \heatmapcell{57.24}{57.24} &
    \heatmapcell{52.01}{52.01} &
    \heatmapcell{48.21}{48.21} &
    \heatmapcell{42.43}{42.43} \\
    Gemma-3-27B-IT & 
    \heatmapcell{67.71}{67.71} &
    \heatmapcell{63.86}{63.86} &
    \heatmapcell{60.92}{60.92} &
    \heatmapcell{55.5}{55.5} &
    \heatmapcell{53.05}{53.05} &
    \heatmapcell{48.98}{48.98} \\
    GPT-4.1 mini & 
    \heatmapcell{70.19}{70.19} &
    \heatmapcell{67.6}{67.6} &
    \heatmapcell{66.49}{66.49} &
    \heatmapcell{62.73}{62.73} &
    \heatmapcell{61.33}{61.33} &
    \heatmapcell{60.05}{60.05} \\
    Qwen3-8B & 
    \heatmapcell{71.66}{71.66} & 
    \heatmapcell{70.41}{70.41} &
    \heatmapcell{68.15}{68.15} & 
    \heatmapcell{65.14}{65.14} &
    \heatmapcell{62.2}{62.2} &
    \heatmapcell{59.22}{59.22} \\
    Qwen3-14B &
    \heatmapcell{71.9}{71.9} & 
    \heatmapcell{69.29}{69.29} & 
    \heatmapcell{69.38}{69.38} &
    \heatmapcell{65.96}{65.96} &
    \heatmapcell{64.89}{64.89} &
    \heatmapcell{61.39}{61.39} \\
    Qwen3-32B & 
    \heatmapcell{71.32}{71.32} & 
    \heatmapcell{69.49}{69.49} &
    \heatmapcell{69.14}{69.14} &
    \heatmapcell{68.65}{68.65} &
    \heatmapcell{64.01}{64.01} &
    \heatmapcell{62.41}{62.41} \\
    Gemini 2.5 Flash-Lite & 
    \heatmapcell{71.6}{71.6} & 
    \heatmapcell{70.81}{70.81} & 
    \heatmapcell{70.69}{70.69} &
    \heatmapcell{70.27}{70.27} &
    \heatmapcell{64.72}{64.72} &
    \heatmapcell{63.49}{63.49} \\
    o4-mini & 
    \heatmapcell{75.95}{75.95} &
    \heatmapcell{75.28}{75.28} &
    \heatmapcell{73.42}{73.42} &
    \heatmapcell{73.73}{73.73} &
    \heatmapcell{69.62}{69.62} &
    \heatmapcell{66.98}{66.98} \\
\bottomrule
\end{tabular}
\end{adjustbox}
\label{tab:static_NIAH_rank_qwen3_0.6B}
\end{table}

\begin{table}[t]
\caption{Static NIAH performance in F1 score $(\times 10^{-2}, \uparrow)$ using hybrid retriever for haystack construction, where retriever-ranked haystack ordering is used. 0 stands for the case without distractors.}
\centering
\begin{adjustbox}{width=0.8\textwidth}
\begin{tabular}{lcccccc}
\toprule
Context Size (\# Tokens) & 0 & 8K & 16K & 32K & 64K & 128K \\
\midrule
    Llama-3.1-8B-Instruct & \heatmapcell{59.8}{59.8} & 
    \heatmapcell{56.71}{56.71} & 
    \heatmapcell{53.41}{53.41} &
    \heatmapcell{43.02}{43.02} &
    \heatmapcell{36.99}{36.99} &
    \heatmapcell{30.22}{30.22} \\
    Llama-3.1-70B-Instruct & 
    \heatmapcell{67.7}{67.7} &
    \heatmapcell{66.09}{66.09} &
    \heatmapcell{62.44}{62.44} &
    \heatmapcell{52.04}{52.04} &
    \heatmapcell{42.33}{42.33} &
    \heatmapcell{25.11}{25.11} \\
    Qwen2.5-7B-Instruct-1M & \heatmapcell{55.56}{55.56} & 
    \heatmapcell{52.54}{52.54} &
    \heatmapcell{48.42}{48.42} &
    \heatmapcell{42.16}{42.16} &
    \heatmapcell{37.5}{37.5} &
    \heatmapcell{35.16}{35.16} \\
    Qwen2.5-14B-Instruct-1M & \heatmapcell{61.76}{61.76} &
    \heatmapcell{57.17}{57.17} &
    \heatmapcell{52.48}{52.48} & 
    \heatmapcell{50.7}{50.7} &
    \heatmapcell{46.24}{46.24} &
    \heatmapcell{42.91}{42.91} \\
    Gemma-3-12B-IT & 
    \heatmapcell{67.49}{67.49} & 
    \heatmapcell{62.26}{62.26} &
    \heatmapcell{56.63}{56.63} &
    \heatmapcell{53.68}{53.68} &
    \heatmapcell{48.62}{48.62} &
    \heatmapcell{44.1}{44.1} \\
    Gemma-3-27B-IT & 
    \heatmapcell{67.71}{67.71} &
    \heatmapcell{65.79}{65.79} &
    \heatmapcell{60.7}{60.7} & 
    \heatmapcell{54.78}{54.78} & 
    \heatmapcell{50.73}{50.73} &
    \heatmapcell{48.4}{48.4} \\
    GPT-4.1 mini & 
    \heatmapcell{70.19}{70.19} & 
    \heatmapcell{68.56}{68.56} &
    \heatmapcell{66.02}{66.02} &
    \heatmapcell{64.42}{64.42} &
    \heatmapcell{60.72}{60.72} &
    \heatmapcell{58.27}{58.27} \\
    Qwen3-8B & 
    \heatmapcell{71.66}{71.66} & 
    \heatmapcell{71.36}{71.36} & 
    \heatmapcell{68.74}{68.74} &
    \heatmapcell{66.52}{66.52} & 
    \heatmapcell{61.54}{61.54} & 
    \heatmapcell{57.29}{57.29} \\
    Qwen3-14B &
    \heatmapcell{71.9}{71.9} & 
    \heatmapcell{70.46}{70.46} &
    \heatmapcell{68.82}{68.82} &
    \heatmapcell{67.29}{67.29} &
    \heatmapcell{64.42}{64.42} &
    \heatmapcell{61.28}{61.28} \\
    Qwen3-32B & 
    \heatmapcell{71.32}{71.32} &
    \heatmapcell{71.19}{71.19} & 
    \heatmapcell{69.98}{69.98} & 
    \heatmapcell{67.55}{67.55} & 
    \heatmapcell{64.66}{64.66} &
    \heatmapcell{62.07}{62.07} \\
    Gemini 2.5 Flash-Lite & 
    \heatmapcell{71.6}{$71.6$} & 
    \heatmapcell{70.78}{$70.78$} &
    \heatmapcell{69.08}{$69.08$} &
    \heatmapcell{67.69}{$67.69$} &
    \heatmapcell{64.83}{$64.83$} &
    \heatmapcell{62.78}{$62.78$} \\
    o4-mini & 
    \heatmapcell{75.95}{$75.95$} & 
    \heatmapcell{76.43}{$76.43$} &
    \heatmapcell{73.54}{$73.54$} &
    \heatmapcell{73.12}{$73.12$} &
    \heatmapcell{68.46}{$68.46$} &
    \heatmapcell{67.9}{$67.9$} \\
\bottomrule
\end{tabular}
\end{adjustbox}
\label{tab:static_NIAH_rank_hybrid}
\end{table}

\begin{table}[t]
\caption{Static NIAH performance in F1 score $(\times 10^{-2}, \uparrow)$ using BM25 + PPR for haystack construction, where retriever-ranked haystack ordering is used. 0 stands for the case without distractors.}
\centering
\begin{adjustbox}{width=0.8\textwidth}
\begin{tabular}{lcccccc}
\toprule
Context Size (\# Tokens) & 0 & 8K & 16K & 32K & 64K & 128K \\
\midrule
    Llama-3.1-8B-Instruct & \heatmapcell{59.8}{59.8} & \heatmapcell{60.52}{60.52} & \heatmapcell{53.63}{53.63} & \heatmapcell{49.81}{49.81} & \heatmapcell{45.87}{45.87} & \heatmapcell{42.8}{42.8}\\
    Llama-3.1-70B-Instruct & 
    \heatmapcell{67.7}{67.7} &
    \heatmapcell{66.49}{66.49} &
    \heatmapcell{64.14}{64.14} &
    \heatmapcell{60.66}{60.66} &
    \heatmapcell{51.26}{51.26} & 
    \heatmapcell{37.08}{37.08} \\
    Qwen2.5-7B-Instruct-1M & \heatmapcell{55.56}{55.56} & \heatmapcell{51.69}{51.69} & \heatmapcell{49.6}{49.6} & \heatmapcell{46.16}{46.16} & \heatmapcell{46.91}{46.91} & \heatmapcell{45.95}{45.95} \\
    Qwen2.5-14B-Instruct-1M & 
    \heatmapcell{61.76}{61.76} &
    \heatmapcell{58.17}{58.17} &
    \heatmapcell{56.17}{56.17} &
    \heatmapcell{54.61}{54.61} &
    \heatmapcell{51.68}{51.68} &
    \heatmapcell{49.78}{49.78} \\
    Gemma-3-12B-IT & 
    \heatmapcell{67.49}{67.49} &
    \heatmapcell{64.11}{64.11} &
    \heatmapcell{60.85}{60.85} &
    \heatmapcell{58.82}{58.82} &
    \heatmapcell{54.43}{54.43} &
    \heatmapcell{47.89}{47.89} \\
    Gemma-3-27B-IT &
    \heatmapcell{67.71}{67.71} & 
    \heatmapcell{65.98}{65.98} & 
    \heatmapcell{63.08}{63.08} &
    \heatmapcell{58.13}{58.13} &
    \heatmapcell{56.23}{56.23} &
    \heatmapcell{53.71}{53.71} \\
    GPT-4.1 mini & 
    \heatmapcell{70.19}{70.19} & 
    \heatmapcell{69.07}{69.07} &
    \heatmapcell{66.67}{66.67} &
    \heatmapcell{64.46}{64.46} &
    \heatmapcell{63.36}{63.36} &
    \heatmapcell{62.2}{62.2} \\
    Qwen3-8B & 
    \heatmapcell{71.66}{71.66} & \heatmapcell{71.32}{71.32} &
    \heatmapcell{71.23}{71.23} & \heatmapcell{69.38}{69.38} &
    \heatmapcell{64.44}{64.44} &
    \heatmapcell{62.37}{62.37} \\
    Qwen3-14B &
    \heatmapcell{71.9}{71.9} & \heatmapcell{72.62}{72.62} & \heatmapcell{71.33}{71.33} &
    \heatmapcell{70.19}{70.19} &
    \heatmapcell{66.31}{66.31} &
    \heatmapcell{66.17}{66.17} \\
    Qwen3-32B & 
    \heatmapcell{71.32}{71.32} &
    \heatmapcell{71.58}{71.58} & 
    \heatmapcell{71.37}{71.37} &
    \heatmapcell{69.13}{69.13} &
    \heatmapcell{67.93}{67.93} & 
    \heatmapcell{65.7}{65.7} \\
    Gemini 2.5 Flash-Lite & 
    \heatmapcell{71.6}{71.6} &
    \heatmapcell{71.95}{71.95} &
    \heatmapcell{71.72}{71.72} &
    \heatmapcell{69.52}{69.52} &
    \heatmapcell{69.18}{69.18} &
    \heatmapcell{64.86}{64.86} \\
    o4-mini & 
    \heatmapcell{75.95}{75.95} &
    \heatmapcell{75.42}{75.42} &
    \heatmapcell{74.82}{74.82} &
    \heatmapcell{74.67}{74.67} &
    \heatmapcell{72.88}{72.88} &
    \heatmapcell{69.88}{69.88} \\
\bottomrule
\end{tabular}
\end{adjustbox}
\label{tab:static_NIAH_rank_bm25_ppr}
\end{table}

\begin{table}[t]
\caption{Static NIAH performance in F1 score $(\times 10^{-2}, \uparrow)$ using Qwen3-0.6B + PPR for haystack construction, where retriever-ranked haystack ordering is used. 0 stands for the case without distractors.}
\centering
\begin{adjustbox}{width=0.8\textwidth}
\begin{tabular}{lcccccc}
\toprule
Context Size (\# Tokens) & 0 & 8K & 16K & 32K & 64K & 128K \\
\midrule
    Llama-3.1-8B-Instruct & \heatmapcell{59.8}{59.8} & \heatmapcell{55.85}{55.85} & \heatmapcell{54.26}{54.26} & \heatmapcell{45.25}{45.25} & \heatmapcell{41.37}{41.37} & \heatmapcell{36.69}{36.69} \\
    Llama-3.1-70B-Instruct & 
    \heatmapcell{67.7}{67.7} &
    \heatmapcell{65.75}{65.75} &
    \heatmapcell{63.01}{63.01} &
    \heatmapcell{55.82}{55.82} &
    \heatmapcell{48.65}{48.65} &
    \heatmapcell{33.42}{33.42} \\
    Qwen2.5-7B-Instruct-1M & \heatmapcell{55.56}{55.56} & \heatmapcell{52.51}{52.51} & \heatmapcell{50.93}{50.93} & \heatmapcell{46.47}{46.47} & \heatmapcell{42.88}{42.88} & \heatmapcell{42.96}{42.96} \\
    Qwen2.5-14B-Instruct-1M & \heatmapcell{61.76}{61.76} &
    \heatmapcell{57.95}{57.95} &
    \heatmapcell{52.83}{52.83} &
    \heatmapcell{54.41}{54.41} &
    \heatmapcell{50.96}{50.96} &
    \heatmapcell{48.82}{48.82} \\
    Gemma-3-12B-IT & 
    \heatmapcell{67.49}{67.49} &
    \heatmapcell{64.25}{64.25} &
    \heatmapcell{59.27}{59.27} &
    \heatmapcell{56.33}{56.33} &
    \heatmapcell{53.72}{53.72} &
    \heatmapcell{47.2}{47.2} \\
    Gemma-3-27B-IT & 
    \heatmapcell{67.71}{67.71} & 
    \heatmapcell{65.08}{65.08} &
    \heatmapcell{61.46}{61.46} &
    \heatmapcell{59.64}{59.64} &
    \heatmapcell{57.39}{57.39} &
    \heatmapcell{50.93}{50.93} \\
    GPT-4.1 mini & 
    \heatmapcell{70.19}{70.19} &
    \heatmapcell{67.64}{67.64} &
    \heatmapcell{65.43}{65.43} &
    \heatmapcell{64.74}{64.74} &
    \heatmapcell{61.92}{61.92} &
    \heatmapcell{61.45}{61.45} \\
    Qwen3-8B & 
    \heatmapcell{71.66}{71.66} & 
    \heatmapcell{71.35}{71.35} &
    \heatmapcell{70.08}{70.08} &
    \heatmapcell{66.23}{66.23} &
    \heatmapcell{62.68}{62.68} &
    \heatmapcell{60.94}{60.94} \\
    Qwen3-14B &
    \heatmapcell{71.9}{71.9} & 
    \heatmapcell{69.97}{69.97} &
    \heatmapcell{69.46}{69.46} &
    \heatmapcell{68.39}{68.39} &
    \heatmapcell{65.33}{65.33} &
    \heatmapcell{62.42}{62.42} \\
    Qwen3-32B & 
    \heatmapcell{71.32}{71.32} &
    \heatmapcell{71.11}{71.11} &
    \heatmapcell{70.8}{70.8} & 
    \heatmapcell{70.28}{70.28} & 
    \heatmapcell{68.26}{68.26} & 
    \heatmapcell{64.55}{64.55} \\
    Gemini 2.5 Flash-Lite & 
    \heatmapcell{71.6}{71.6} &
    \heatmapcell{71.48}{71.48} &
    \heatmapcell{70.43}{70.43} & 
    \heatmapcell{70.33}{70.33} &
    \heatmapcell{66.05}{66.05} & 
    \heatmapcell{66.53}{66.53} \\
    o4-mini & 
    \heatmapcell{75.95}{75.95} & \heatmapcell{73.73}{73.73} &
    \heatmapcell{74.18}{74.18} &
    \heatmapcell{72.7}{72.7} & 
    \heatmapcell{70.24}{70.24} &
    \heatmapcell{69.03}{69.03} \\
\bottomrule
\end{tabular}
\end{adjustbox}
\label{tab:static_NIAH_rank_qwen3_0.6B_ppr}
\end{table}

\begin{table}[t]
\caption{Static NIAH performance in F1 score $(\times 10^{-2}, \uparrow)$ using hybrid + PPR retriever for haystack construction, where retriever-ranked haystack ordering is used. 0 stands for the case without distractors.}
\centering
\begin{adjustbox}{width=0.8\textwidth}
\begin{tabular}{lcccccc}
\toprule
Context Size (\# Tokens) & 0 & 8K & 16K & 32K & 64K & 128K \\
\midrule
    Llama-3.1-8B-Instruct & \heatmapcell{59.8}{59.8} & 
    \heatmapcell{55.04}{55.04} &
    \heatmapcell{50.45}{50.45} &
    \heatmapcell{46.64}{46.64} &
    \heatmapcell{45.41}{45.41} &
    \heatmapcell{38.11}{38.11} \\
    Llama-3.1-70B-Instruct & 
    \heatmapcell{67.7}{67.7} &
    \heatmapcell{65.85}{65.85} &
    \heatmapcell{61.56}{61.56} &
    \heatmapcell{56.6}{56.6} &
    \heatmapcell{48.64}{48.64} &
    \heatmapcell{36.22}{36.22} \\
    Qwen2.5-7B-Instruct-1M & \heatmapcell{55.56}{55.56} &
    \heatmapcell{52.06}{52.06} &
    \heatmapcell{49.07}{49.07} &
    \heatmapcell{46.02}{46.02} &
    \heatmapcell{44.31}{44.31} &
    \heatmapcell{45.65}{45.65} \\
    Qwen2.5-14B-Instruct-1M & \heatmapcell{61.76}{61.76} &
    \heatmapcell{57.92}{57.92} &
    \heatmapcell{52.93}{52.93} &
    \heatmapcell{51.29}{51.29} &
    \heatmapcell{51.14}{51.14} &
    \heatmapcell{49.88}{49.88} \\
    Gemma-3-12B-IT & 
    \heatmapcell{67.49}{67.49} & 
    \heatmapcell{62.95}{62.95} &
    \heatmapcell{58.43}{58.43} &
    \heatmapcell{57.35}{57.35} &
    \heatmapcell{54}{54} &
    \heatmapcell{48.8}{48.8} \\
    Gemma-3-27B-IT & 
    \heatmapcell{67.71}{67.71} & 
    \heatmapcell{65.45}{65.45} &
    \heatmapcell{60.97}{60.97} &
    \heatmapcell{59.56}{59.56} &
    \heatmapcell{57.55}{57.55} &
    \heatmapcell{52.51}{52.51} \\
    GPT-4.1 mini & 
    \heatmapcell{70.19}{70.19} & 
    \heatmapcell{69.48}{69.48} &
    \heatmapcell{66.63}{66.63} &
    \heatmapcell{65.23}{65.23} &
    \heatmapcell{64.73}{64.73} &
    \heatmapcell{62.09}{62.09} \\
    Qwen3-8B & 
    \heatmapcell{71.66}{71.66} & \heatmapcell{71.37}{71.37} &
    \heatmapcell{69.22}{69.22} &
    \heatmapcell{68.45}{68.45} & \heatmapcell{63.59}{63.59} &
    \heatmapcell{63.98}{63.98} \\
    Qwen3-14B &
    \heatmapcell{71.9}{71.9} & \heatmapcell{70.32}{70.32} &
    \heatmapcell{69}{69} &
    \heatmapcell{67.97}{67.97} &
    \heatmapcell{67}{67} &
    \heatmapcell{63.85}{63.85} \\
    Qwen3-32B & 
    \heatmapcell{71.32}{71.32} &
    \heatmapcell{71.11}{71.11} &
    \heatmapcell{70.93}{70.93} & 
    \heatmapcell{69.29}{69.29} & 
    \heatmapcell{65.8}{65.8} & 
    \heatmapcell{64.14}{64.14} \\
    Gemini 2.5 Flash-Lite & 
    \heatmapcell{71.6}{$71.6$} & 
    \heatmapcell{71.86}{$71.86$} &
    \heatmapcell{68.79}{$68.79$} &
    \heatmapcell{69.51}{$69.51$} &
    \heatmapcell{68.07}{$68.07$} &
    \heatmapcell{66.07}{$66.07$} \\
    o4-mini & 
    \heatmapcell{75.95}{$75.95$} & 
    \heatmapcell{75.1}{$75.1$} &
    \heatmapcell{73.7}{$73.7$} &
    \heatmapcell{73.61}{$73.61$} &
    \heatmapcell{70.23}{$70.23$} &
    \heatmapcell{70.06}{$70.06$} \\
\bottomrule
\end{tabular}
\end{adjustbox}
\label{tab:static_NIAH_rank_hybrid_ppr}
\end{table}

\section{Raw Experiment Results for Static NIAH Averaged Over Three Random Haystack Orderings}

\label{appendix:static_NIAH_random}

\begin{itemize}[leftmargin=*]
    \item BM25: Table~\ref{tab:static_NIAH_random_bm25}
    \item Qwen3-0.6B: Table~\ref{tab:static_NIAH_random_qwen3_0.6B}
    \item Hybrid: Table~\ref{tab:static_NIAH_random_hybrid}
    \item BM25 + PPR: Table~\ref{tab:static_NIAH_random_bm25_ppr}
    \item Qwen3-0.6B + PPR: Table~\ref{tab:static_NIAH_random_qwen3_0.6B_ppr}
    \item Hybrid + PPR: Table~\ref{tab:static_NIAH_random_hybrid_ppr}
\end{itemize}

\begin{table}[t]
\caption{Static NIAH performance in F1 score $(\times 10^{-2}, \uparrow)$ using BM25 for haystack construction, where we average the results over three random Haystack orderings. 0 stands for the case without distractors.}
\centering
\begin{adjustbox}{width=\textwidth}
\begin{tabular}{lcccccc}
\toprule
Context Size (\# Tokens) & 0 & 8K & 16K & 32K & 64K & 128K \\
\midrule
    Llama-3.1-8B-Instruct & \heatmapcell{59.8}{59.8} &
    \heatmapcell{56.18}{$56.18 \pm 0.85$} &
    \heatmapcell{52.25}{$52.25 \pm 1.21$} &
    \heatmapcell{44.71}{$44.71 \pm 0.86$} & 
    \heatmapcell{40.43}{$40.43 \pm 1.78$}
    & \heatmapcell{34.3}{$34.3 \pm 0.76$} 
    \\
    Llama-3.1-70B-Instruct & 
    \heatmapcell{67.7}{67.7} &
    \heatmapcell{66.51}{$66.51 \pm 0.15$} &
    \heatmapcell{63.79}{$63.79 \pm 1.21$} &
    \heatmapcell{58.41}{$58.41 \pm 0.57$} &
    \heatmapcell{49.14}{$49.14 \pm 1$} &
    \heatmapcell{30.58}{$30.58 \pm 0.72$} \\
    Qwen2.5-7B-Instruct-1M & \heatmapcell{55.56}{55.56} &
    \heatmapcell{50.7}{$50.7 \pm 1.3$} &
    \heatmapcell{45.73}{$45.73 \pm 0.66$} & 
    \heatmapcell{39.22}{$39.22 \pm 0.17$}
    & 
    \heatmapcell{36.85}{$36.85 \pm 0.38$} 
    & \heatmapcell{33.66}{$33.66 \pm 0.47$} 
    \\
    Qwen2.5-14B-Instruct-1M & 
    \heatmapcell{61.76}{61.76} &
    \heatmapcell{58.32}{$58.32 \pm 0.73$} &
    \heatmapcell{53.95}{$53.95 \pm 1.4$} &
    \heatmapcell{48.86}{$48.86 \pm 0.75$} &
    \heatmapcell{44.16}{$44.16 \pm 0.7$} &
    \heatmapcell{37.7}{$37.7 \pm 0.42$} \\
    Gemma-3-12B-IT & 
    \heatmapcell{67.49}{67.49} & 
    \heatmapcell{62.66}{$62.66 \pm 0.97$} &
    \heatmapcell{57.18}{$57.18 \pm 0.78$} &
    \heatmapcell{52.39}{$52.39 \pm 1.69$} &
    \heatmapcell{45.88}{$45.88 \pm 1.31$} &
    \heatmapcell{33.39}{$33.39 \pm 0.42$} \\
    Gemma-3-27B-IT & 
    \heatmapcell{67.71}{67.71} & 
    \heatmapcell{63.17}{$63.17 \pm 0.65$} &
    \heatmapcell{60.53}{$60.53 \pm 0.26$} &
    \heatmapcell{53.19}{$53.19 \pm 1.44$} &
    \heatmapcell{47.35}{$47.35 \pm 0.94$} &
    \heatmapcell{38.93}{$38.93 \pm 1.21$} \\
    GPT-4.1 mini & 
    \heatmapcell{70.19}{70.19} & 
    \heatmapcell{69.04}{$69.04 \pm 0.27$} &
    \heatmapcell{67.41}{$67.41 \pm 0.26$} &
    \heatmapcell{64.57}{$64.57 \pm 0.86$} &
    \heatmapcell{60.28}{$60.28 \pm 0.4$} &
    \heatmapcell{56.84 }{$56.84 \pm 0.36$} \\
    Qwen3-8B & 
    \heatmapcell{71.66}{71.66} & 
    \heatmapcell{71.3}{$71.3 \pm 0.81$} &
    \heatmapcell{71.02}{$71.02 \pm 0.82$} & 
    \heatmapcell{67.13}{$67.13 \pm 0.39$} & 
    \heatmapcell{58.53}{$58.53 \pm 1.33$} &
    \heatmapcell{53.48}{$53.48 \pm 1.55$} \\
    Qwen3-14B &
    \heatmapcell{71.9}{71.9} & 
    \heatmapcell{71.39}{$71.39 \pm 0.43$} &
    \heatmapcell{69.8}{$69.8 \pm 1.18$} &
    \heatmapcell{68.39}{$68.39 \pm 0.69$} &
    \heatmapcell{66.22}{$66.22 \pm 0.41$} &
    \heatmapcell{61.1}{$61.1 \pm 0.62$} \\
    Qwen3-32B & 
    \heatmapcell{71.32}{71.32} & 
    \heatmapcell{71.73}{$71.73 \pm 0.46$} &
    \heatmapcell{71.22}{$71.22 \pm 0.9$} &
    \heatmapcell{68.94}{$68.94 \pm 1.1$} &
    \heatmapcell{67.71}{$67.71 \pm 0.61$} &
    \heatmapcell{62.11}{$62.11 \pm 1.81$} \\
    Gemini 2.5 Flash-Lite & 
    \heatmapcell{71.6}{71.6} & 
    \heatmapcell{72.16}{$72.16 \pm 0.89$} &
    \heatmapcell{72.07}{$72.07 \pm 0.73$} &
    \heatmapcell{69.52}{$69.52 \pm 1.35$} &
    \heatmapcell{67.66}{$67.66 \pm 0.19$} &
    \heatmapcell{65.61}{$65.61 \pm 0.34$} \\
    o4-mini & 
    \heatmapcell{75.95}{75.95} &
    \heatmapcell{75.6}{$75.6 \pm 0.41$} &
    \heatmapcell{74.73}{$74.73 \pm 0.43$} &
    \heatmapcell{73.39}{$73.39 \pm 0.2$} &
    \heatmapcell{70.48}{$70.48 \pm 1.09$} &
    \heatmapcell{67.24}{$67.24 \pm 0.57$} \\
\bottomrule
\end{tabular}
\end{adjustbox}
\label{tab:static_NIAH_random_bm25}
\end{table}

\begin{table}[t]
\caption{Static NIAH performance in F1 score $(\times 10^{-2}, \uparrow)$ using Qwen3-0.6B for haystack construction, where we average the results over three random Haystack orderings. 0 stands for the case without distractors.}
\centering
\begin{adjustbox}{width=\textwidth}
\begin{tabular}{lcccccc}
\toprule
Context Size (\# Tokens) & 0 & 8K & 16K & 32K & 64K & 128K \\
\midrule
    Llama-3.1-8B-Instruct & \heatmapcell{59.8}{$59.8$} & \heatmapcell{54.87}{$54.87 \pm 0.75$} & \heatmapcell{50.71}{$50.71 \pm 0.89$} & \heatmapcell{43.41}{$43.41 \pm 0.6$} & \heatmapcell{39.08}{$39.08 \pm 1.00$}
    & \heatmapcell{33.33}{$33.33 \pm 0.5$} \\
    Llama-3.1-70B-Instruct & 
    \heatmapcell{67.7}{67.7} &
    \heatmapcell{65.69}{$65.69 \pm 0.77$} &
    \heatmapcell{62.52}{$62.52 \pm 1.01$} &
    \heatmapcell{56.81}{$56.81 \pm 1.1$} &
    \heatmapcell{47.87}{$47.87 \pm 0.2$} &
    \heatmapcell{26.49}{$26.49 \pm 0.73$} \\
    Qwen2.5-7B-Instruct-1M 
    & \heatmapcell{55.56}{$55.56$} 
    & \heatmapcell{50.79}{$50.79 \pm 1.29$} 
    & \heatmapcell{46.75}{$46.75 \pm 0.85$} 
    & \heatmapcell{41.05}{$41.05 \pm 1.00$} 
    & \heatmapcell{34.85}{$34.85 \pm 0.50$} 
    & \heatmapcell{29.55}{$29.55 \pm 0.46$} \\
    Qwen2.5-14B-Instruct-1M & 
    \heatmapcell{61.76}{61.76} &
    \heatmapcell{57.1}{$57.1 \pm 0.39$} &
    \heatmapcell{52.83}{$52.83 \pm 1.81$} & \heatmapcell{48.7}{$48.7 \pm 0.22$} &
    \heatmapcell{40.78}{$40.78 \pm 0.88$} &
    \heatmapcell{35.36}{$35.36 \pm 0.28$} \\
    Gemma-3-12B-IT & 
    \heatmapcell{67.49}{67.49} &
    \heatmapcell{61.2}{$61.2 \pm 0.2$} &
    \heatmapcell{55.51}{$55.51 \pm 1.0$} &
    \heatmapcell{51.24}{$51.24 \pm 0.46$} &
    \heatmapcell{42.66}{$42.66 \pm 0.79$} &
    \heatmapcell{32.37}{$32.37 \pm 1.66$} \\
    Gemma-3-27B-IT & 
    \heatmapcell{67.71}{67.71} & 
    \heatmapcell{62.87}{$62.87 \pm 0.39$} &
    \heatmapcell{57.77}{$57.77 \pm 0.26$} &
    \heatmapcell{51.71}{$51.71 \pm 0.37$} &
    \heatmapcell{42.95}{$42.95 \pm 1.05$} &
    \heatmapcell{34.67}{$34.67 \pm 0.38$} \\
    GPT-4.1 mini & 
    \heatmapcell{70.19}{70.19} & 
    \heatmapcell{68.2}{$68.2 \pm 0.53$} &
    \heatmapcell{66.09}{$66.09 \pm 1.67$} &
    \heatmapcell{63.57}{$63.57 \pm 0.45$} &
    \heatmapcell{59.97}{$59.97 \pm 0.55$} &
    \heatmapcell{56.74}{$56.74 \pm 1.38$} \\
    Qwen3-8B & 
    \heatmapcell{71.66}{71.66} & 
    \heatmapcell{71.69}{$71.69 \pm 0.81$} &
    \heatmapcell{69.9}{$69.9 \pm 0.52$} &
    \heatmapcell{64.12}{$64.12 \pm 0.82$} &
    \heatmapcell{54.24}{$54.24 \pm 0.41$} &
    \heatmapcell{48.89}{$48.89 \pm 1.62$} \\
    Qwen3-14B &
    \heatmapcell{71.9}{71.9} & 
    \heatmapcell{69.83}{$69.83 \pm 0.98$} &
    \heatmapcell{69.04}{$69.04 \pm 1.14$} &
    \heatmapcell{68.03}{$68.03 \pm 0.13$} &
    \heatmapcell{64.21}{$64.21 \pm 0.52$} &
    \heatmapcell{56.77}{$56.77 \pm 1.45$} \\
    Qwen3-32B & 
    \heatmapcell{71.32}{71.32} &
    \heatmapcell{70.66}{$70.66 \pm 1.22$} &
    \heatmapcell{70.29}{$70.29 \pm 0.76$} &
    \heatmapcell{68.68}{$68.68 \pm 0.56$} & 
    \heatmapcell{64.65}{$64.65 \pm 0.39$} &
    \heatmapcell{57.11}{$57.11 \pm 1.22$} \\
    Gemini 2.5 Flash-Lite & 
    \heatmapcell{71.6}{71.6} & 
    \heatmapcell{71.39}{$71.39 \pm 0.49$} &
    \heatmapcell{70.61}{$70.61 \pm 0.68$} &
    \heatmapcell{69.67}{$69.67 \pm 0.87$} &
    \heatmapcell{67.39}{$67.39 \pm 0.35$} &
    \heatmapcell{62.58}{$62.58 \pm 0.64$} \\
    o4-mini & 
    \heatmapcell{75.95}{75.95} &
    \heatmapcell{74.54}{$74.54 \pm 0.17$} &
    \heatmapcell{75.33}{$75.33 \pm 0.43$} &
    \heatmapcell{72.93}{$72.93 \pm 0.44$} & 
    \heatmapcell{69.72}{$69.72 \pm 0.59$} &
    \heatmapcell{63.91}{$63.91 \pm 1.39$} \\
\bottomrule
\end{tabular}
\end{adjustbox}
\label{tab:static_NIAH_random_qwen3_0.6B}
\end{table}

\begin{table}[t]
\caption{Static NIAH performance in F1 score $(\times 10^{-2}, \uparrow)$ using hybrid retriever for haystack construction, where we average the results over three random Haystack orderings. 0 stands for the case without distractors.}
\centering
\begin{adjustbox}{width=\textwidth}
\begin{tabular}{lcccccc}
\toprule
Context Size (\# Tokens) & 0 & 8K & 16K & 32K & 64K & 128K \\
\midrule
    Llama-3.1-8B-Instruct & \heatmapcell{59.8}{$59.8$} & \heatmapcell{54.95}{$54.95 \pm 1.16$} & \heatmapcell{51.33}{$51.33 \pm 0.85$} & \heatmapcell{43.77}{$43.77 \pm 1.42$} & \heatmapcell{37.42}{$37.42 \pm 0.92$} & \heatmapcell{33.34}{$33.34 \pm 0.58$} \\
    Llama-3.1-70B-Instruct & 
    \heatmapcell{67.7}{67.7} & 
    \heatmapcell{66.29}{$66.29 \pm 0.19$} &
    \heatmapcell{62.54}{$62.54 \pm 0.25$} &
    \heatmapcell{57.51}{$57.51 \pm 0.05$} &
    \heatmapcell{47.76}{$47.76 \pm 1.77$} &
    \heatmapcell{28.82}{$28.82 \pm 1.01$} \\
    Qwen2.5-7B-Instruct-1M & \heatmapcell{55.56}{$55.56$} & \heatmapcell{52.46}{$52.46 \pm 1.21$} & \heatmapcell{46.7}{$46.7 \pm 1.14$} & \heatmapcell{40.32}{$40.32 \pm 0.78$} & \heatmapcell{35.26}{$35.26 \pm 1.37$} & \heatmapcell{32.52}{$32.52 \pm 0.61$} \\
    Qwen2.5-14B-Instruct-1M & \heatmapcell{61.76}{61.76} &
    \heatmapcell{58.28}{$58.28 \pm 0.29$} &
    \heatmapcell{52.69}{$52.69 \pm 0.46$} &
    \heatmapcell{47.98}{$47.98 \pm 0.8$} &
    \heatmapcell{42.9}{$42.9 \pm 0.71$} &
    \heatmapcell{36.16}{$36.16 \pm 1.04$} \\
    Gemma-3-12B-IT & 
    \heatmapcell{67.49}{67.49} &
    \heatmapcell{62.41}{$62.41 \pm 0.43$} &
    \heatmapcell{56.83}{$56.83 \pm 0.77$} &
    \heatmapcell{51.66}{$51.66 \pm 0.25$} & 
    \heatmapcell{44.11}{$44.11 \pm 0.63$} &
    \heatmapcell{32.51}{$32.51 \pm 1.72$} \\
    Gemma-3-27B-IT & 
    \heatmapcell{67.71}{67.71} &
    \heatmapcell{64.26}{$64.26 \pm 0.43$} &
    \heatmapcell{57.56}{$57.56 \pm 0.35$} &
    \heatmapcell{52.08}{$52.08 \pm 0.22$} &
    \heatmapcell{44.95}{$44.95 \pm 1.07$} &
    \heatmapcell{36.85}{$36.85 \pm 1.16$} \\
    GPT-4.1 mini & 
    \heatmapcell{70.19}{70.19} & 
    \heatmapcell{68.46}{$68.46 \pm 0.53$} &
    \heatmapcell{66.4}{$66.4 \pm 0.81$} &
    \heatmapcell{62.48}{$62.48 \pm 0.39$} &
    \heatmapcell{59.72}{$59.72 \pm 0.38$} &
    \heatmapcell{56.77}{$56.77 \pm 0.81$} \\
    Qwen3-8B & 
    \heatmapcell{71.66}{71.66} & 
    \heatmapcell{72.03}{$72.03 \pm 0.68$} & 
    \heatmapcell{69.57}{$69.57 \pm 0.42$} & 
    \heatmapcell{65.96}{$65.96 \pm 0.97$} & 
    \heatmapcell{57.68}{$57.68 \pm 0.65$} & 
    \heatmapcell{51.85}{$51.85 \pm 0.86$} \\
    Qwen3-14B &
    \heatmapcell{71.9}{71.9} &
    \heatmapcell{70.21}{$70.21 \pm 0.15$} &
    \heatmapcell{69.97}{$69.97 \pm 0.4$} &
    \heatmapcell{66.74}{$66.74 \pm 0.37$} &
    \heatmapcell{65.4}{$65.4 \pm 0.71$} &
    \heatmapcell{59.58}{$59.58 \pm 1.13$} \\
    Qwen3-32B & 
    \heatmapcell{71.32}{71.32} & 
    \heatmapcell{71.61}{$71.61 \pm 0.69$} & 
    \heatmapcell{70.66}{$70.66 \pm 0.84$} &
    \heatmapcell{68.19}{$68.19 \pm 1.14$} &
    \heatmapcell{63.68}{$63.68 \pm 0.59$} & 
    \heatmapcell{57.99}{$57.99 \pm 0.65$} \\
    Gemini 2.5 Flash-Lite & 
    \heatmapcell{71.6}{$71.6$} & 
    \heatmapcell{71.56}{$71.56 \pm 0.81$} &
    \heatmapcell{70.04}{$70.04 \pm 0.6$} &
    \heatmapcell{69.24}{$69.24 \pm 0.78$} &
    \heatmapcell{66.75}{$66.75 \pm 0.65$} &
    \heatmapcell{64.06}{$64.06 \pm 0.51$} \\
    o4-mini & 
    \heatmapcell{75.95}{$75.95$} & 
    \heatmapcell{74.67}{$74.67 \pm 0.32$} &
    \heatmapcell{74.73}{$74.73 \pm 1.23$} &
    \heatmapcell{72.88}{$72.88 \pm 0.38$} &
    \heatmapcell{70.22}{$70.22 \pm 0.18$} &
    \heatmapcell{65.64}{$65.64 \pm 0.2$} \\
\bottomrule
\end{tabular}
\end{adjustbox}
\label{tab:static_NIAH_random_hybrid}
\end{table}

\begin{table}[t]
\caption{Static NIAH performance in F1 score $(\times 10^{-2}, \uparrow)$ using BM25 + PPR for haystack construction, where we average the results over three random Haystack orderings. 0 stands for the case without distractors.}
\centering
\begin{adjustbox}{width=\textwidth}
\begin{tabular}{lcccccc}
\toprule
Context Size (\# Tokens) & 0 & 8K & 16K & 32K & 64K & 128K \\
\midrule
    Llama-3.1-8B-Instruct & \heatmapcell{59.8}{59.8} & \heatmapcell{56.76}{$56.76 \pm 0.34$} & \heatmapcell{51.34}{$51.34 \pm 0.64$} & \heatmapcell{45.13}{$45.13 \pm 1.19$} & \heatmapcell{41.23}{$41.23 \pm 1.16$} & \heatmapcell{37.44}{$37.44 \pm 2.2$}\\
    Llama-3.1-70B-Instruct & 
    \heatmapcell{67.7}{67.7} &
    \heatmapcell{66.83}{$66.83 \pm 0.41$} &
    \heatmapcell{63.39}{$63.39 \pm 0.96$} &
    \heatmapcell{59.46}{$59.46 \pm 0.95$} &
    \heatmapcell{50.4}{$50.4 \pm 1.22$} &
    \heatmapcell{33.53}{$33.53 \pm 1.22$} \\
    Qwen2.5-7B-Instruct-1M & \heatmapcell{55.56}{55.56} & \heatmapcell{52.14}{$52.14 \pm 0.51$} &
    \heatmapcell{45.37}{$45.37 \pm 1.58$} & 
    \heatmapcell{42.73}{$42.73 \pm 1.63$} & 
    \heatmapcell{37.8}{$37.8 \pm 1.88$} & 
    \heatmapcell{37.74}{$37.74 \pm 0.19$}\\
    Qwen2.5-14B-Instruct-1M & \heatmapcell{61.76}{61.76} &
    \heatmapcell{58.73}{$58.73 \pm 0.73$} &
    \heatmapcell{53.72}{$53.72 \pm 1.03$} &
    \heatmapcell{49.81}{$49.81 \pm 1.26$} &
    \heatmapcell{44.38}{$44.38 \pm 1.29$} &
    \heatmapcell{41.94}{$41.94 \pm 0.79$} \\
    Gemma-3-12B-IT & 
    \heatmapcell{67.49}{67.49} & 
    \heatmapcell{62.8}{$62.8 \pm 0.42$} &
    \heatmapcell{58.08}{$58.08 \pm 0.49$} &
    \heatmapcell{52.37}{$52.37 \pm 0.85$} &
    \heatmapcell{45.95}{$45.95 \pm 1.31$} &
    \heatmapcell{38.5}{$38.5 \pm 0.21$} \\
    Gemma-3-27B-IT & 
    \heatmapcell{67.71}{67.71} & 
    \heatmapcell{64.31}{$64.31 \pm 0.83$} &
    \heatmapcell{60.06}{$60.06 \pm 0.64$} &
    \heatmapcell{53.15}{$53.15 \pm 1.13$} & 
    \heatmapcell{49.49}{$49.49 \pm 0.64$} &
    \heatmapcell{42.65}{$42.65 \pm 0.94$} \\
    GPT-4.1 mini & 
    \heatmapcell{70.19}{70.19} &
    \heatmapcell{69.12}{$69.12 \pm 0.28$} &
    \heatmapcell{67.36}{$67.36 \pm 0.96$} &
    \heatmapcell{64.48}{$64.48 \pm 0.71$} &
    \heatmapcell{60.93}{$60.93 \pm 0.7$} &
    \heatmapcell{59.26}{$59.26 \pm 1.29$} \\
    Qwen3-8B & 
    \heatmapcell{71.66}{71.66} & \heatmapcell{71.71}{$71.71 \pm 0.35$} &
    \heatmapcell{70.6}{$70.6 \pm 0.78$} &
    \heatmapcell{66.74}{$66.74 \pm 0.81$} &
    \heatmapcell{60.06}{$60.06 \pm 0.4$} & 
    \heatmapcell{57.87}{$57.87 \pm 0.37$} \\
    Qwen3-14B &
    \heatmapcell{71.9}{71.9} & 
    \heatmapcell{70.93}{$70.93 \pm 0.11$} &
    \heatmapcell{70.24}{$70.24 \pm 0.81$} &
    \heatmapcell{68.22}{$68.22 \pm 0.51$} &
    \heatmapcell{66.72}{$66.72 \pm 1.13$} &
    \heatmapcell{62.8}{$62.8 \pm 0.65$} \\
    Qwen3-32B &
    \heatmapcell{71.32}{71.32} & 
    \heatmapcell{71.48}{$71.48 \pm 0.25$} & \heatmapcell{70.86}{$70.86 \pm 0.26$} & \heatmapcell{69.2}{$69.2 \pm 0.96$} &
    \heatmapcell{66.73}{$66.73 \pm 0.96$} &
    \heatmapcell{63.91}{$63.91 \pm 0.68$}  \\
    Gemini 2.5 Flash-Lite & 
    \heatmapcell{71.6}{71.6} & 
    \heatmapcell{72.66}{$72.66 \pm 1.48$} &
    \heatmapcell{72.69}{$72.69 \pm 0.66$} &
    \heatmapcell{69.43}{$69.43 \pm 0.96$} &
    \heatmapcell{66.61}{$66.61 \pm 0.52$} &
    \heatmapcell{65.64}{$65.64 \pm 0.26$} \\
    o4-mini & 
    \heatmapcell{75.95}{75.95} & 
    \heatmapcell{75.86}{$75.86 \pm 0.71$} &
    \heatmapcell{74.77}{$74.77 \pm 0.8$} &
    \heatmapcell{73.73}{$73.73 \pm 0.22$} &
    \heatmapcell{71.73}{$71.73 \pm 0.36$} &
    \heatmapcell{68.96}{$68.96 \pm 0.47$} \\
\bottomrule
\end{tabular}
\end{adjustbox}
\label{tab:static_NIAH_random_bm25_ppr}
\end{table}

\begin{table}[t]
\caption{Static NIAH performance in F1 score $(\times 10^{-2}, \uparrow)$ using Qwen3-0.6B + PPR for haystack construction, where we average the results over three random Haystack orderings. 0 stands for the case without distractors.}
\centering
\begin{adjustbox}{width=\textwidth}
\begin{tabular}{lcccccc}
\toprule
Context Size (\# Tokens) & 0 & 8K & 16K & 32K & 64K & 128K \\
\midrule
    Llama-3.1-8B-Instruct & \heatmapcell{59.8}{$59.8$} & \heatmapcell{55.22}{$55.22 \pm 1.19$} & \heatmapcell{51.4}{$51.4 \pm 1.12$} & \heatmapcell{46.64}{$46.64 \pm 0.94$} & \heatmapcell{41.6}{$41.6 \pm 1.04$} & \heatmapcell{36.71}{$36.71 \pm 0.6$} \\
    Llama-3.1-70B-Instruct & 
    \heatmapcell{67.7}{67.7} &
    \heatmapcell{66.49}{$66.49 \pm 0.78$} &
    \heatmapcell{62.85}{$62.85 \pm 0.07$} &
    \heatmapcell{57.63}{$57.63 \pm 0.3$} &
    \heatmapcell{50.4}{$50.4 \pm 0.95$} &
    \heatmapcell{33.35}{$33.35 \pm 0.65$} \\
    Qwen2.5-7B-Instruct-1M & \heatmapcell{55.56}{$55.56$} & \heatmapcell{55.56}{$51.54 \pm 0.68$} & \heatmapcell{47.34}{$47.34 \pm 1.24$} & \heatmapcell{44.77}{$44.77 \pm 0.48$} & \heatmapcell{41.14}{$41.14 \pm 1.06$} & \heatmapcell{36.05}{$36.05 \pm 1.55$} \\
    Qwen2.5-14B-Instruct-1M & \heatmapcell{61.76}{61.76} &
    \heatmapcell{58.45}{$58.45 \pm 0.29$} &
    \heatmapcell{53.54}{$53.54 \pm 0.49$} &
    \heatmapcell{49.4}{$49.4 \pm 1.65$} &
    \heatmapcell{44.76}{$44.76 \pm 1.43$} &
    \heatmapcell{39.46}{$39.46 \pm 1.15$} \\
    Gemma-3-12B-IT & 
    \heatmapcell{67.49}{67.49} &
    \heatmapcell{62.78}{$62.78 \pm 0.38$} &
    \heatmapcell{56.4}{$56.4 \pm 0.75$} &
    \heatmapcell{53.29}{$53.29 \pm 0.87$} &
    \heatmapcell{46.8}{$46.8 \pm 0.77$} &
    \heatmapcell{38.47}{$38.47 \pm 0.21$} \\
    Gemma-3-27B-IT & 
    \heatmapcell{67.71}{67.71} & 
    \heatmapcell{63.72}{$63.72 \pm 0.52$} &
    \heatmapcell{59.01}{$59.01 \pm 1.53$} &
    \heatmapcell{54.35}{$54.35 \pm 0.84$} &
    \heatmapcell{49.01}{$49.01 \pm 1.02$} &
    \heatmapcell{40.79}{$40.79 \pm 0.93$} \\
    GPT-4.1 mini & 
    \heatmapcell{70.19}{70.19} & 
    \heatmapcell{68.08}{$68.08 \pm 1.18$} &
    \heatmapcell{66.18}{$66.18 \pm 0.3$} &
    \heatmapcell{63.23}{$63.23 \pm 0.56$} &
    \heatmapcell{61.09}{$61.09 \pm 0.61$} &
    \heatmapcell{57.49}{$57.49 \pm 0.31$} \\
    Qwen3-8B & 
    \heatmapcell{71.66}{71.66} & 
    \heatmapcell{71.53}{$71.53 \pm 0.59$} &
    \heatmapcell{69.82}{$69.82 \pm 0.89$} &
    \heatmapcell{67.09}{$67.09 \pm 0.52$} &
    \heatmapcell{58.84}{$58.84 \pm 1.45$} &
    \heatmapcell{57.84}{$57.84 \pm 1.53$} \\
    Qwen3-14B &
    \heatmapcell{71.9}{71.9} & 
    \heatmapcell{70.46}{$70.46 \pm 1.0$} &
    \heatmapcell{70.19}{$70.19 \pm 0.77$} &
    \heatmapcell{67.61}{$67.61 \pm 0.1$} & 
    \heatmapcell{66.08}{$66.08 \pm 0.66$} &
    \heatmapcell{61.4}{$61.4 \pm 0.64$} \\
    Qwen3-32B & 
    \heatmapcell{71.32}{71.32} &
    \heatmapcell{70.63}{$70.63 \pm 0.57$} &
    \heatmapcell{70.19}{$70.19 \pm 0.22$} & 
    \heatmapcell{68.77}{$68.77 \pm 1.35$} &
    \heatmapcell{65.64}{$65.64 \pm 0.47$} &
    \heatmapcell{61.54}{$61.54 \pm 1.12$} \\
    Gemini 2.5 Flash-Lite & 
    \heatmapcell{71.6}{$71.6$} & 
    \heatmapcell{72.13}{$72.13 \pm 1.34$} &
    \heatmapcell{70.21}{$70.21 \pm 0.25$} &
    \heatmapcell{69.26}{$69.26 \pm 0.77$} &
    \heatmapcell{67.64}{$67.64 \pm 0.64$} &
    \heatmapcell{65.26}{$65.26 \pm 1.6$} \\
    o4-mini & 
    \heatmapcell{75.95}{$75.95$} &
    \heatmapcell{74.49}{$74.49 \pm 0.38$} &
    \heatmapcell{74.35}{$74.35 \pm 0.19$} &
    \heatmapcell{73.86}{$73.86 \pm 0.51$} &
    \heatmapcell{70.81}{$70.81 \pm 0.63$} &
    \heatmapcell{67.71}{$67.71 \pm 0.41$} \\
\bottomrule
\end{tabular}
\end{adjustbox}
\label{tab:static_NIAH_random_qwen3_0.6B_ppr}
\end{table}

\begin{table}[t]
\caption{Static NIAH performance in F1 score $(\times 10^{-2}, \uparrow)$ using hybrid + PPR retriever for haystack construction, where we average the results over three random Haystack orderings. 0 stands for the case without distractors.}
\centering
\begin{adjustbox}{width=\textwidth}
\begin{tabular}{lcccccc}
\toprule
Context Size (\# Tokens) & 0 & 8K & 16K & 32K & 64K & 128K \\
\midrule
    Llama-3.1-8B-Instruct & 
    \heatmapcell{59.8}{$59.8$} & 
    \heatmapcell{55.19}{$55.19 \pm 1.37$} & 
    \heatmapcell{51.69}{$51.69 \pm 0.71$} &
    \heatmapcell{45.57}{$45.57 \pm 1.02$} &
    \heatmapcell{41.97}{$41.97 \pm 0.54$} &
    \heatmapcell{38.99}{$38.99 \pm 1.78$} \\
    Llama-3.1-70B-Instruct & 
    \heatmapcell{67.7}{67.7} &
    \heatmapcell{66.97}{$66.97 \pm 0.78$} &
    \heatmapcell{62.81}{$62.81 \pm 1.17$} &
    \heatmapcell{58.75}{$58.75 \pm 0.6$} &
    \heatmapcell{51.36}{$51.36 \pm 0.61$} &
    \heatmapcell{33.16}{$33.16 \pm 1.12$} \\
    Qwen2.5-7B-Instruct-1M & 
    \heatmapcell{55.56}{$55.56$} & 
    \heatmapcell{52.21}{$52.21 \pm 1.02$} &
    \heatmapcell{47.8}{$47.8 \pm 0.19$} &
    \heatmapcell{44.18}{$44.18 \pm 0.38$} &
    \heatmapcell{40.59}{$40.59 \pm 0.49$} &
    \heatmapcell{39.17}{$39.17 \pm 1.27$} \\
    Qwen2.5-14B-Instruct-1M & \heatmapcell{61.76}{61.76} &
    \heatmapcell{58.03}{$58.03 \pm 0.55$} &
    \heatmapcell{53.63}{$53.63 \pm 0.79$} &
    \heatmapcell{50.8}{$50.8 \pm 0.69$} &
    \heatmapcell{46.9}{$46.9 \pm 1.25$} &
    \heatmapcell{42.15}{$42.15 \pm 0.7$} \\
    Gemma-3-12B-IT & 
    \heatmapcell{67.49}{67.49} & 
    \heatmapcell{62.2}{$62.2 \pm 0.98$} &
    \heatmapcell{57.27}{$57.27 \pm 1.04$} & 
    \heatmapcell{53.32}{$53.32 \pm 0.47$} &
    \heatmapcell{48.2}{$48.2 \pm 0.88$} &
    \heatmapcell{37.51}{$37.51 \pm 4.94$} \\
    Gemma-3-27B-IT & 
    \heatmapcell{67.71}{67.71} & 
    \heatmapcell{64.02}{$64.02 \pm 1.13$} &
    \heatmapcell{58.84}{$58.84 \pm 0.49$} & \heatmapcell{55.6}{$55.6 \pm 0.68$} &
    \heatmapcell{50.27}{$50.27 \pm 0.14$} &
    \heatmapcell{41.95}{$41.95 \pm 1.08$} \\
    GPT-4.1 mini & 
    \heatmapcell{70.19}{70.19} & 
    \heatmapcell{68.7}{$68.7 \pm 0.76$} &
    \heatmapcell{66.51}{$66.51 \pm 0.81$} &
    \heatmapcell{64.43}{$64.43 \pm 0.61$} &
    \heatmapcell{61.79}{$61.79 \pm 1.28$} &
    \heatmapcell{60.75}{$60.75 \pm 0.31$} \\
    Qwen3-8B & 
    \heatmapcell{71.66}{71.66} & \heatmapcell{71.5}{$71.5 \pm 0.71$} &
    \heatmapcell{69.7}{$69.7 \pm 0.32$} & 
    \heatmapcell{66.95}{$66.95 \pm 0.51$} &
    \heatmapcell{60.46}{$60.46 \pm 0.28$} &
    \heatmapcell{58.62}{$58.62 \pm 0.52$} \\
    Qwen3-14B &
    \heatmapcell{71.9}{71.9} & 
    \heatmapcell{70.14}{$70.14 \pm 0.32$} &
    \heatmapcell{68.83}{$68.83 \pm 0.74$} &
    \heatmapcell{66.87}{$66.87 \pm 2.12$} &
    \heatmapcell{66.81}{$66.81 \pm 0.78$} &
    \heatmapcell{62.13}{$62.13 \pm 0.47$} \\
    Qwen3-32B & 
    \heatmapcell{71.32}{71.32} &
    \heatmapcell{71.96}{$71.96 \pm 0.67$} &
    \heatmapcell{71.03}{$71.03 \pm 0.57$} &
    \heatmapcell{70.13}{$70.13 \pm 0.43$} & 
    \heatmapcell{65.37}{$65.37 \pm 0.55$} & 
    \heatmapcell{63.4}{$63.4 \pm 0.62$}  \\
    Gemini 2.5 Flash-Lite & 
    \heatmapcell{71.6}{$71.6$} &
    \heatmapcell{71.57}{$71.57 \pm 0.87$} &
    \heatmapcell{70.39}{$70.39 \pm 0.6$} &
    \heatmapcell{69.07}{$69.07 \pm 0.35$} &
    \heatmapcell{66.42}{$66.42 \pm 0.4$} &
    \heatmapcell{65.27}{$65.27 \pm 0.52$} \\
    o4-mini & 
    \heatmapcell{75.95}{$75.95$} &
    \heatmapcell{75.59}{$75.59 \pm 0.19$} &
    \heatmapcell{74.14}{$74.14 \pm 0.7$} &
    \heatmapcell{73.68}{$73.68 \pm 0.23$} &
    \heatmapcell{71.03}{$71.03 \pm 0.61$} &
    \heatmapcell{68.13}{$68.13 \pm 0.4$} \\
\bottomrule
\end{tabular}
\end{adjustbox}
\label{tab:static_NIAH_random_hybrid_ppr}
\end{table}

\section{Raw Experiment Results for Dynamic NIAH}

\label{appendix:dynamic_NIAH_raw}

See Table~\ref{tab:dynamic_NIAH_bm25_ppr} and Table~\ref{tab:dynamic_NIAH_qwen}.

\begin{table}[t]
\caption{Dynamic NIAH performance in F1 score $(\times 10^{-2}, \uparrow)$ using BM25 + PPR for haystack construction, where retriever-ranked haystack ordering is used. 0 stands for the case without distractors.}
\centering
\begin{adjustbox}{width=0.8\textwidth}
\begin{tabular}{lccccccc}
\toprule
Context Size (\# Tokens) & \# Rounds & 0 & 8K & 16K & 32K & 64K & 128K \\
\midrule
    Llama-3.1-8B-Instruct & 1 & 
    \heatmapcell{56.86}{$56.86$} & 
    \heatmapcell{60.46}{$60.46$} &
    \heatmapcell{52.06}{$52.06$} &
    \heatmapcell{51.93}{$51.93$} & 
    \heatmapcell{42.42}{$42.42$} &
    \heatmapcell{39.97}{$39.97$} \\
    & 2 & 
    \heatmapcell{50.88}{$50.88$} &
    \heatmapcell{49.52}{$49.52$} &
    \heatmapcell{51.41}{$51.41$} & 
    \heatmapcell{39.18}{$39.18$} &
    \heatmapcell{37.71}{$37.71$} & 
    \heatmapcell{27.88}{$27.88$} \\
    & 3 & 
    \heatmapcell{26.51}{$26.51$} & 
    \heatmapcell{33.79}{$33.79$} &
    \heatmapcell{37.04}{$37.04$} &
    \heatmapcell{28.63}{$28.63$} &
    \heatmapcell{23.79}{$23.79$} &
    \heatmapcell{32.77}{$32.77$} \\
    & max 3 & 
    \heatmapcell{34.52}{$34.52$} &
    \heatmapcell{43.16}{$43.16$} &
    \heatmapcell{34.85}{$34.85$} &
    \heatmapcell{31.62}{$31.62$} &
    \heatmapcell{30.45}{$30.45$} &
    \heatmapcell{29.66}{$29.66$} \\
    \midrule
    Qwen2.5-7B-Instruct-1M & 
    1 & 
    \heatmapcell{60.59}{$60.59$} &
    \heatmapcell{55.16}{$55.16$} & 
    \heatmapcell{52.63}{$52.63$} & 
    \heatmapcell{45.96}{$45.96$} &
    \heatmapcell{46.3}{$46.3$} & 
    \heatmapcell{48.12}{$48.12$} \\
    & 2 & 
    \heatmapcell{37.64}{$37.64$} &
    \heatmapcell{46.31}{$46.31$} & 
    \heatmapcell{37.53}{$37.53$} & 
    \heatmapcell{28.57}{$28.57$} &
    \heatmapcell{34.78}{$34.78$} & 
    \heatmapcell{36.8}{$36.8$} \\
    & 3 & 
    \heatmapcell{24}{$24$} &
    \heatmapcell{25.35}{$25.35$} &
    \heatmapcell{28.14}{$28.14$} &
    \heatmapcell{22.88}{$22.88$} &
    \heatmapcell{20.98}{$20.98$} &
    \heatmapcell{21.93}{$21.93$} \\
    & max 3 & 
    \heatmapcell{43.61}{$43.61$} & 
    \heatmapcell{38.99}{$38.99$} &
    \heatmapcell{35.95}{$35.95$} &
    \heatmapcell{36.51}{$36.51$} &
    \heatmapcell{34.5}{$34.5$} &
    \heatmapcell{34.97}{$34.97$} \\
    \midrule 
    Gemma-3-12B-IT & 1 &
    \heatmapcell{66.54}{$66.54$} &
    \heatmapcell{63.48}{$63.48$} &
    \heatmapcell{61.94}{$61.94$} &
    \heatmapcell{57.27}{$57.27$} &
    \heatmapcell{54.64}{$54.64$} & \heatmapcell{45.51}{$45.51$} 
    \\
    & 2 & 
    \heatmapcell{48.91}{$48.91$} & 
    \heatmapcell{47.96}{$47.96$} & 
    \heatmapcell{53.82}{$53.82$} &
    \heatmapcell{40.63}{$40.63$} & 
    \heatmapcell{41.77}{$41.77$} &
    \heatmapcell{36.3}{$36.3$} \\
    & 3 & 
    \heatmapcell{39.07}{$39.07$} & 
    \heatmapcell{38.94}{$38.94$} &
    \heatmapcell{40.31}{$40.31$} & 
    \heatmapcell{32.48}{$32.48$} & 
    \heatmapcell{35.18}{$35.18$} &
    \heatmapcell{29.28}{$29.28$} \\
    & max 3 & 
    \heatmapcell{64.88}{$64.88$} &
    \heatmapcell{60.1}{$60.1$} &
    \heatmapcell{62.19}{$62.19$} &
    \heatmapcell{52.01}{$52.01$} &
    \heatmapcell{51.11}{$51.11$} & 
    \heatmapcell{40.87}{$40.87$} \\
    \midrule 
    Qwen3-8B & 1 & 
    \heatmapcell{71.98}{$71.98$} &
    \heatmapcell{74.3}{$74.3$} &
    \heatmapcell{74.28}{$74.28$} & 
    \heatmapcell{71.93}{$71.93$} &
    \heatmapcell{67.54}{$67.54$} &
    \heatmapcell{68.72}{$68.72$} \\
    & 2 & 
    \heatmapcell{62.89}{$62.89$} &
    \heatmapcell{58.54}{$58.54$} &
    \heatmapcell{58.24}{$58.24$} & 
    \heatmapcell{53.1}{$53.1$} &
    \heatmapcell{54.79}{$54.79$} &
    \heatmapcell{54.58}{$54.58$} \\
    & 3 & 
    \heatmapcell{49.87}{$49.87$} & 
    \heatmapcell{47.34}{$47.34$} &
    \heatmapcell{52.3}{$52.3$} &
    \heatmapcell{42.98}{$42.98$} &
    \heatmapcell{46.79}{$46.79$} &
    \heatmapcell{46.22}{$46.22$} \\
    & max 3 & 
    \heatmapcell{69.06}{$69.06$} &
    \heatmapcell{69.76}{$69.76$} & 
    \heatmapcell{63.86}{$63.86$} &
    \heatmapcell{64.25}{$64.25$} & 
    \heatmapcell{59.01}{$59.01$} & 
    \heatmapcell{59.83}{$59.83$} \\
    \midrule
    Gemini 2.5 Flash-Lite & 
    1 & 
    \heatmapcell{74.62}{$74.62$} & 
    \heatmapcell{73.29}{$73.29$} &
    \heatmapcell{72.67}{$72.67$} & 
    \heatmapcell{72.42}{$72.42$} &
    \heatmapcell{69.4}{$69.4$} &
    \heatmapcell{65.38}{$65.38$} \\
    & 2 & 
    \heatmapcell{69.3}{$69.3$} & 
    \heatmapcell{72.75}{$72.75$} & 
    \heatmapcell{64.79}{$64.79$} & 
    \heatmapcell{61.44}{$61.44$} &
    \heatmapcell{63.12}{$63.12$} &
    \heatmapcell{61.78}{$61.78$} \\
    & 3 & 
    \heatmapcell{71.46}{$71.46$} &
    \heatmapcell{71.25}{$71.25$} &
    \heatmapcell{63.92}{$63.92$} & 
    \heatmapcell{63.2}{$63.2$} &
    \heatmapcell{60.49}{$60.49$} &
    \heatmapcell{57.03}{$57.03$} \\
    & max 3 & 
    \heatmapcell{71.55}{$71.55$} &
    \heatmapcell{70.97}{$70.97$} &
    \heatmapcell{69.73}{$69.73$} &
    \heatmapcell{65.83}{$65.83$} & 
    \heatmapcell{65.97}{$65.97$} &
    \heatmapcell{64.98}{$64.98$} \\
    \midrule 
    Gemini 2.5 Pro & 1 &
    \heatmapcell{75.86}{$75.86$} &
    \heatmapcell{72.78}{$72.78$} &
    \heatmapcell{74.97}{$74.97$} & 
    \heatmapcell{70.45}{$70.45$} &
    \heatmapcell{69.34}{$69.34$} & 
    \heatmapcell{69.65}{$69.65$} \\
    & 2 & 
    \heatmapcell{72.66}{$72.66$} & 
    \heatmapcell{65.99}{$65.99$} &
    \heatmapcell{64.39}{$64.39$} &
    \heatmapcell{60.68}{$60.68$} & 
    \heatmapcell{60.98}{$60.98$} & 
    \heatmapcell{60.88}{$60.88$} \\
    & 3 & 
    \heatmapcell{69.06}{$69.06$} &
    \heatmapcell{70.38}{$70.38$} &
    \heatmapcell{66.36}{$66.36$} & \heatmapcell{58.63}{$58.63$} & \heatmapcell{55.98}{$55.98$} &
    \heatmapcell{54.75}{$54.75$} \\
    & max 3 & 
    \heatmapcell{68.36}{$68.36$} & 
    \heatmapcell{71.08}{$71.08$} & 
    \heatmapcell{67.85}{$67.85$} & 
    \heatmapcell{63.87}{$63.87$} & 
    \heatmapcell{66.35}{$66.35$} &
    \heatmapcell{62.03}{$62.03$} \\
    \midrule
    GPT-5 mini & 1 & 
    \heatmapcell{79.87}{$79.87$} & 
    \heatmapcell{78.04}{$78.04$} &
    \heatmapcell{79.1}{$79.1$} &
    \heatmapcell{75.78}{$75.78$} &
    \heatmapcell{75.27}{$75.27$} &
    \heatmapcell{72.51}{$72.51$} \\
    & 2 & 
    \heatmapcell{56.48}{$56.48$} &
    \heatmapcell{58.96}{$58.96$} &
    \heatmapcell{58.24}{$58.24$} &
    \heatmapcell{55.54}{$55.54$} &
    \heatmapcell{52.15}{$52.15$} &
    \heatmapcell{51.79}{$51.79$} \\
    & 3 & 
    \heatmapcell{40.81}{$40.81$} & 
    \heatmapcell{41}{$41$} &
    \heatmapcell{32.89}{$32.89$} & 
    \heatmapcell{38.66}{$38.66$} & 
    \heatmapcell{36.39}{$36.39$} &
    \heatmapcell{40.74}{$40.74$} \\
    & max 3 & 
    \heatmapcell{73.47}{$73.47$} & 
    \heatmapcell{70.4}{$70.4$} & 
    \heatmapcell{71.42}{$71.42$} & 
    \heatmapcell{68.2}{$68.2$} &
    \heatmapcell{71.81}{$71.81$} & 
    \heatmapcell{65.46}{$65.46$} \\
    \midrule 
    GPT-5 & 1 & 
    \heatmapcell{78.28}{$78.28$} & 
    \heatmapcell{77.36}{$77.36$} &
    \heatmapcell{77.2}{$77.2$} &
    \heatmapcell{75.03}{$75.03$} & \heatmapcell{72.98}{$72.98$} &
    \heatmapcell{75.15}{$75.15$} \\
    & 2 & 
    \heatmapcell{72.72}{$72.72$} & \heatmapcell{72.33}{$72.33$} &
    \heatmapcell{68.73}{$68.73$} &
    \heatmapcell{70.17}{$70.17$} &
    \heatmapcell{66.29}{$66.29$} &
    \heatmapcell{68.58}{$68.58$} \\
    & 3 & 
    \heatmapcell{71.7}{$71.7$} & 
    \heatmapcell{70.13}{$70.13$} & 
    \heatmapcell{67.2}{$67.2$} & 
    \heatmapcell{66.45}{$66.45$} &
    \heatmapcell{64.07}{$64.07$} & 
    \heatmapcell{63.31}{$63.31$} \\
    & max 3 & 
    \heatmapcell{76.47}{$76.47$} & 
    \heatmapcell{74.89}{$74.89$} &
    \heatmapcell{75.06}{$75.06$} & 
    \heatmapcell{73.56}{$73.56$} &
    \heatmapcell{74.03}{$74.03$} &
    \heatmapcell{71.96}{$71.96$} \\
\bottomrule
\end{tabular}
\end{adjustbox}
\label{tab:dynamic_NIAH_bm25_ppr}
\end{table}

\begin{table}[t]
\caption{Dynamic NIAH performance in F1 score $(\times 10^{-2}, \uparrow)$ using Qwen3-0.6B for haystack construction, where retriever-ranked haystack ordering is used. 0 stands for the case without distractors.}
\centering
\begin{adjustbox}{width=0.8\textwidth}
\begin{tabular}{lccccccc}
\toprule
Context Size (\# Tokens) & \# Rounds & 0 & 8K & 16K & 32K & 64K & 128K \\
\midrule
    Llama-3.1-8B-Instruct & 1 & 
    \heatmapcell{56.86}{$56.86$} &
    \heatmapcell{56.76}{$56.76$} &
    \heatmapcell{53.41}{$53.41$} &
    \heatmapcell{49.45}{$49.45$} &
    \heatmapcell{42.81}{$42.81$} &
    \heatmapcell{30.15}{$30.15$} \\
    & 2 & 
    \heatmapcell{56.68}{$56.68$} &
    \heatmapcell{45.59}{$45.59$} &
    \heatmapcell{51.41}{$51.41$} & 
    \heatmapcell{39.18}{$39.18$} &
    \heatmapcell{37.71}{$37.71$} &
    \heatmapcell{27.88}{$27.88$} \\
    & 3 &
    \heatmapcell{36.68}{$36.68$} &
    \heatmapcell{30.48}{$30.48$} &
    \heatmapcell{38.74}{$38.74$} &
    \heatmapcell{27.87}{$27.87$} &
    \heatmapcell{29.65}{$29.65$} &
    \heatmapcell{26.08}{$26.08$} \\
    & max 3 & 
    \heatmapcell{38.5}{$38.5$} & 
    \heatmapcell{31.12}{$31.12$} &
    \heatmapcell{37.8}{$37.8$} &
    \heatmapcell{31.7}{$31.7$} &
    \heatmapcell{33.55}{$33.55$} & 
    \heatmapcell{25.24}{$25.24$} \\
    \midrule
    Qwen2.5-7B-Instruct-1M & 
    1 & 
    \heatmapcell{60.59}{$60.59$} & 
    \heatmapcell{55.45}{$55.45$} &
    \heatmapcell{50.54}{$50.54$} &
    \heatmapcell{44.36}{$44.36$} &
    \heatmapcell{39.97}{$39.97$} &
    \heatmapcell{35.51}{$35.51$} \\
    & 2 & 
    \heatmapcell{32.44}{$32.44$} &
    \heatmapcell{35.22}{$35.22$} &
    \heatmapcell{37.24}{$37.24$} &
    \heatmapcell{37.41}{$37.41$} &
    \heatmapcell{38.49}{$38.49$} & 
    \heatmapcell{35.63}{$35.63$} \\
    & 3 & 
    \heatmapcell{26.1}{$26.1$} & 
    \heatmapcell{23.45}{$23.45$} &
    \heatmapcell{22.02}{$22.02$} &
    \heatmapcell{24.84}{$24.84$} &
    \heatmapcell{20.33}{$20.33$} &
    \heatmapcell{27.18}{$27.18$} \\
    & max 3 & 
    \heatmapcell{44.36}{$44.36$} & 
    \heatmapcell{36.92}{$36.92$} &
    \heatmapcell{33.9}{$33.9$} &
    \heatmapcell{31.46}{$31.46$} & 
    \heatmapcell{38.7}{$38.7$} & 
    \heatmapcell{38.12}{$38.12$} \\
    \midrule 
    Gemma-3-12B-IT & 1 &
    \heatmapcell{66.54}{$66.54$} &
    \heatmapcell{63.77}{$63.77$} & 
    \heatmapcell{58.67}{$58.67$} & 
    \heatmapcell{52.66}{$52.66$} &
    \heatmapcell{55.18}{$55.18$} & 
    \heatmapcell{47.59}{$47.59$} \\
    & 2 & 
    \heatmapcell{55.13}{$55.13$} &
    \heatmapcell{51.63}{$51.63$} &
    \heatmapcell{49.39}{$49.39$} &
    \heatmapcell{47.28}{$47.28$} &
    \heatmapcell{44.81}{$44.81$} &
    \heatmapcell{35.55}{$35.55$} \\
    & 3 & 
    \heatmapcell{40.07}{$40.07$} & 
    \heatmapcell{43.32}{$43.32$} & 
    \heatmapcell{37.54}{$37.54$} &
    \heatmapcell{32.23}{$32.23$} & 
    \heatmapcell{32.57}{$32.57$} &
    \heatmapcell{23.7}{$23.7$} \\
    & max 3 & 
    \heatmapcell{61.37}{$61.37$} &
    \heatmapcell{63.83}{$63.83$} & 
    \heatmapcell{59.1}{$59.1$} & 
    \heatmapcell{58.86}{$58.86$} & 
    \heatmapcell{57.05}{$57.05$} & 
    \heatmapcell{50.16}{$50.16$} \\
    \midrule 
    Qwen3-8B & 1 & 
    \heatmapcell{71.98}{$71.98$} & 
    \heatmapcell{69.61}{$69.61$} & 
    \heatmapcell{65.56}{$65.56$} &
    \heatmapcell{69.44}{$69.44$} &
    \heatmapcell{62.41}{$62.41$} &
    \heatmapcell{55.57}{$55.57$}\\
    & 2 & 
    \heatmapcell{58.95}{$58.95$} &
    \heatmapcell{59.54}{$59.54$} & 
    \heatmapcell{62.48}{$62.48$} &
    \heatmapcell{63.91}{$63.91$} & 
    \heatmapcell{57.94}{$57.94$} &
    \heatmapcell{49.83}{$49.83$} \\
    & 3 & 
    \heatmapcell{49.26}{$49.26$} &
    \heatmapcell{46.95}{$46.95$} &
    \heatmapcell{50.32}{$50.32$} & 
    \heatmapcell{49.84}{$49.84$} &
    \heatmapcell{40.14}{$40.14$} &
    \heatmapcell{41.64}{$41.64$} \\
    & max 3 & 
    \heatmapcell{69.48}{$69.48$} &
    \heatmapcell{65.31}{$65.31$} & 
    \heatmapcell{63.26}{$63.26$} & 
    \heatmapcell{59.78}{$59.78$} & 
    \heatmapcell{55.2}{$55.2$} & 
    \heatmapcell{54.32}{$54.32$} \\
    \midrule
    Gemini 2.5 Flash-Lite & 
    1 & 
    \heatmapcell{74.62}{$74.62$} & 
    \heatmapcell{72.25}{$72.25$} &
    \heatmapcell{73.99}{$73.99$} & 
    \heatmapcell{71.92}{$71.92$} &
    \heatmapcell{70.23}{$70.23$} &
    \heatmapcell{63.28}{$63.28$} \\
    & 
    2 & 
    \heatmapcell{68.04}{$68.04$} & 
    \heatmapcell{66.85}{$66.85$} & 
    \heatmapcell{65.56}{$65.56$} & 
    \heatmapcell{64.87}{$64.87$} &
    \heatmapcell{62.67}{$62.67$} &
    \heatmapcell{62.76}{$62.76$} \\
     &
    3 & 
    \heatmapcell{69.62}{$69.62$} & 
    \heatmapcell{64.97}{$64.97$} &
    \heatmapcell{64.6}{$64.6$} &
    \heatmapcell{58.91}{$58.91$} &
    \heatmapcell{63.56}{$63.56$} &
    \heatmapcell{63.23}{$63.23$} \\
    & max 3 & 
    \heatmapcell{69.35}{$69.35$} & 
    \heatmapcell{66.77}{$66.77$} &
    \heatmapcell{67.34}{$67.34$} & 
    \heatmapcell{68.09}{$68.09$} &
    \heatmapcell{63.23}{$63.23$} & 
    \heatmapcell{64.87}{$64.87$} \\
    \midrule 
    Gemini 2.5 Pro &
    1 & 
    \heatmapcell{75.86}{$75.86$} & 
    \heatmapcell{73.34}{$73.34$} & 
    \heatmapcell{72.20}{$72.20$} &
    \heatmapcell{73.9}{$73.9$} &
    \heatmapcell{69.61}{$69.61$} &
    \heatmapcell{68.52}{$68.52$} \\
    & 2 & 
    \heatmapcell{69.07}{$69.07$} & 
    \heatmapcell{65.81}{$65.81$} & 
    \heatmapcell{65.95}{$65.95$} &
    \heatmapcell{62.05}{$62.05$} &
    \heatmapcell{61.16}{$61.16$} &
    \heatmapcell{58.08}{$58.08$} \\
    & 3 &
    \heatmapcell{67.26}{$67.26$} &
    \heatmapcell{63.84}{$63.84$} &
    \heatmapcell{62.91}{$62.91$} &
    \heatmapcell{61.02}{$61.02$} &
    \heatmapcell{53.09}{$53.09$} &
    \heatmapcell{51.43}{$51.43$} \\
    & max 3 &
    \heatmapcell{71.37}{$71.37$} &
    \heatmapcell{68.28}{$68.28$} & 
    \heatmapcell{65.16}{$65.16$} & 
    \heatmapcell{63.12}{$63.12$} & 
    \heatmapcell{59.73}{$59.73$} &
    \heatmapcell{60.93}{$60.93$} \\
    \midrule
    GPT-5 mini & 1 & 
    \heatmapcell{79.87}{$79.87$} &
    \heatmapcell{77.18}{$77.18$} &
    \heatmapcell{76.97}{$76.97$} &
    \heatmapcell{73.68}{$73.68$} &
    \heatmapcell{72.14}{$72.14$} &
    \heatmapcell{70.14}{$70.14$} \\
    & 2 & 
    \heatmapcell{60.24}{$60.24$} & 
    \heatmapcell{57.4}{$57.4$} &
    \heatmapcell{57.2}{$57.2$} & 
    \heatmapcell{53.82}{$53.82$} & 
    \heatmapcell{51.23}{$51.23$} & 
    \heatmapcell{53.33}{$53.33$} \\
    & 3 & 
    \heatmapcell{39.63}{$39.63$} & 
    \heatmapcell{41.02}{$41.02$} & 
    \heatmapcell{38.99}{$38.99$} & 
    \heatmapcell{35.26}{$35.26$} &
    \heatmapcell{37.73}{$37.73$} & 
    \heatmapcell{37.92}{$37.92$} \\
    & max 3 & 
    \heatmapcell{73.27}{$73.27$} &
    \heatmapcell{71.7}{$71.7$} & 
    \heatmapcell{70.77}{$70.77$} & 
    \heatmapcell{68.98}{$68.98$} &
    \heatmapcell{67.59}{$67.59$} &
    \heatmapcell{65.59}{$65.59$} \\
    \midrule 
    GPT-5 & 1 & 
    \heatmapcell{78.28}{$78.28$} &
    \heatmapcell{74.8}{$74.8$} &
    \heatmapcell{76.31}{$76.31$} &
    \heatmapcell{76.98}{$76.98$} &
    \heatmapcell{73.47}{$73.47$} & 
    \heatmapcell{71.7}{$71.7$} \\
    & 2 & 
    \heatmapcell{71.92}{$71.92$} & 
    \heatmapcell{69.38}{$69.38$} &
    \heatmapcell{70.28}{$70.28$} &
    \heatmapcell{72.96}{$72.96$} & 
    \heatmapcell{68.26}{$68.26$} &
    \heatmapcell{66.3}{$66.3$} \\
    & 3 & 
    \heatmapcell{71.21}{$71.21$} & 
    \heatmapcell{68.35}{$68.35$} &
    \heatmapcell{63.01}{$63.01$} &
    \heatmapcell{68.52}{$68.52$} & 
    \heatmapcell{63.35}{$63.35$} &
    \heatmapcell{65.48}{$65.48$} \\
    & max 3 & 
    \heatmapcell{76.22}{$76.22$} & 
    \heatmapcell{76.22}{$76.16$} & 
    \heatmapcell{71.92}{$71.92$} &
    \heatmapcell{71.15}{$71.15$} &
    \heatmapcell{72.67}{$72.67$} & 
    \heatmapcell{72.95}{$72.95$} \\
\bottomrule
\end{tabular}
\end{adjustbox}
\label{tab:dynamic_NIAH_qwen}
\end{table}

\section{Implementation Details}

We employ vLLM for LLM inference~\citep{kwon2023efficient}.

\end{document}